\documentclass[runningheads]{llncs}

\usepackage{graphicx}

    \usepackage{amsmath}
    \usepackage{amssymb}
    \usepackage{booktabs}

    \usepackage[table]{xcolor}
    
    \usepackage{tabularx}
    \newcolumntype{Y}{>{\centering\arraybackslash}X}
    \usepackage{array}
    
    \usepackage{float}
    \usepackage{array,multirow,graphicx}
    \graphicspath{ {./images/} }
    
    \usepackage{mathtools}
    
    \usepackage[justification=centering,skip=0pt]{subcaption}
    \usepackage{placeins}
    \usepackage[export]{adjustbox}
    \usepackage{amsmath}
    \usepackage{xspace}
    \usepackage{makecell}

    \usepackage[pagebackref,breaklinks,colorlinks]{hyperref}

\usepackage{tikz}
\usepackage{comment}
\usepackage{amsmath,amssymb} 
\usepackage{color}

\usepackage[accsupp]{axessibility}

    \usepackage[capitalize]{cleveref}
    \crefname{section}{Sec.}{Secs.}
    \Crefname{section}{Section}{Sections}
    \Crefname{table}{Table}{Tables}
    \crefname{table}{Tab.}{Tabs.}
    \crefname{appendix}{App.}{Apps.}
    \Crefname{appendix}{Appendix}{Appendices}
    \crefname{equation}{Eq.}{Eqs.}
    \Crefname{equation}{Equation}{Equation}

    \DeclareMathOperator*{\argmax}{arg\,max}
    \DeclareMathOperator*{\argmin}{arg\,min}
    \DeclareMathOperator{\sign}{sign}
    \newcolumntype{C}[1]{>{\centering\arraybackslash}p{#1}}
    \newcolumntype{L}[1]{>{\raggedright\arraybackslash}p{#1}}

    \newtheorem{refproof}{Proof}

    \newcommand{\inner}[1]{\left\langle#1\right\rangle}
    \def\R{\mathbb{R}}

    \newcommand{\norm}[1]{\left\|#1\right\|}

    \def\argmax{\mathop{\rm arg\,max}\limits}
    \def\argmin{\mathop{\rm arg\,min}\limits}
    
    \def\maxop{\mathop{\rm max}\limits}
    \def\sign{\mathop{\rm sign}\limits}

    \def\min{\mathop{\rm min}\nolimits}
    \def\max{\mathop{\rm max}\nolimits}

    \newcolumntype{C}[1]{>{\centering\arraybackslash}p{#1}}
    \newcolumntype{L}[1]{>{\raggedright\arraybackslash}p{#1}}
    \newcolumntype{R}[1]{>{\raggedleft\arraybackslash}p{#1}}

    \newcommand{\changed}[1]{{#1}}

\def\onedot{\ifx\@let@token.\else.\null\fi\xspace}
\def\eg{\emph{e.g}\onedot}

\newif\ifreview
\reviewfalse

\ifreview
	\usepackage{lineno}

	\linenumbers
\fi

\begin{document}

\def\SubNumber{89}

\def\GCPRTrack{Fast Review Track}

\title{Sparse Visual Counterfactual Explanations in Image Space}

\ifreview
\else

	\author{Valentyn Boreiko \and
	Maximilian Augustin \and
	Francesco Croce \and Philipp Berens \and Matthias Hein}
	
	\authorrunning{V. Boreiko et al.}

	\institute{University of Tübingen}
\fi

\maketitle              

\begin{abstract}
       Visual counterfactual explanations (VCEs) in image space are an important tool to understand decisions of image classifiers as they show under which changes of the image the decision of the classifier would change. Their generation in image space is challenging and requires robust models due to the problem of adversarial examples.
       Existing techniques to generate VCEs in image space suffer from spurious changes in the background.
       Our novel perturbation model for VCEs together with its efficient optimization via our novel Auto-Frank-Wolfe scheme yields sparse VCEs which lead to subtle changes specific for the target class.
       Moreover, we show
       that VCEs can be used to detect undesired behavior of ImageNet classifiers due to spurious features in the ImageNet dataset. Code is available under \url{https://github.com/valentyn1boreiko/SVCEs_code}.
    \end{abstract}

    \section{Introduction}
    The black-box nature of decisions made by neural networks is one of the main obstacles for the widespread use of machine learning in industry and science. It is likely that future regulatory steps will strengthen the ``right for an explanation'', which is currently already implemented in a weak form in the GDPR \cite{wachter2018counterfactual} and is included as ``transparency of an AI system'' in a draft for regulating AI of the European Union, at least concerning the use of AI in safety critical systems \cite{EU2021}. Apart from directly interpretable classifiers like linear models or decision trees, a variety of model-agnostic explanation techniques has been proposed: sensitivity based explanations \cite{BaeEtAl2010}, explanations based on feature attributions \cite{BacEtAl2015}, saliency maps \cite{simonyan2014deep,Selvaraju_2019,etmann2019connection,wang2020smoothed,SrinivasFleuret2019}, Shapley additive explanations \cite{SHAPley_explanations}, and local fits of interpretable models \cite{Ribeiro-Lime}, see \cite{marcinkev2020interpretability} for a recent overview. 
    
    \setlength\tabcolsep{1pt}
    
    \begin{figure*}[hbt!]
     
     \centering
     \begin{tabular}{c|c|c}
                   \hline
                   \multicolumn{1}{c}{Original}&\multicolumn{1}{C{.4\textwidth}}{AFW, $l_{1.5}$}&\multicolumn{1}{C{.4\textwidth}}{APGD, $l_2$}\\
                   \hline
     \begin{subfigure}{0.2\textwidth}\centering
     
     \caption*{{\normalsize cougar: 0.42}}
     \includegraphics[width=.96\textwidth]{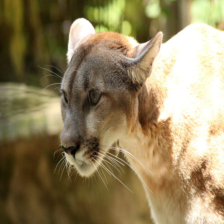}
     \end{subfigure}&
     \begin{subfigure}{0.4\textwidth}\centering
     
     \caption*{{\normalsize$\rightarrow$cheetah: 0.99}}
                     
     \includegraphics[width=.48\textwidth]{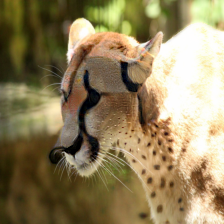}
     \includegraphics[width=.48\textwidth]{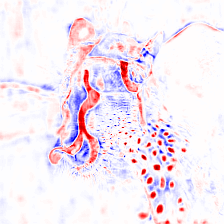}
     \end{subfigure}&\begin{subfigure}{0.4\textwidth}\centering
     
     \caption*{{\normalsize$\rightarrow$cheetah: 0.99}}
                     
     \includegraphics[width=.48\textwidth]{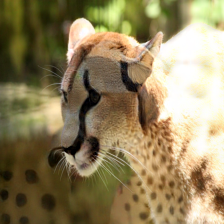}
     \includegraphics[width=.48\textwidth]{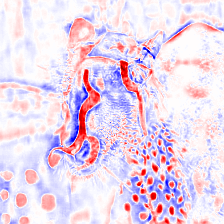}
     
     \end{subfigure}

      \end{tabular}
     
     \caption{\label{fig:motivating_example} 
     \textbf{VCEs for two different threat models.} VCEs together with difference maps for the change ``cougar $\longrightarrow$ cheetah'' for an adversarially robust ImageNet model \cite{engstrom2019adversarial_short,croce2021adversarial_short}. Our novel $l_{1.5}$-VCEs yield more sparse changes which are mainly focused on the object compared to the previously considered $l_2$-VCEs \cite{santurkar2019image_short,augustin2020_short}.
      }

     \end{figure*}

    Another candidate are counterfactual explanations (CEs) introduced in \cite{wachter2018counterfactual} as a form of instance-specific explanations close to human reasoning \cite{Mil2017}. Humans often justify  decisions by
    counterfactual reasoning: ``I would have decided for $X$, if $Y$ had been true''. One of their biggest advantages in contrast to feature attribution and other mentioned above methods is that CEs \cite{wachter2018counterfactual,dhurandhar2018explanations,Mothilal2020counterfactual,Barocas2020,Pawlowski2020counterfactual,verma2020counterfactual,schut2021generating} are actionable explanations \cite{wachter2018counterfactual} and thus are close to what the GDPR requires. Current approaches to generate CEs for classifier decisions can be summarized by answering the question:  ``What is the minimal change $\delta$ of the input $x$, so that the perturbed input $x+\delta$ is classified as the desired target class with sufficiently high confidence and is realistic?''. From the developer's perspective, counterfactuals are interesting for debugging as they allow  to detect spurious features which the classifier has picked up. We refer to \cite{verma2020counterfactual} for a recent extensive overview on the literature of counterfactual explanations who note five criteria for CEs: i) \textbf{validity:} the changed input $x+\delta$ should have the desired target class, ii) \textbf{actionability:} the change $\delta$ should be possible to be realized by the human, iii) \textbf{sparsity:} the change $\delta$ should be sparse so that the change is interpretable for humans, iv) \textbf{realism:} the changed input $x+\delta$ should lie close to the data manifold, v) \textbf{causality:} CEs should maintain causal relations between features. 
    Interestingly,  \cite{verma2020counterfactual} noted that most papers they reviewed just evaluate on tabular datasets or MNIST. The reason for this is that the process of generating CEs for high-dimensional image datasets (which we will refer to as visual counterfactual explanations, or VCEs for short) is very similar to that of generating adversarial examples \cite{SzeEtAl2014} which just exploit non-robust features of the classifier and thus show no class-specific changes required for VCEs. Thus, realistic VCEs require either (adversarially) robust models as in \cite{santurkar2019image_short,augustin2020_short} or that the images are implicitly restricted via the usage of a generative model 
   \cite{HenEtAl2016,HenEtAl2018,SamEtAl2018,chang2018explaining,GoyEtAl2019,schutte2021using}. Very recently visual counterfactuals based on generative models have been proposed \cite{explaining_in_style,GoyEtAl2019,sanchez2022diffusion} but no code has been released so far or it is restricted to MNIST. These methods require to specify the amount of ``classifier guidance'' which might be difficult to be choose as we discuss in \cref{app:diffusion-vces}. For this reason, in this work we investigate the generation of VCEs directly in image space, instead of working in the latent space, and purely based on the classifier, thus showing its behavior without the influence of an auxiliary model.
    We make the following contributions: 
    i) we show that the $l_2$-metric used for the generation of VCEs in \cite{santurkar2019image_short,augustin2020_short} leads to changes all over the image (see \cref{fig:motivating_example}) which are unrelated to the object. This is in particular true for ImageNet models;
    ii) we propose a new model for sparse VCEs based on the $l_p$-metric for $p=1.5$. Since an efficient projection onto $l_{1.5}$-balls is not available, we develop a novel Auto-Frank-Wolfe (AFW) optimization scheme with an adaptive step-size for the generation of $l_{1.5}$-VCEs. The resulting VCEs are more sparse and ``subtle'' as confirmed by a user study;
    iii) we illustrate that VCEs are useful to detect spurious features in ImageNet classifiers, e.g., we detect the spurious feature ``watermark'' in the class granny smith due to a bias in the training set and
    show that our findings transfer to other ImageNet classifiers. This shows the utility of VCEs as a ``debugging tool'' for ML classifiers.

    \section{Visual Counterfactual Explanations (VCEs)}\label{sec:vces_intro}
    In this section, we first discuss the previously considered formulation of Visual Counterfactual Explanations (VCEs) of \cite{augustin2020_short} in the image space and the required kind of (adversarial) robustness of the classifier.
    Then we discuss a novel perturbation model which overcomes the partially non-object-related changes of the VCEs proposed in \cite{augustin2020_short}. For the optimization over this perturbation model we provide in \cref{sec:afw} a novel adaptive Frank-Wolfe scheme. 
    
    We assume in the paper that the classifier, 
    $f:\R^d \rightarrow \Delta_K$, where $\Delta_K := \{ w \in \mathbb{R}_{\geq 0}^K | \sum_{i=1}^K w_i = 1 \}$ is the probability simplex, outputs for every input $x$ a
    probability distribution $\hat{p}_f(y|x)$ ($y \in \{1,\ldots,K\}$) over the classes. The $l_p$-distance on $\R^d$ is defined as: $\norm{x-y}_p = \big(\sum_{i=1}^d |x_i-y_i|^p\big)^\frac{1}{p}.$

    \subsection{Formulation {
    and properties} of 
    VCEs}\label{sec:vces_definition}

    Counterfactual explanations for a given classifier are instance-wise explanations.
    Informally speaking, a visual counterfactual explanation for an input image $x_0$ is a new image $\hat{x}$ which is visually similar and as close as possible to a real image, but class-specific features have been changed such that the classifier now assigns to $\hat{x}$ a desired target class different from than one assigned to $x_0$ (counterfactual). In addition, it is often interesting which features appear if one aims to make the classifier maximally confident in its decision (same as for $x_0$).
    
    {
    \textbf{VCEs via constrained optimization:}} In \cite{wachter2018counterfactual}  (see also  \cite{Mothilal2020counterfactual,verma2020counterfactual}) they suggest to determine counterfactuals by the following optimization problem:
    \begin{equation}\label{eq:CE}
    \hat{x} = \argmin_{x \in \R^d} L(k,f(x)) + \lambda \, d(x_0,x),
    \end{equation}
    where $L$ is a loss function, e.g. cross-entropy loss, $L(k,f(x))=-\log \hat{p}_f(k|x)$, $k$ is the desired target class and $d:\R^d \times \R^d \rightarrow \R$ a distance, measuring similarity of $x_0$ and $\hat{x}$. If the decision of the classifier for $\hat{x}$ changes to the target class $k$, then the counterfactual is ``valid''. The advantage of valid counterfactuals, compared to feature attribution methods or other instance-wise explanation techniques, 
    is that the change $\hat{x}-x_0$ is actionable, in the sense that the user understands, how to influence and change the decision of the classifier.
    As $\lambda$ has no direct interpretation, we employ the related and more interpretable objective of \cite{augustin2020_short}
    \begin{equation}\label{eq:VCE-RATIO} \argmax_{x \in [0,1]^d \cap B(x_0,\epsilon)} \log \hat{p}_f(k|x),
    \end{equation}
    where $B(x_0,\epsilon)=\{x \in \R^d \,|\, d(x,x_0)\leq \epsilon\}$. The constraint, $x \in [0,1]^d$,
    is necessary as we want to generate valid images. The choice of the distance metric is crucial for the quality of the VCEs (see \cref{sec:threatmodel}). The new free parameter $\epsilon$ can be interpreted as ``perturbation budget'' with respect to the chosen metric. 
\begin{figure*}[ht]
     \centering
     \setlength\tabcolsep{1pt}
     \begin{tabular}{c|c|c|c|c}
                   \cline{2-5}
                   \multicolumn{1}{c|}{} & \multicolumn{1}{c}{Original}&\multicolumn{1}{C{.27\textwidth}}{Non-robust}&\multicolumn{1}{C{.27\textwidth}}{Madry\cite{robustness_short}} &\multicolumn{1}{C{.27\textwidth}}{Madry\cite{robustness_short} + FT}\\
                   \hline
    \rotatebox[origin=c]{90}{ImageNet} & 
     \begin{subfigure}{0.135\textwidth}\centering
     \caption*{{\normalsize $\rightarrow$megalith}}
     \includegraphics[width=.96\textwidth]{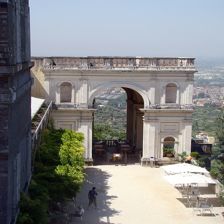}
     \end{subfigure}&
     \begin{subfigure}{0.27\textwidth}\centering
     \caption*{{\normalsize $p_\mathrm{i}$:0.00, $p_\mathrm{e}$:1.00}}

      \includegraphics[width=.48\textwidth]{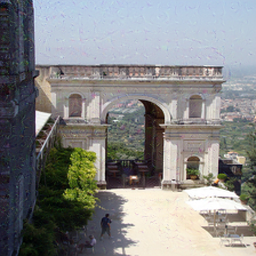}
      \includegraphics[width=.48\textwidth]{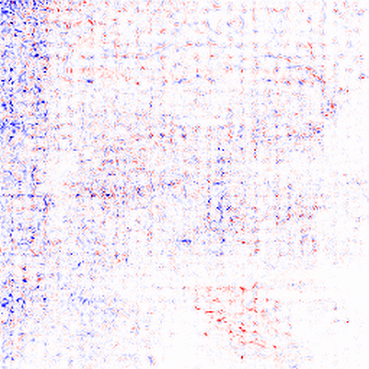}
     \end{subfigure}&
     \begin{subfigure}{0.27\textwidth}\centering
     \caption*{{\normalsize $p_\mathrm{i}$:0.00, $p_\mathrm{e}$:1.00}}
     \includegraphics[width=.48\textwidth]{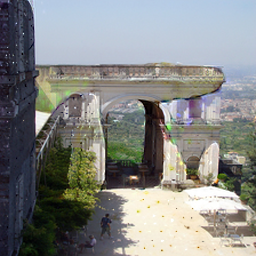}
     \includegraphics[width=.48\textwidth]{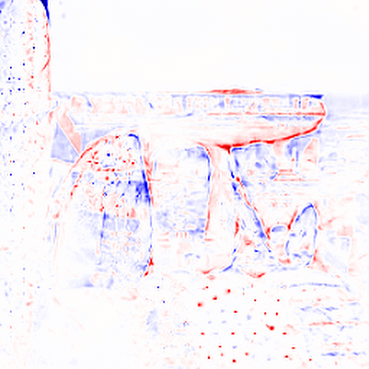}
     \end{subfigure} &
     \begin{subfigure}{0.27\textwidth}\centering
    \caption*{{\normalsize $p_\mathrm{i}$:0.00, $p_\mathrm{e}$:1.00}}
     \includegraphics[width=.48\textwidth]{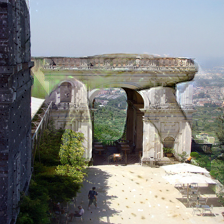}
     \includegraphics[width=.48\textwidth]{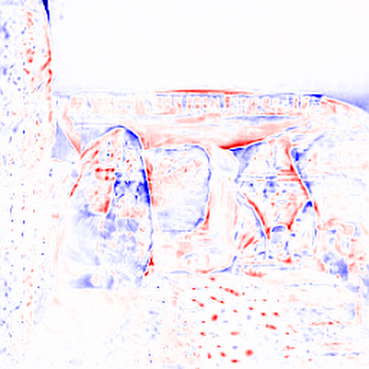}
     \end{subfigure}\\
     \hline
     \rotatebox[origin=c]{90}{ImageNet-O} & 
     \begin{subfigure}{0.135\textwidth}\centering
     \caption*{{\normalsize$\rightarrow$jellyfish}}
     \includegraphics[width=.96\textwidth]{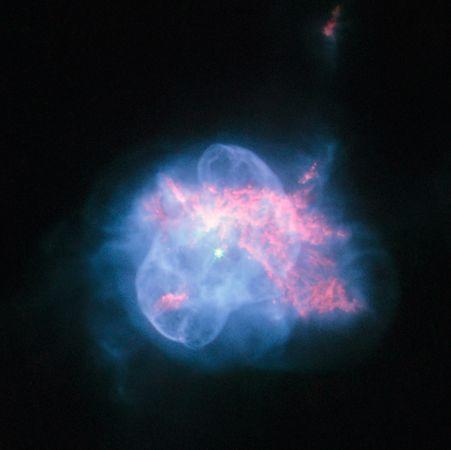}
     \end{subfigure}&
     \begin{subfigure}{0.27\textwidth}\centering
     \caption*{{\normalsize $p_\mathrm{i}$:0.47, $p_\mathrm{e}$:1.00}}

     \includegraphics[width=.48\textwidth]{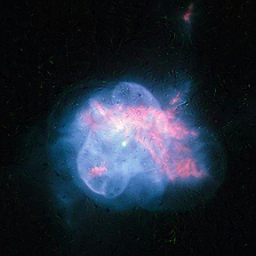}
     \includegraphics[width=.48\textwidth]{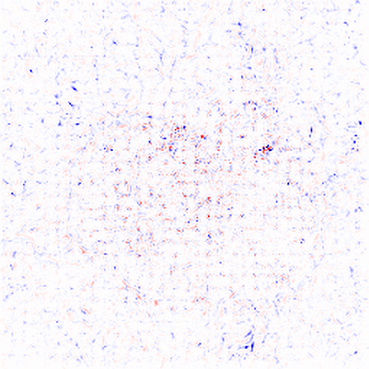}
     \end{subfigure}&
     \begin{subfigure}{0.27\textwidth}\centering
     \caption*{{\normalsize $p_\mathrm{i}$:0.75, $p_\mathrm{e}$:1.00}}
     \includegraphics[width=.48\textwidth]{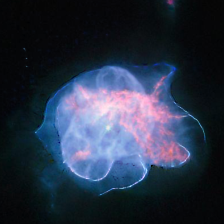}
     \includegraphics[width=.48\textwidth]{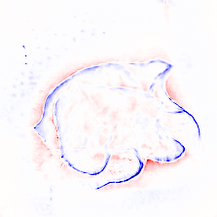}
     \end{subfigure} &
     \begin{subfigure}{0.27\textwidth}\centering
    \caption*{{\normalsize $p_\mathrm{i}$:0.66, $p_\mathrm{e}$:1.00}}
     \includegraphics[width=.48\textwidth]{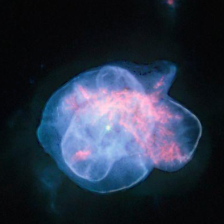}
     \includegraphics[width=.48\textwidth]{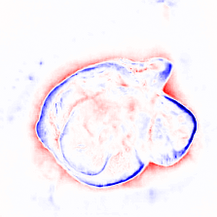}
     \end{subfigure}\\

     \hline
      \end{tabular}
     \caption{\label{fig:motivating_robustness} 
    \textbf{Dependence of $l_{1.5}$-VCEs on robustness of the model (for in- and out-of-distribution).} VCEs for a non-robust ResNet50, a robust ResNet trained with $l_2$-Adversarial Training \cite{robustness_short} and the same model finetuned for multiple norm robustness (Madry\cite{robustness_short}+FT\cite{croce2021adversarial_short}). For the non-robust model, VCEs 
     achieve high confidence without introducing any class-specific features or meaningful changes. For the robust models, VCEs  only achieve high confidence by adding meaningful changes for the given target class (indicated by $\rightarrow$ in the original image). The multiple-norm robust model yields higher quality images than the original $l_2$-robust model.
      }
     \end{figure*}
    
    {\textbf{VCEs and Robustness:}}
    It has  been noted in \cite{wachter2018counterfactual} that counterfactuals generated via \cref{eq:CE} are equivalent to targeted adversarial examples. In \cite{wachter2018counterfactual} this did not cause problems as they only handled very low-dimensional problems.
    {
    In fact, adversarial attacks often maximize a surrogate loss, in this case the log-probability, to induce misclassification into the target class.}
    However, adversarial attacks \cite{SzeEtAl2014,MadEtAl2018} on non-robust image classifiers typically show no class-specific changes, see Fig. \ref{fig:motivating_robustness}. 
    {
    The standard method to increase robustness to adversarial attacks is adversarial training \cite{MadEtAl2018} based on projected gradient descent (PGD). Notably,} \cite{TsiEtAl18,santurkar2019image_short} have observed that adversarially robust models  have strong generative properties, {
    which is closely related to the explainability of the classifier decisions}. In \cref{fig:motivating_robustness} we show the VCEs for a robust ResNet50\cite{robustness_short} trained with adversarial training (Madry \cite{robustness_short}) and the same model enhanced with multiple-norm finetuning \cite{croce2021adversarial_short} (Madry\cite{robustness_short}+FT) and a non-robust model. The examples confirm that for meaningful VCE generation, a robust model is needed. 
    Throughout the rest of the paper, we show VCEs for the robust Madry\cite{robustness_short}+FT model. In the Appendix, we furthermore explore what kind of robustness is required for the VCE generation and how VCEs differ between robust models on both ImageNet and CIFAR10.

    {
    \textbf{Properties of VCEs:}}
    Following \cite{Mothilal2020counterfactual,verma2020counterfactual}, we aim to achieve the following main properties for our VCEs: i) \textbf{validity:} 
        from \cref{eq:VCE-RATIO}, 
        one sees that, for 
        a given perturbation budget, we find the VCE with maximum  probability $\hat{p}_f( k | \hat{x})$ in the target class $k$ for 
        $f$; ii) \textbf{sparsity:} $\hat{x}$ should be visually similar to $x_0$ and only contain sparse changes which is exactly the reason for our 
        considered distance metric, see \cref{sec:threatmodel}; iii) \textbf{realism:} $\hat{x}$ should lie on the data manifold and look like a real image. For qualitatitve results we show examples of VCEs and for quantitative analysis we use the Frechet Inception Distance (FID) \cite{heusel2017fid} both on VCEs generated from in-distribution test set, and from out-distribution samples, see \cref{sec:evaluation}, and a user-study in \cref{sec:threatmodel}.

    We stress that our primary goal is to explain the inner workings of  a given classifier and not necessarily to generate the best looking images. We demonstrate in \cref{sec:debugging} that our VCEs can be successfully used to reveal undesired behavior of ImageNet classifiers due to biases in the ImageNet dataset.

      \begin{table}[t] 
       
        \caption{\textbf{ImageNet:}
        Accuracy and $l_{1.5}$-, $l_2$-robust accuracy (RA) at $\epsilon_{1.5}=12.5,\epsilon_2=2$ for the $l_2$-adv. robust model of Madry\cite{robustness_short}
        and the fine-tuned Madry\cite{robustness_short}+FT for multiple-norm robustness \cite{croce2021adversarial_short}, and
        FID scores for $l_1, l_{1.5}$- and $l_2$-VCEs generated on in(ID)- and out-distribution(OD) images and their average. The Madry\cite{robustness_short}+FT model achieves the best FID score for $l_{1.5}$-VCEs.
        \label{tab:l1.5_FIDs}}
       \centering
       \small
      \begin{tabular}{c|c|c|c||c|c|c}

         &
         \multicolumn{3}{c}{\textbf{Accuracies}}&
         \multicolumn{3}{c}{\textbf{FID scores (ID/OD/AVG)}}\\
         \hline

           & Acc.
           & $l_2$-RA
           & $l_{1.5}$-RA
           & $l_{1}$-VCE, $\epsilon=400$
           & $l_{1.5}$-VCE, $\epsilon=50$ 
           & $l_2$-VCE, $\epsilon=12$
           \\

        \hline
        Madry\cite{robustness_short} 
        & \cellcolor{green!25} 57.9
        & \cellcolor{green!25} 45.7
        & 37.4
        & 13.6/41.6/27.6
        & \textbf{8.4}/24.3/16.4 
        & \textbf{8.4/22.8/15.6}\\
        \cite{robustness_short} +FT 
        & 57.5
        & 44.6
        & \cellcolor{green!25} 40.1
        & 9.6/35.7/22.6
        &  \textbf{6.9/22.6/14.8} 
        & 7.9/23.1/15.5
        \\
        \hline
      \end{tabular}

    \end{table}

    \subsection{{
    Generation and} evaluation of VCEs}\label{sec:evaluation}
    {
    We generate VCEs by approximately solving the non-convex problem \cref{eq:VCE-RATIO} with a small computational bugdet. We thus use the efficient APGD \cite{croce2020reliable} (only available for $l_1,l_2$ and $l_\infty$) or our adaptive Frank-Wolfe scheme AFW (see Section \ref{sec:afw}). For both we use a budget of 5 random restarts each with 75 iterations.}
    Typical deep learning classifiers are not calibrated, that is their decisions are either over- or underconfident \cite{pmlr-v70-guo17a}. We calibrate them using temperature rescaling by minimizing the expected calibration error (ECE) on a holdout validation set, so that confidence values are comparable, see \cref{app:calibration}.

     For the quantitative evaluation of the image quality of VCEs produced by different methods and classifiers,
     we use several metrics. First, FID scores \cite{heusel2017fid} by generating $10.000$ VCEs from the test set for the in-distribution (ID) evaluation where the target class is the second most likely class computed by using an ensemble of all classifiers used in the comparison, see  
     \cref{tab:benchmark-cifar10} (top) in \cref{app:further-evaluation}. An evaluation using FID scores on the in-distribution test (FID ID) set only is in our setting problematic, \changed{as methods with no (or minimal) change would get the best FID-score}. Thus, we also use  VCEs generated from out-of-distribution images 
     (ImagetNet-A and ImageNet-O \cite{hendrycks2021nae}) where the target label corresponds to the decision of an ensemble with all classifiers used in the comparison, 
     see \cref{tab:benchmark-cifar10} (bottom) in \cref{app:further-evaluation}.
     As  out-of-distribution images are not part of the data distribution, non-trivial changes are required to turn them into images of the in-distribution. Thus methods with almost no change will suffer here from large FID scores as the images are far from the in-distribution. In our experience from the qualitative inspection of the images, the average (AVG) of FID-scores on  in-distribution (FID ID) and  out-of-distribution images (FID OD) reflects best the realism and quality of the VCEs. Note that our FID scores cannot be directly compared to the ones of generative models as VCEs are not based on sampling. We just use the FID scores as a quantitative way to compare the different classifiers and perturbation models for VCEs. Moreover, we evaluate the utility of $l_{p}$-VCEs in a user study in \cref{sec:threatmodel}.

    \begin{figure*}[t]
     
     \centering
     \small
     \begin{tabular}{c|c|cc|cc|cc}
    \cline{2-8}
    \multicolumn{1}{c|}{} & \multicolumn{1}{C{.11\textwidth}}{{
    Original}} 
     & \multicolumn{1}{C{.13\textwidth}}{$l_{1}$, $\epsilon=400$}
     & \multicolumn{1}{C{.13\textwidth}}{Diff. map}
    &\multicolumn{1}{C{.13\textwidth}}{$l_{1.5}$, $\epsilon=50$}
    & \multicolumn{1}{C{.13\textwidth}}{Diff. map}
    &\multicolumn{1}{C{.13\textwidth}}{$l_2$, $\epsilon=12$}
    & \multicolumn{1}{C{.13\textwidth}}{Diff. map}\\
    \hline
    \rotatebox[origin=c]{90}{ILSVRC2012} & \begin{subfigure}{0.13\textwidth}\centering
     \caption*{tiger beetle:$0.85$}
     \includegraphics[width=1\textwidth]{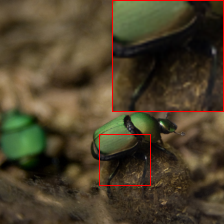}
     \end{subfigure}
     &\begin{subfigure}{0.13\textwidth}\centering
     \caption*{ 
     dung beetle:$0.88$}
     \includegraphics[width=1\textwidth]{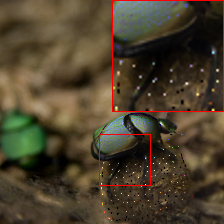}
     \end{subfigure}
      &\begin{subfigure}{0.13\textwidth}\centering
      \caption*{
     \newline}
         \includegraphics[width=1\textwidth]{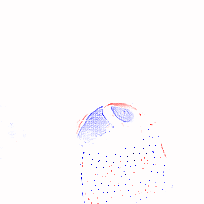}
         \end{subfigure}
     &\begin{subfigure}{0.13\textwidth}\centering
     \caption*{
     dung beetle:$0.99$}
     \includegraphics[width=1\textwidth]{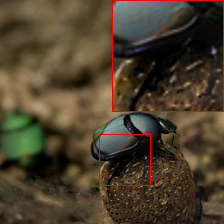}
     \end{subfigure}
     
      &\begin{subfigure}{0.13\textwidth}\centering
      \caption*{
     \newline}
         \includegraphics[width=1\textwidth]{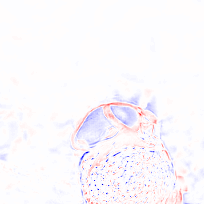}
         \end{subfigure}
         
     &\begin{subfigure}{0.13\textwidth}\centering
     \caption*{
     dung beetle:$1.00$}
     \includegraphics[width=1\textwidth]{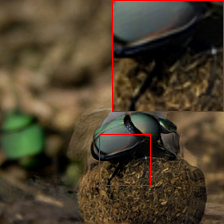}
     \end{subfigure}
      &\begin{subfigure}{0.13\textwidth}\centering
      \caption*{
     \newline}
         \includegraphics[width=1\textwidth]{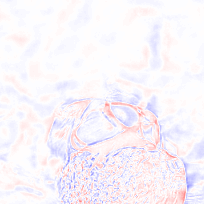}
         \end{subfigure}\\

     \end{tabular}

     \caption{\label{tab:afw-varying-eps-short}
     \textbf{ImageNet: $l_p$-VCEs for $p \in \{1, 1.5, 2\}$.} $l_p$-VCEs  into  correct class for a misclassified Image of class ``dung beetle''.
     for the multiple-norm adversarially robust model Madry\cite{robustness_short}+FT. $l_1$-VCEs are too sparse and introduce artefacts and $l_2$-VCEs change the background. Our $l_{1.5}$-VCEs are sparse and object-related (see difference maps right from the respective $l_p$-VCEs).
     }
     \end{figure*}

    \begin{figure}[htb!]
     \centering
     \small
     \begin{tabular}{c|ccc|ccc|}
     \hline
     \multicolumn{1}{c|}{Original}&  \multicolumn{3}{c|}{$l_{1.5}$-VCE} &    \multicolumn{3}{c|}{$l_{1.5}$-VCE}\\
     \multicolumn{1}{c|}{} & \multicolumn{1}{C{.135\textwidth}}{
     $\epsilon=50$} & 
     \multicolumn{1}{C{.135\textwidth}}{$\epsilon=75$} &
     \multicolumn{1}{C{.135\textwidth}|}{$\epsilon=100$}
     & \multicolumn{1}{C{.135\textwidth}}{
     $\epsilon=50$} & 
     \multicolumn{1}{C{.135\textwidth}}{$\epsilon=75$} &
     \multicolumn{1}{C{.135\textwidth}|}{$\epsilon=100$}\\
     \hline
       
     \multirow{2}{0.135\textwidth}{\begin{subfigure}{0.135\textwidth}\centering
     \caption*{{\tiny Gila monster: 0.12}}
     \includegraphics[width=1\textwidth]{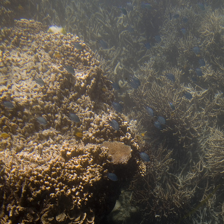}
     \end{subfigure}} &
     
     \begin{subfigure}{0.135\textwidth}\centering
     \caption*{{\tiny $\rightarrow$reef: 0.92}}
     \includegraphics[width=1\textwidth]{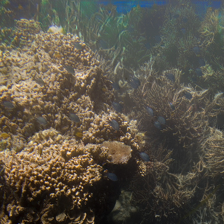}
     \end{subfigure} &

     \begin{subfigure}{0.135\textwidth}\centering
     \caption*{{\tiny $\rightarrow$reef: 0.98}}
     \includegraphics[width=1\textwidth]{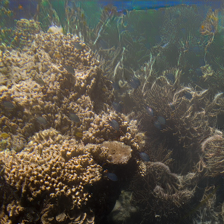}
     \end{subfigure} &

     \begin{subfigure}{0.135\textwidth}\centering
     \caption*{{\tiny $\rightarrow$reef: 1.00}}
     \includegraphics[width=1\textwidth]{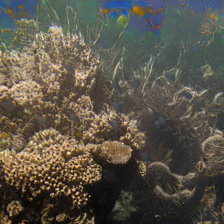}
     \end{subfigure} 
     
      &
     
     \begin{subfigure}{0.135\textwidth}\centering
     \caption*{{\tiny $\rightarrow$cliff: 0.92}}
     \includegraphics[width=1\textwidth]{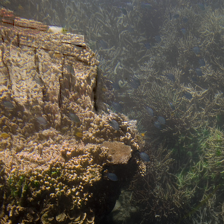}
     \end{subfigure} &

     \begin{subfigure}{0.135\textwidth}\centering
     \caption*{{\tiny $\rightarrow$cliff: 0.96}}
     \includegraphics[width=1\textwidth]{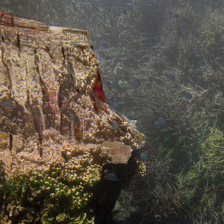}
     \end{subfigure} &

     \begin{subfigure}{0.135\textwidth}\centering
     \caption*{{\tiny $\rightarrow$cliff: 0.99}}
     \includegraphics[width=1\textwidth]{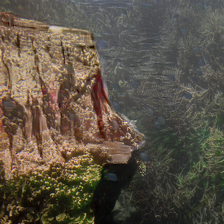}
     \end{subfigure} \\

     &     
     
     \begin{subfigure}{0.135\textwidth}\centering
     \caption*{{\tiny $\rightarrow$valley: 0.79}}
     \includegraphics[width=1\textwidth]{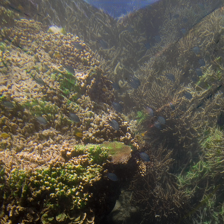}
     \end{subfigure} &

     \begin{subfigure}{0.135\textwidth}\centering
     \caption*{{\tiny $\rightarrow$valley: 0.94}}
     \includegraphics[width=1\textwidth]{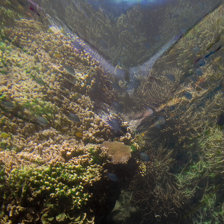}
     \end{subfigure} &

     \begin{subfigure}{0.135\textwidth}\centering
     \caption*{{\tiny$\rightarrow$valley: 0.98}}
     \includegraphics[width=1\textwidth]{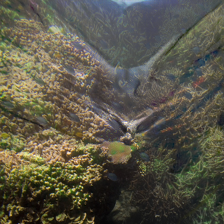}
     \end{subfigure}

     &     
     
     \begin{subfigure}{0.135\textwidth}\centering
     \caption*{{\tiny $\rightarrow$volcano: 0.91}}
     \includegraphics[width=1\textwidth]{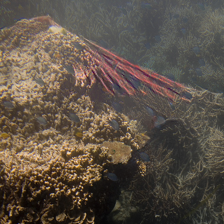}
     \end{subfigure} &

     \begin{subfigure}{0.135\textwidth}\centering
     \caption*{{\tiny $\rightarrow$volcano: 1.00}}
     \includegraphics[width=1\textwidth]{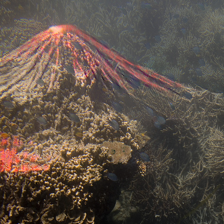}
     \end{subfigure} &

     \begin{subfigure}{0.135\textwidth}\centering
     \caption*{{\tiny $\rightarrow$volcano: 1.00}}
     \includegraphics[width=1\textwidth]{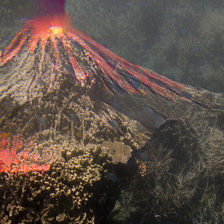}
     \end{subfigure} \\

     \hline
    
    
    \multirow{2}{*}{
     \begin{subfigure}{0.135\textwidth}\centering
     \caption*{ \tiny fig: 0.27}
     \includegraphics[width=1\textwidth]{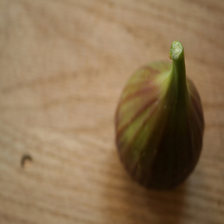}
     \end{subfigure}} &



     
     \begin{subfigure}{0.135\textwidth}\centering
     \caption*{\tiny$\rightarrow$ strawberry: 0.97}
     \includegraphics[width=1\textwidth]{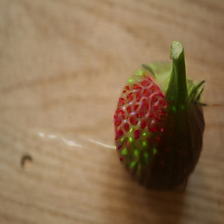}
     \end{subfigure} &

     \begin{subfigure}{0.135\textwidth}\centering
     \caption*{ \tiny$\rightarrow$ strawberry: 1.00}
     \includegraphics[width=1\textwidth]{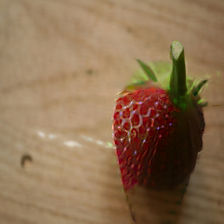}
     \end{subfigure} &

     \begin{subfigure}{0.135\textwidth}\centering
     \caption*{ \tiny$\rightarrow$ strawberry: 1.00}
     \includegraphics[width=1\textwidth]{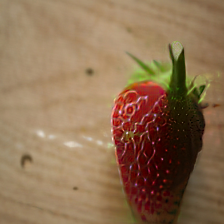}
     \end{subfigure}
     
     &
     
     \begin{subfigure}{0.135\textwidth}\centering
     \caption*{ \tiny$\rightarrow$ pineapple: 0.99}
     \includegraphics[width=1\textwidth]{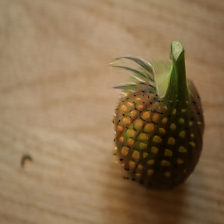}
     \end{subfigure} &

     \begin{subfigure}{0.135\textwidth}\centering
     \caption*{ \tiny$\rightarrow$ pineapple: 1.00}
     \includegraphics[width=1\textwidth]{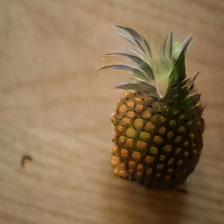}
     \end{subfigure} &

     \begin{subfigure}{0.135\textwidth}\centering
     \caption*{ \tiny$\rightarrow$ pineapple: 1.00}
     \includegraphics[width=1\textwidth]{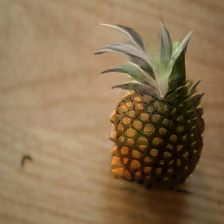}
     \end{subfigure}\\&
     
     \begin{subfigure}{0.135\textwidth}\centering
     \caption*{ \tiny$\rightarrow$ custard apple: 0.95}
     \includegraphics[width=1\textwidth]{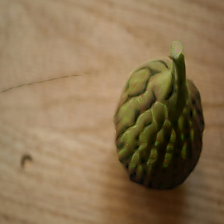}
     \end{subfigure} &

     \begin{subfigure}{0.135\textwidth}\centering
     \caption*{ \tiny$\rightarrow$ custard apple: 0.99}
     \includegraphics[width=1\textwidth]{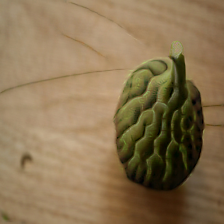}
     \end{subfigure} &

     \begin{subfigure}{0.135\textwidth}\centering
     \caption*{ \tiny$\rightarrow$ custard apple: 1.00}
     \includegraphics[width=1\textwidth]{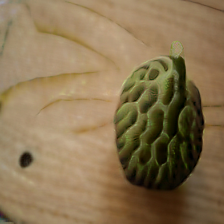}
     \end{subfigure}&
     
     \begin{subfigure}{0.135\textwidth}\centering
     \caption*{ \tiny $\rightarrow$ pomegran.: 0.76}
     \includegraphics[width=1\textwidth]{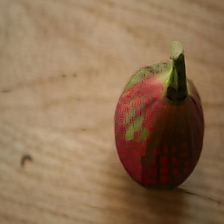}
     \end{subfigure} &

     \begin{subfigure}{0.135\textwidth}\centering
     \caption*{ \tiny $\rightarrow$ pomegran.: 0.97}
     \includegraphics[width=1\textwidth]{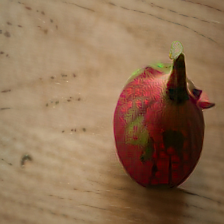}
     \end{subfigure} &

     \begin{subfigure}{0.135\textwidth}\centering
     \caption*{ \tiny $\rightarrow$ pomegran.: 0.99}
     \includegraphics[width=1\textwidth]{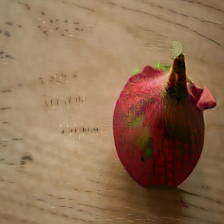}
     \end{subfigure} \\

     \hline
     \end{tabular}
      \caption{\label{tab:imagent_valley} \textbf{ImageNet: $l_{1.5}$-VCEs.} (First two rows) $l_{1.5}$-VCEs for Madry\cite{robustness_short}+FT with varying radii for a misclassified image of class ``coral reef'' for the target classes: ``coral reef'', ``cliff'', ``valley'' and ``volcano'' (same wordnet category ``geological formation'').
      (Second two rows) $l_{1.5}$-VCEs for Madry\cite{robustness_short}+FT with varying radii for an image of class ``fig'' for the target classes: ``strawberry'', ``pineapple'', ``custard apple'', and `pomegranate` (same wordnet category ``edible fruit'').
      The changes are sparse and subtle for radius 50 and get more pronounced for larger budgets. 
    }
     \end{figure}

    \subsection{Sparse VCEs via the $l_{1.5}$-metric}\label{sec:threatmodel}
    The perturbation budget  of VCEs in \cref{eq:VCE-RATIO}, in particular the chosen distance metric, is crucial for the generation of realistic VCEs. It might seem natural to use for $l_2$-adversarially robust models
    also the $l_2$-metric for the perturbation budget of the VCEs. However, as we show in  \cref{fig:motivating_example} and  \cref{tab:afw-varying-eps-short}, the problem of the $l_2$-budget is that one typically gets non-sparse changes over the full image which are not centered on the object. Aiming at sparse VCEs it seems like the $l_1$-metric might be well-suited as it is known to lead to sparse changes. However, as one can see in \cref{tab:afw-varying-eps-short}, the changes are in fact extremely sparse and often show color artefacts: \eg for the dung beetle, single pixels are changed to non-natural extreme colors. As a compromise between $l_1$ (too sparse) and $l_2$ (non-sparse), we propose to use the $l_{1.5}$-metric for the perturbation model in \cref{eq:VCE-RATIO}. In \cref{fig:motivating_example} and \cref{tab:afw-varying-eps-short}, for 
    ImageNet 
    the changes of $l_{1.5}$-VCEs are sparse and localized on the object. For the generation of the FID scores for $l_{1.5}$-VCEs, we used 
    $\epsilon=50$ for ImageNet. Apart from the better FID-scores of $l_{1.5}$-VCEs we quantify in Tab.~\ref{fig:imagenet_s} in \cref{app:further-evaluation} for ImageNet the concentration of the changes on the actual object using the 
    pixel-wise segmentations of ImageNet test images in \cite{gao2021luss}.

    We found the chosen radii to work well for most images. However, for visualizing the VCEs to a user, the best option is to let the user see how the changes evolve as one changes the radius in an interactive fashion. Rather subtle changes with a small budget are already sufficient for some images, whereas for other images larger budgets are necessary due to a significant change of color or shape. As such an interactive process cannot be shown, we provide panels with different radii of the perturbation model in \cref{tab:imagent_valley} and the Appendix.

    {\textbf{FID evaluation:}}
    For a quantitative evaluation, we compute FID scores for our ImageNet models in \cref{tab:l1.5_FIDs}, where we use for the ImageNet validation set the second predicted class as target (ID) and for out-of-distribution images (OD) from ImageNet-O/ImageNet-A we generate the VCE for the predicted class. The FID scores indicate that $l_{1.5}$-VCEs have higher \textbf{realism} and \textbf{sparsity} than $l_1$- and $l_2$-VCEs, on both in- and out-of-distribution images. 
    
     \begin{figure*}[t]
     
     \centering
     \begin{tabular}{c|c|c|c}
                   \hline 
              \multicolumn{1}{c}{Original}&\multicolumn{1}{C{.27\textwidth}}{APGD, $l_{1}$}&\multicolumn{1}{C{.27\textwidth}}{AFW, $l_{1.5}$}&\multicolumn{1}{C{.27\textwidth}}{APGD, $l_{2}$}\\
                   \hline

     \begin{subfigure}{0.14\textwidth}\centering
     
     \caption*{{\tiny hummingb.:$0.35$}}
     \includegraphics[width=.96\textwidth]{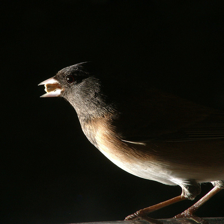}
     \end{subfigure}&
     \begin{subfigure}{0.28\textwidth}\centering

     \caption*{{\tiny r.:$0.61$, m.:$0.72$, s.:$0.61$}}
     \includegraphics[width=.47\textwidth]{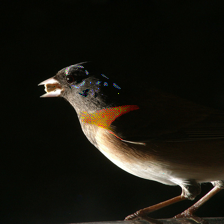}
     \includegraphics[width=.47\textwidth]{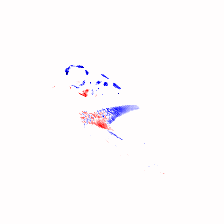}
     
     \end{subfigure}&\begin{subfigure}{0.28\textwidth}\centering
     
     \caption*{{\tiny r.:$0.89$, m.:$0.89$, s.:$0.72$}}
     \includegraphics[width=.47\textwidth,cfbox=blue 1pt 0pt]{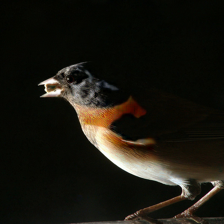}
     \includegraphics[width=.47\textwidth]{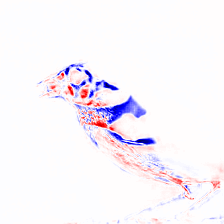}
     
     \end{subfigure} 
     &\begin{subfigure}{0.28\textwidth}\centering

     \caption*{{\tiny r.:$0.78$, m.:$0.78$, s.:$0.61$}}
     \includegraphics[width=.47\textwidth]{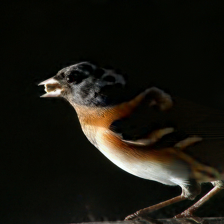}
     \includegraphics[width=.47\textwidth]{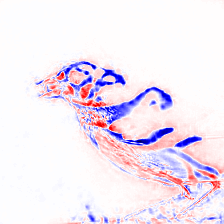}
     
     \end{subfigure}\\

     \hline

          \begin{subfigure}{0.14\textwidth}\centering
     
     \caption*{{\tiny paddle:$0.11$}}
     \includegraphics[width=.96\textwidth]{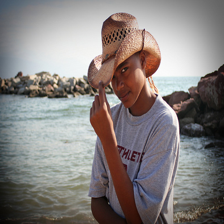}
     \end{subfigure}&
     \begin{subfigure}{0.28\textwidth}\centering

     \caption*{{\tiny r.:$0.06$, m.:$0.17$, s.:$0.06$}}
     \includegraphics[width=.47\textwidth,cfbox=red 1pt 0pt]{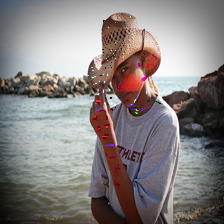}
     \includegraphics[width=.47\textwidth]{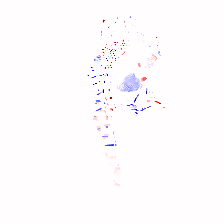}
     
     \end{subfigure}&\begin{subfigure}{0.28\textwidth}\centering
     
     \caption*{{\tiny r.:$0.28$, m.:$0.11$, s.:$0.44$}}
     \includegraphics[width=.47\textwidth]{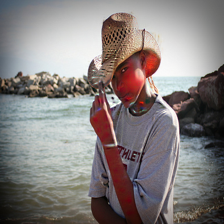}
     \includegraphics[width=.47\textwidth]{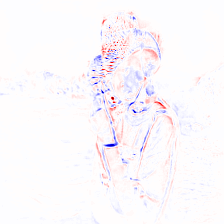}
     
     \end{subfigure} 
     &\begin{subfigure}{0.28\textwidth}\centering

     \caption*{{\tiny r.:$0.39$, m.:$0.11$, s.:$0.5$}}
     \includegraphics[width=.47\textwidth]{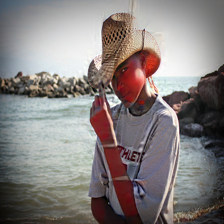}
     \includegraphics[width=.47\textwidth]{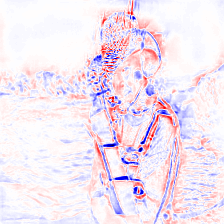}
    \end{subfigure}

      \end{tabular}
    
     \caption{\label{fig:main_user_study} \textbf{Best (blue) and worst (red) rated images from the user study.} $l_p$-VCEs for $p \in \{1,1.5,2\}$ for the change ``hummingbird $\rightarrow$ brambling'' (top row) and ``paddle $\rightarrow$ bearskin'' (bottom row) for Madry\cite{robustness_short}+FT, with \textbf{realism (r)}, \textbf{meaningful (m)}, \textbf{subtle (s)} fractions from the user study (r:0.61 means that 61\% of the users considered this image to be realistic). }

     \end{figure*}

    {\textbf{User study:}}
     We perform a user study (18 participants) to compare $l_{1.5}$-, $l_1$-, and $l_2$-VCEs (the Madry \cite{robustness_short}+FT model on ImageNet is used). 
    For each target image (94 in total), we show  $l_p$-VCEs for $p \in \{1,1.5,2\}$, to the users, who can choose which ones satisfy the following properties (none or multiple 
    answers are allowed): i) \textbf{realism}, ii) ``meaningful features in the target class are introduced'' (\textbf{meaningful}), iii) ``subtle, yet understandable changes are introduced'' (\textbf{subtle}). 
    The percentages for $l_1$-, $l_{1.5}$-, and $l_2$-VCEs are: \textbf{realism} - 23.5\%, \textbf{38.2\%}, 33.8\%; \textbf{meaningful} - 37.5\%, 63.1\%, \textbf{64.0\%}; \textbf{subtle} -  34.7\%, \textbf{49.1\%}, 41.5\%. While the difference of $l_{1.5}$-VCEs compared to $l_2$-VCEs is small for meaningfulness, $l_{1.5}$-VCEs are considered more subtle and realistic. In \cref{fig:main_user_study}, we show the best and worst rated images from the user study. Note that for the worst one, the changes into the target class ''bearskin'' are not achievable in the given budget and thus all methods fail to produce meaningful images.\\ 
    {\textbf{Details about the user study:}}
    Participants are researchers in machine learning (volunteers) not working on VCEs themselves and neither being exposed to the generated images or compared methods before. The p-values, using the two-sample two-sided binomial test, for the best and next best methods are: $0.02$ for \textbf{realism} of $l_{1.5}$ vs $l_{2}$, 
    $0.6$ for \textbf{meaningful} of $l_2$ vs $l_{1.5}$, and $5.3\cdot 10^{-5}$ for \textbf{subtle} of $l_{1.5}$ vs $l_{2}$, that is $l_{1.5}$ outperforms statistically significantly $l_2$ in realism/subtle (significance level $0.05$).

    \section{Auto-Frank-Wolfe for $l_p$-VCEs}
    \label{sec:afw}
    \begin{figure}[t]
      \centering{%
       \includegraphics[width=0.5\textwidth]{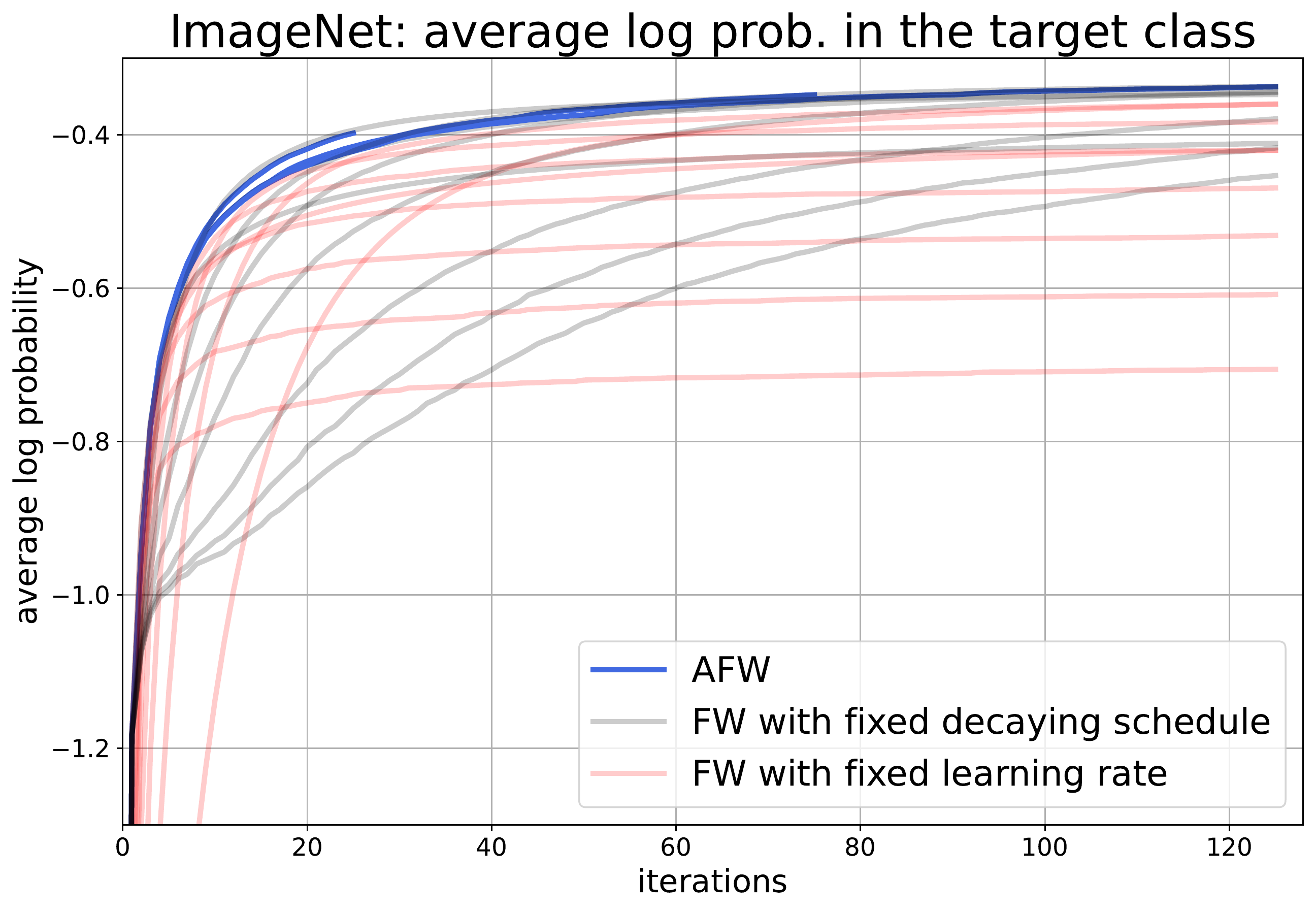}%
       \includegraphics[width=0.5\textwidth]{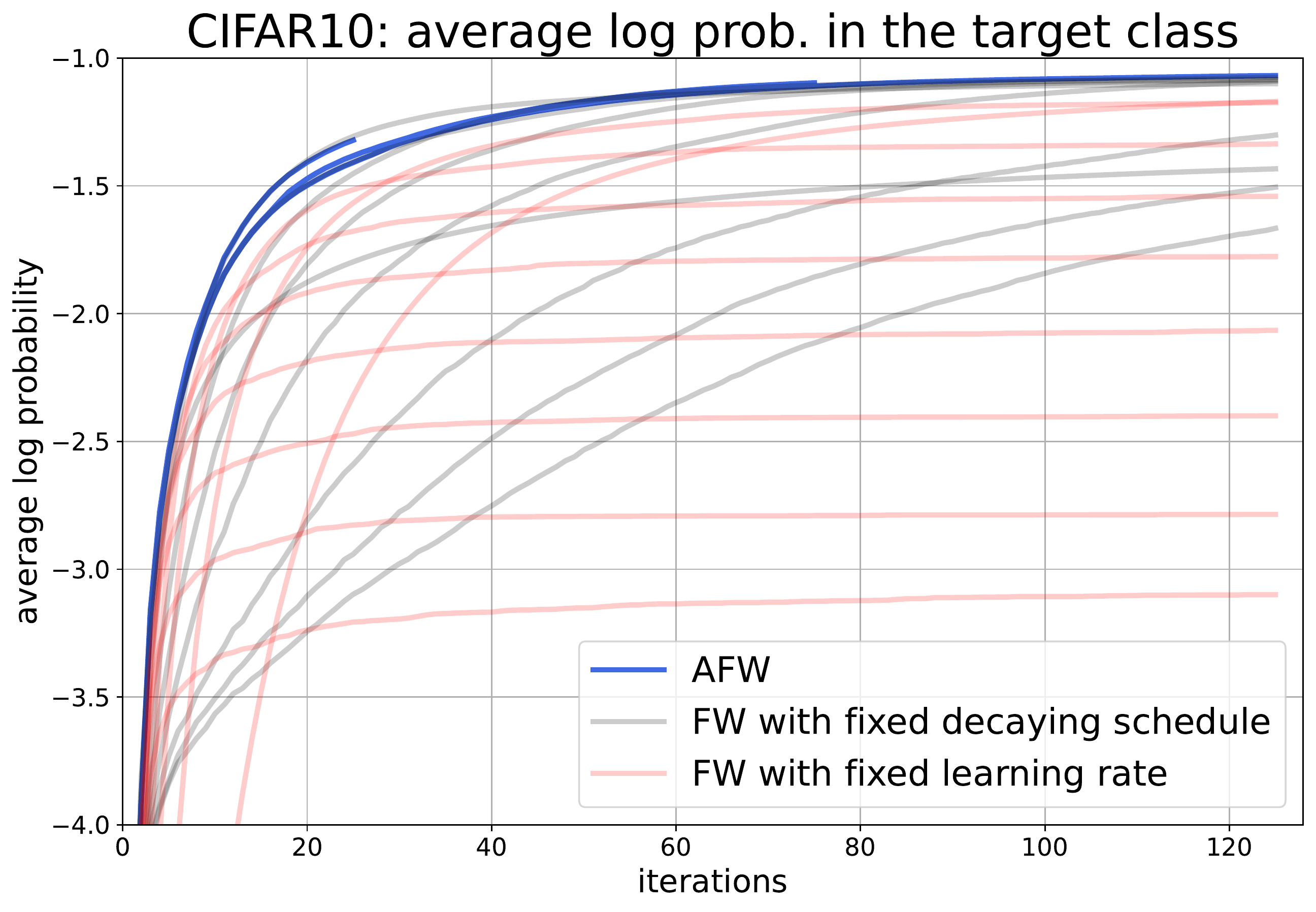}%
       }
      \caption{\label{fig:apgd_vs_afw_probs_fig} \textbf{Comparison of our adaptive Frank-Wolfe scheme (AFW) vs non-adaptive FW.} We plot avg. log probability of the $l_{1.5}$-VCEs (ImageNet: $\epsilon=50$, CIFAR10: $\epsilon=6$) with the second predicted class as target for ImageNet (left) and CIFAR10 (right). Our novel AFW (blue curve) is adaptive to the budget, without any need for additional hyperparameter tuning. We show the result of AFW for a budget of 25, 75 and 125. We observe that it matches the best or is better than FW with fixed learning rate (red) or various decaying stepsize schemes (green) for the given number of iterations for both datasets.
    }
      \end{figure}
    For deep models, the optimization problem for $l_p$-VCEs 
    \begin{equation} \label{eq:maxlikelihood_constrained}
          \maxop_{x \in B_p(x_0,\epsilon) \cap [0, 1]^d} \log \hat{p}(y | x),
    \end{equation}
    is non-convex and related to targeted adversarial attacks, for which AutoPGD (APGD) \cite{croce2020reliable} has been shown to be very effective. APGD requires projections onto $l_p$-balls which are available either in closed form for $l_2$ and $l_\infty$
    or can be computed efficiently for $l_1$. However, for $p\notin \{1,2,\infty\}$, there is no such projection available and one cannot use APGD. Thus, in order to generate $l_p$-VCEs for $p>1$, we propose
    an adaptive version of the Frank-Wolfe (FW) algorithm \cite{frank-wolfe,pmlr-v28-jaggi13}, which we call Auto-Frank-Wolfe (AFW). FW has the advantage that it is projection-free and thus allows to use more complex constraint sets. In particular, we can use arbitrary $l_p$ norm balls for $p > 1$.

    \textbf{Auto-Frank-Wolfe:} At each iteration $k$, FW maximizes the first-order Taylor expansion at the iterate $x^k$ of 
    the objective in the feasible set 
    , i.e.
    \begin{align} s^k = \argmax_{s \in B_p(x_0,\epsilon) \cap [0, 1]^d} \inner{s, \nabla_{x^k} \log \hat{p}(y | x^k)}, \label{eq:lmo}
    \end{align}
    and the next iterate is the convex combination
    \begin{equation}\label{eq:fw_conv_comb}
         x^{k+1} = (1-\gamma^k)x^k + \gamma^k s^k.
    \end{equation}
    The choice of the learning rate $\gamma^k \in (0,1)$ is crucial for the success of the algorithm: in the context of adversarial attacks, \cite{chen2019a} use a fixed value $\gamma_0$ for every $k$, while \cite{JMLR:v18:14-348,tsiligkaridis2020understanding} decrease it as $\frac{\gamma_0}{\gamma_0 + k}$. In both cases the schedule is agnostic of the total budget of iterations, and $\gamma_0$ needs to be tuned. Thus, we propose to use an adaptive scheme for choosing $\gamma^k$ at each iteration as $\gamma^k=\frac{M}{2+\sqrt{k}}$ where $M\leq 2$ is adapted during the optimization. This yields our AFW attack which automatically adapts to different budgets (details of AFW in \cref{app:afw_details}).
    
    \textbf{Considering box-constraints}: Prior FW-based attacks \cite{chen2019a,tsiligkaridis2020understanding} do not consider the image domain constraints $[0, 1]^d$ but rather solve \cref{eq:lmo} for $l_p$-ball constraints only (which has a closed form solution) and clip it to $[0, 1]^d$. This is suboptimal, especially when $p$ is close to 1, see 
    \cite{croce2021mind}. The following proposition shows that it is possible to solve \cref{eq:lmo} efficiently  in the intersection $B_p(x_0,\epsilon) \cap [0, 1]^d$ for $p > 1$ (proof in \cref{app:afw_proof}, $p=1$ is more simple, see \cite{croce2021mind},).
 
    \begin{proposition}\label{thm:fw_intersection}
    Let $w \in \R^d$, $x \in [0,1]^d$, $\epsilon > 0$ and $p > 1$. 
    The solution $\delta^*$ of the optimization problem
    \begin{equation}
        \argmax_{\delta \in \R^d
        } \inner{w, \delta} \quad \textrm{s.th.} \; \norm{\delta}_p \leq \epsilon, \; x + \delta \in [0, 1]^d
    \end{equation}
    is given, with the convention $\sign{0} = 0$, by
    $$\delta^*_i = \min\left\lbrace\gamma_i, \left(\frac{|w_i|}{p\mu^*}\right)^\frac{1}{p - 1}\right\rbrace \sign{w_i}, \quad i=1, \ldots, d,$$
    where $\gamma_i = \max\{-x_i \sign{w_i}, (1-x_i)\sign{w_i}\}$ and $\mu^* > 0$ can be computed in $O(d\log d)$ time.
    \end{proposition}
    
    \textbf{Experiments:} To evaluate the effectiveness of AFW, we compare its performance when optimizing \cref{eq:maxlikelihood_constrained} in the $l_{1.5}$-ball of radius $50$. We use different budgets of $25,75,125$ iterations (75 is used for generating all VCEs), and test a variety of fixed parameters for the FW attacks of \cite{chen2019a} ($\gamma_0 \in\{0.1, 0.2, ..., 0.9\}$ as constant stepsize and $\gamma_0 \in \{1, 5, ..., 25, 50, 75, 100\}$ and $M \in \{2,...\}$ for stepsize decaying with $k$). \Cref{fig:apgd_vs_afw_probs_fig} shows that AFW achieves the maximal objective (log probability of the target class) for 75 and is second best for the budget of 25 and 125 iterations. Thus AFW adapts to the given budget and outperforms FW with fixed stepsize schemes. In \cref{app:afw_additional}, we provide additional experiments.

    \begin{figure}[hbt!]
     \centering
     \small
     \begin{tabular}{c|c|c|c}
     \hline
     \multicolumn{1}{c|}{Orig.} & 
     \multicolumn{1}{c|}{{\footnotesize $l_{1.5}$-VCE, $\epsilon = 50$}
     } &
     \multicolumn{1}{c|}{Watermark}&
     \multicolumn{1}{c}{Train set}\\
     \hline

     \begin{subfigure}{0.22\columnwidth}\centering
     \caption*{bell pepper: 0.95}
     \includegraphics[width=1\columnwidth]{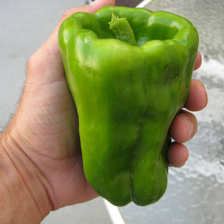}
     \end{subfigure} &
     
     \begin{subfigure}{0.22\columnwidth}\centering
     \caption*{$\rightarrow$GS: 0.94}
     \includegraphics[width=1\columnwidth]{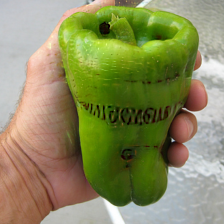}
     \end{subfigure} &

     \begin{subfigure}{0.22\columnwidth}\centering
     \caption*{$\rightarrow$GS: 0.65}
     \includegraphics[width=1\columnwidth]{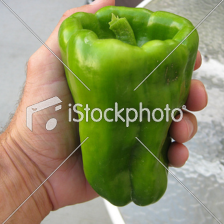}
     \end{subfigure} &

     \begin{subfigure}{0.22\columnwidth}\centering
     \caption*{GS}
     \includegraphics[width=1\columnwidth]{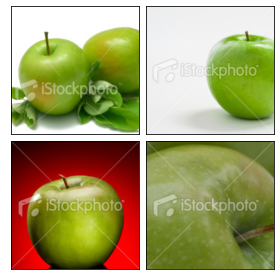}
     \end{subfigure} 
     \end{tabular}
      \caption{\label{fig:spurious_features} \textbf{Spurious feature: watermark.} The $l_{1.5}$-VCE with target class ``granny smith'' (GS) for Madry\cite{robustness_short}+FT shows that the model has associated a spurious ``text'' feature with this class. This is likely due to ``iStockphoto'' watermarked images in its training set (right). Adding the watermark changes the decision of the classifier to GS.
    }
     \end{figure}

    \section{Finding spurious features with $l_{1.5}$-VCEs} \label{sec:debugging}

    Neural networks are excellent feature extractors and very good at finding correlations in the data. 
    This makes them susceptible to learn spurious features which are not task-related \cite{hohman2020summit,brendel2018bagnets,carter2019exploring,goh2021multimodal,zech2018confounding}, to the extent that the actual object is not recognized, \eg a cow on a beach is not recognized without the appearance of the spurious feature ``pasture/grass'' \cite{beery2018recognition,singla2021understanding}. We show how VCEs can be used for finding such spurious features. While an automatic pipeline is beyond the scope of this paper, we believe that this can be done with minimal human supervision. 
    
    \textbf{Failure A, Watermark text as spurious feature for ``granny smith'':} We detected this failure  when creating VCEs for the target class ``granny smith''. We consistently observed text-like features on the generated $l_{1.5}$-VCEs which are obviously not related to this class. In 
     \cref{fig:spurious_features} we illustrate the $l_{1.5}$-VCE for an image from the class ``bell pepper''. More examples are in \cref{app:watermarks}.
    \changed{Since almost none of the ``granny smith'' training images contains text,} we came up with the hypothesis that the reason is a high percentage of watermarked images in the training set. Manual inspection showed that 90 out of 1300 training images contain a watermark, out of which 53 contain the one of ``iStockphoto'' (see the rightmost image in \cref{fig:spurious_features}). While watermarked images appear in several classes in ImageNet, this significant fraction of one type of watermark in the training set seems to only be present in the class ``granny smith''. We tested this hypothesis by simulating the watermark of ``iStockphoto'' on the test set, for an example see the second image from the right in     \cref{fig:spurious_features} and \cref{app:watermarks} for more details. In \cref{tab:granny-smith-analysis} we show that adding the ``iStockphoto''-watermark has a negative influence on top-1 accuracy not only for the adv. robust model Madry\cite{robustness_short}+FT but also other non-robust models. The reason is that this particular watermark contains features (lines) which are characteristic for classes such as bow, safety pin, reel. However, even though the ``iStockphoto''-watermark contains no characteristic features of ``granny smith'', adding the watermark leads to significantly worse precision of the granny smith class (basically an increase in false positives while false negatives stay the same). Interestingly, even an accurate model such as NS-B7 \cite{xie2020selftraining} shows this effect although trained using the much larger non-public JFT-300M dataset, suggesting that JFT-300M contains these artefacts as well.
    
    \begin{table}[tb]
        \centering
        \small
        \caption{\label{tab:granny-smith-analysis} We show top-1 accuracy for the test set, and precision/recall for the class ``granny smith'' (GS) on the test set before and after adding the ``iStockphoto''-watermark. Note that the watermark has a quite significant impact on accuracy and in particular on the precision of GS (more false positives), which confirms our hypothesis on the bias induced by the GS training set.}
        \begin{tabular}{c|c|c|c|c|c|c}
         \hline
         & \multicolumn{3}{c|}{Original} & \multicolumn{3}{c}{w. Watermark}\\
         \hline
         & 
        \multicolumn{1}{C{.1\textwidth}|}{Top-1} & \multicolumn{2}{c|}{GS vs Rest}  &  \multicolumn{1}{C{.1\textwidth}|}{Top-1} & \multicolumn{2}{c}{GS vs Rest}  \\
        \multicolumn{1}{C{.14\textwidth}|}{Model} & \multicolumn{1}{C{.1\textwidth}|}{Acc.}
        & \multicolumn{1}{C{.06\textwidth}}{Prec.} 
        & \multicolumn{1}{C{.06\textwidth}|}{Rec.}
        & \multicolumn{1}{C{.1\textwidth}|}{Acc.}
        & \multicolumn{1}{C{.06\textwidth}}{Prec.} 
        & \multicolumn{1}{C{.06\textwidth}}{Rec.} \\
        \hline\hline
        
        Madry\cite{robustness_short}+FT     & 57.5 & 61.1 & 73.3  & 50.4  & 43.8 & 70.0\\
        
        ResNet50\cite{robustness_short}     & 76.0 & 90.3 & 93.3  & 62.3 & 53.2 & 83.3 \\

        NS-Eff. B7 \cite{xie2020selftraining} & 86.6 & 90.3 & 93.3 & 84.1 & 68.3 & 93.3 \\

        \hline
        \end{tabular}

    \end{table}

    \begin{figure}[t]
     \centering
     \small
     \begin{tabular}{c|cc|c}
     \hline
     \multicolumn{1}{c|}{Orig.} & 
     \multicolumn{2}{c|}{$l_{1.5}$-VCE, $\epsilon = 100$} &
     \multicolumn{1}{c}{Train set}\\
     \hline

     \begin{subfigure}{0.22\columnwidth}\centering
     \caption*{coral reef: 0.58}
     \includegraphics[width=1\columnwidth]{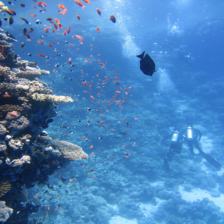}
     \end{subfigure} &
     
     \begin{subfigure}{0.22\columnwidth}\centering
     \caption*{$\rightarrow$t. shark: 0.96}
     \includegraphics[width=1\columnwidth]{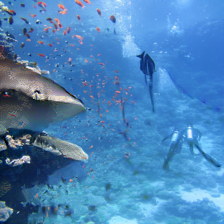}
     \end{subfigure} &

     \begin{subfigure}{0.22\columnwidth}\centering
     \caption*{$\rightarrow$w. shark: 0.99}
     \includegraphics[width=1\columnwidth]{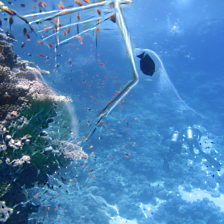}
     \end{subfigure} &

     \begin{subfigure}{0.22\columnwidth}\centering
     \caption*{white shark}
     \includegraphics[width=1\columnwidth]{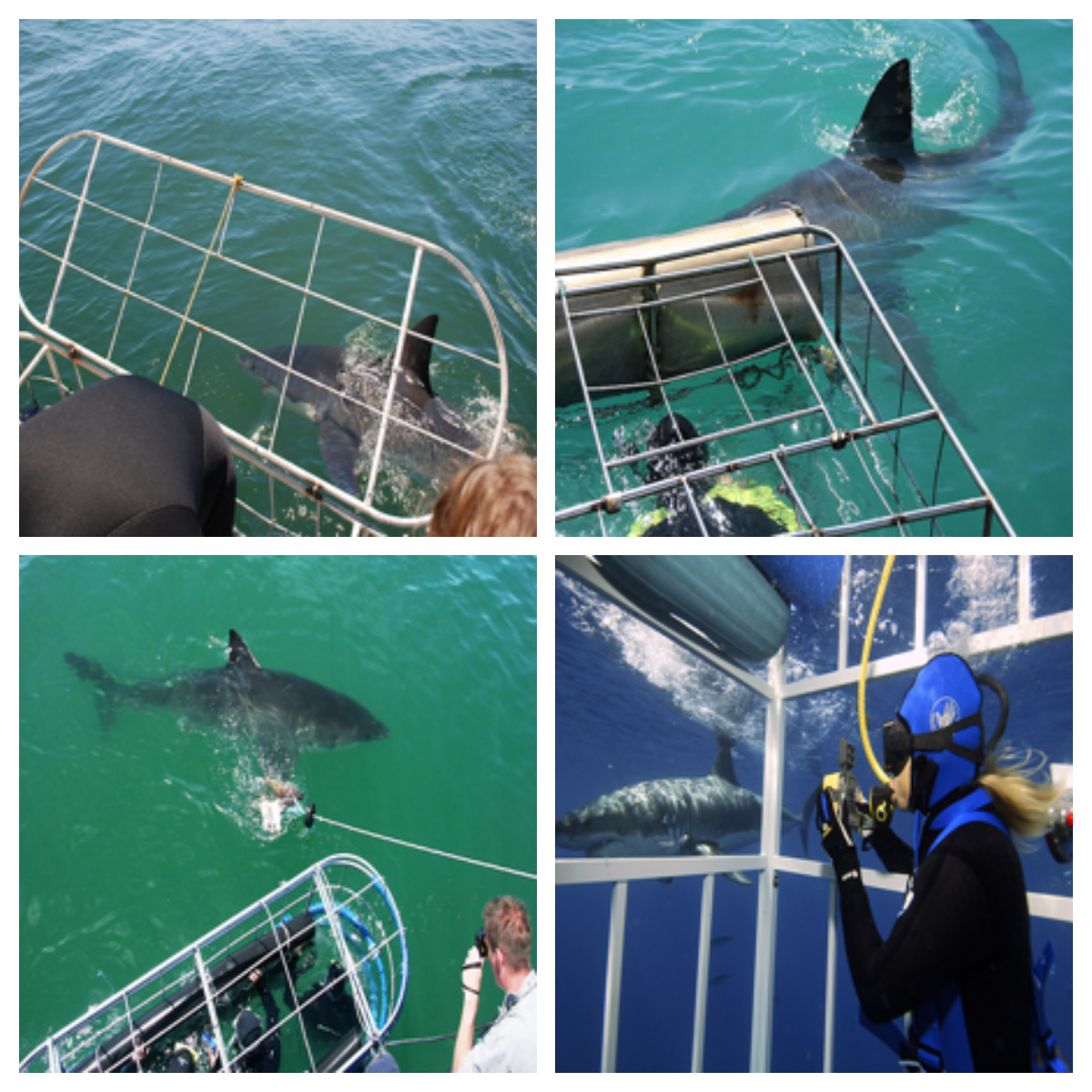}
     \end{subfigure} 
     
     \end{tabular}
      \caption{\label{tab:imagent_shark} \textbf{Spurious feature: cage.} The $l_{1.5}$-VCE for an image from class coral reef with target ``tiger shark'' shows a tiger shark, but with target ``white shark'' grid-like structures as spurious feature. The training set of white shark (right) contains many images with cages.
    }
    
     \end{figure}

    \textbf{Failure B, Cages as spurious feature for ``white shark'':}
    The next failure was detected using $l_{1.5}$-VCEs for the shark classes where very frequently grid-like structures appear - but only for VCEs with  target class ``white shark'' not for ``tiger shark'' or ``hammerhead''. A typical situation is shown in \cref{tab:imagent_shark}, where the original image is from class ``coral reef''. The VCE for ``tiger shark'' shows a shark coming from the left whereas the VCE for ``white shark'' shows just a grid-like structure, see \cref{app:sharks} for more such VCEs. An inspection of the ``white shark'' training set reveals that many of the images contain parts of cages protecting the photographing diver, see the rightmost image in \cref{tab:imagent_shark} for an illustration. The model has picked up on this feature as a fairly dominant one for  white sharks, which is clearly spurious and undesirable. Interestingly, the VCEs allow us to find such artefacts even without images of white-sharks which is an advantage over saliency or feature attribution methods.

    \section{Discussion and Limitations}
    We have shown that our $l_{1.5}$-VCEs are sparse and do subtle changes located on the object of interest even for ImageNet resolution. We have shown that $l_{1.5}$-VCEs are a useful debugging tool for detecting spurious features which the classifier has picked up. 
    Not all VCEs are meaningful which can have different reasons: i) the perturbation budget is too small to be able to change to more distinct classes, ii) the model has not picked up the right features, or iii) VCEs show spurious features, as discussed in \cref{sec:debugging}. However, these ``limitations'' just reflect that our classifiers are not yet perfect and are not a failure of VCEs. In the future it will be interesting to generate an automatic pipeline for the detection of spurious features with minimal human supervision.

    \section*{Acknowledgement}
    M.H., P.B., and V.B. acknowledge support by the the DFG Excellence Cluster Machine Learning - New Perspectives for Science, EXC 2064/1, Project number 390727645. 
   \clearpage
    
     {\small
    \bibliographystyle{splncs04}

    }    
    \clearpage
    
    \appendix
    \section*{Overview of Appendix}
    In the following we present several additional results and experimental details.
    
    \begin{itemize}
    \item In \cref{sec:app-robustness} we present more examples of $l_{1.5}$-VCEs  in the same wordnet categories and examine qualitatively and quantitatively the influence of different threat models used during adversarial training (AT) on the quality of $l_{1.5}$- and $l_2$-VCEs.
    \item In \cref{app:further-evaluation} we discuss additional evaluation of $l_p$-VCEs using FID scores on CIFAR10 and based on pixel-level segmentations in ImageNet-S \cite{gao2021luss}.
    \item In \cref{app:afw} we discuss details of AFW and provide the proof of \cref{thm:fw_intersection}. In this section, we compare the performance of AFW, APGD and FW with different hyper-parameters on CIFAR10 models. 
    \item In \cref{sec:exp-details} we discuss experimental details.
    \item In \cref{sec:spurious} 
    we provide more details on the spurious features we discovered using VCEs and present in \cref{app:fishes} an additional example of a spurious feature where our VCEs show that human features are associated with the class ``tench''. 
    \item In \cref{app:diffusion-vces} we show how VCEs using guided and regularized diffusion process following \cite{avrahami2021blended} can be generated.
    
    Currently, this is the only method which uses a generative model, works on ImageNet, provides code and does something similar to the generation of VCEs.
    \item In \cref{app:user-study} we show the best and the worst examples from the user study. 
    \item In \cref{app:random-selection} we show randomly selected VCEs for both ImageNet and CIFAR10.
    \item In \cref{app:comparing-threat-models} we compare $3$ threat models for different radii.

    \end{itemize}
    
    \section{
    VCEs with robust models
    } \label{sec:app-robustness}
    We study several aspects of the generation of VCEs with (adversarially) robust models.
    
     \subsection{$l_{1.5}$-VCEs for ImageNet and CIFAR10.}
\begin{figure*}[htb!]
     \centering

\begin{tabular}{c|ccc|ccc|}
     \hline
     \multicolumn{1}{c|}{Original}&  
     
     \multicolumn{1}{C{.135\textwidth}}{
     $\epsilon_{1.5}=50$} & 
     \multicolumn{1}{C{.135\textwidth}}{$\epsilon_{1.5}=75$} &
     \multicolumn{1}{C{.135\textwidth}|}{$\epsilon_{1.5}=100$}
     & \multicolumn{1}{C{.135\textwidth}}{
     $\epsilon_{1.5}=50$} & 
     \multicolumn{1}{C{.135\textwidth}}{$\epsilon_{1.5}=75$} &
     \multicolumn{1}{C{.135\textwidth}|}{$\epsilon_{1.5}=100$}\\
     \hline

     \begin{subfigure}{0.135\textwidth}\centering
     \caption*{\tiny trium. arch: 0.79}
     \includegraphics[width=1\textwidth]{images/imagenet_temple/original.png}
     \end{subfigure}&

     \begin{subfigure}{0.135\textwidth}\centering
     \caption*{\tiny $\rightarrow$trium. arch: 1.00}
     \includegraphics[width=1\textwidth]{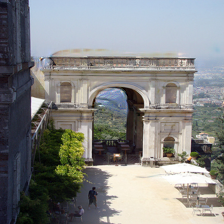}
     \end{subfigure} &

     \begin{subfigure}{0.135\textwidth}\centering
     \caption*{\tiny $\rightarrow$trium. arch: 1.00}
     \includegraphics[width=1\textwidth]{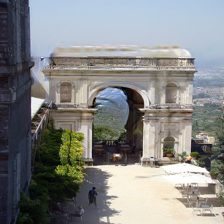}
     \end{subfigure} &

     \begin{subfigure}{0.135\textwidth}\centering
     \caption*{\tiny $\rightarrow$trium. arch: 1.00}
     \includegraphics[width=1\textwidth]{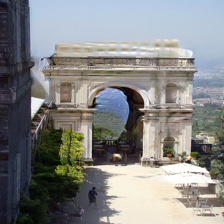}
     \end{subfigure} &
     
     \begin{subfigure}{0.135\textwidth}\centering
     \caption*{\tiny $\rightarrow$megalith: 0.90}
     \includegraphics[width=1\textwidth]{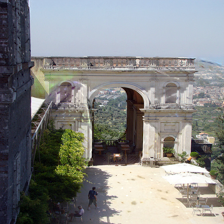}
     \end{subfigure} &

     \begin{subfigure}{0.135\textwidth}\centering
     \caption*{\tiny $\rightarrow$megalith: 1.00}
     \includegraphics[width=1\textwidth]{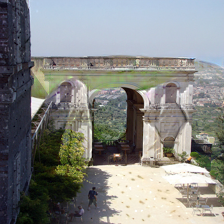}
     \end{subfigure} &

     \begin{subfigure}{0.135\textwidth}\centering
     \caption*{\tiny $\rightarrow$megalith: 1.00}
     \includegraphics[width=1\textwidth]{images/imagenet_temple/target_1_radius_2.png}
     \end{subfigure} \\
\hline

     \end{tabular}\\

      \caption{\label{fig:app-IN-VCEs1}\textbf{Further  $l_{1.5}$-VCEs for WordNet synset classes} for the Madry\cite{robustness_short}+FT model with varying radii. ILSVRC2012 images are transformed into different classes from the same WordNet synset as the original. }
     \end{figure*}

     For ImageNet we use as in the main paper the 
     Madry $l_2$-robust model \cite{robustness_short} plus finetuning for multiple-norm robustness \cite{croce2021adversarial_short} to generate more $l_{1.5}$-VCEs for classes belonging to a related WordNet synset in Figure \ref{fig:app-IN-VCEs1} and \ref{fig:app-IN-VCEs2}. In detail, we traverse the WordNet tree starting at the root node and compute  for each inner node the number of ILSVRC2012 leaf-nodes below it. If a node has between 3 and 10 ILSVRC2012 leafs below it, we add it and do not further process any child nodes. If it has more than 10 
     leafs below it, we process each child node the same way.
     By doing this, we can create clusters of related classes that can act as meaningful targets for our VCEs. For each cluster, we then randomly sample an image from one of the ILSVRC2012 leafs and create VCEs into all classes in that cluster. This is particularly important on ImageNet, as it contains very different classes and for some pairs changing one into the other is not feasible via a subtle change and would require large budgets to modify the entire image content. 
     One can see that our $l_{1.5}$-VCEs realize quite subtle class-specific changes of the original image.
    
     \subsection{Ablation of threat models during AT.} 

     We have seen in \cref{tab:cifar10-bench-fid} that in order to generate realistic $l_2$ VCEs one needs (adversarially) robust models. In this subsection we want to investigate, which threat model used for adversarial training VCEs is sufficient to produce realistic VCEs. We do this study for models trained both on CIFAR10 and ImageNet and check the corresponding $l_{1.5}$- and $l_2$-VCEs.

     \textbf{CIFAR10.} For CIFAR10 we train seven models with adversarial training all with PreActResNet-18 architecture \cite{he2016identity}: five using a $l_2$-threat model with radii $\epsilon_2 \in \{0.1, 0.25, 0.5, 0.75, 1\}$, one with $l_{1}$-threat model with radius $\epsilon_1 = 12 $, and one with $l_{\infty}$-threat model with radius $\epsilon_{\infty} = 8 / 255$.

     As it can be seen from the \cref{tab:ablation-cifar10-l2,tab:ablation-cifar10-1-5,tab:cifar10-abl-fid}, for $l_2$-AT a certain sufficiently large radius $\epsilon_2$ is required. Whereas $0.1$ seems not sufficient, $0.25,0.5,0.75$ look visually similar, while for $1.0$ one observes also more artefacts again. The same is true for the $l_\infty$-robust model. The best model is actually the $l_1$-robust model which produces at least in \cref{tab:ablation-cifar10-l2,tab:ablation-cifar10-1-5} the best visual quality apart form the multiple-norm fine-tuned model GU+FT which we have used throughout the paper and which achieves the best visual quality. Thus one take away message is that multiple-norm robust models are working for all $l_p$-VCEs with $p\in \{1.5,2\}$ due to their simultaneous adversarial robustness in all threat models. These observations are also supported by our quantitative evaluation via FID-scores in \cref{,tab:cifar10-abl-fid}.
    
    Finally, by comparing \cref{tab:ablation-cifar10-l2,tab:ablation-cifar10-1-5}, we can see again that $l_{1.5}$-VCEs produce more sparse and object-related changes.

     \textbf{ImageNet.} For ImageNet-1k we compare six models with ResNet50 architecture \cite{he2016identity} which are either taken from \cite{robustness_short} or are fine-tuned versions of these models. 
     The first one is non-robust, which we denote by \textit{RN-50}. Two are pre-trained models from \cite{robustness_short}: \textit{Madry} $l_2$, which corresponds to \textit{Madry} model in the main paper and \textit{Madry} $l_{\infty}$. First one is $l_2$-adversarially trained with $\epsilon_2=3$. Second one is $l_{\infty}$-adversarially trained with $\epsilon_{\infty}=4/255$. The other three models we obtained by fine-tuning \textit{Madry} $l_{2}$ and \textit{Madry} $l_{\infty}$. One is \textit{Madry} $l_2$\textit{+FT}, which corresponds to the \textit{Madry+FT} model in the main paper. It is obtained by using multiple-norm robust fine-tuning for 3 epochs with $\epsilon_1=255$ and $\epsilon_{\infty}=4/255$.
    The other one is \textit{Madry} $l_{\infty}$\textit{+FT}, which is obtained by the same fine-tuning, but for 1 epoch. The last one, \textit{Madry} $l_2$+$l_1$\textit{FT}, is obtained by fine-tuning for 1 epoch only wrt $l_1$. From \cref{fig:ablation-imagenet-1-5,fig:ablation-imagenet-l2,tab:in1k-abl-fid} similar observations can be made, that is multiple-norm fine-tuning can significantly improve both the image quality, and ID/OD FIDs by increasing $l_1$-, $l_{1.5}$- and $l_2$-RA, and that $l_{1.5}$ produce more object-related changes.

     \begin{figure*}[htb!]
     \centering

    \begin{tabular}{c|ccc|ccc|}
     \hline
     \multicolumn{1}{c|}{Original} & \multicolumn{1}{C{.135\textwidth}}{$\epsilon_{1.5}=50$} & 
     \multicolumn{1}{C{.135\textwidth}}{$\epsilon_{1.5}=75$} &
     \multicolumn{1}{C{.135\textwidth}|}{$\epsilon_{1.5}=100$} &
     \multicolumn{1}{C{.135\textwidth}}{$\epsilon_{1.5}=50$} & 
     \multicolumn{1}{C{.135\textwidth}}{$\epsilon_{1.5}=75$} &
     \multicolumn{1}{C{.135\textwidth}|}{$\epsilon_{1.5}=100$}
     \\
\hline

     \end{tabular}\\

      \caption{\label{fig:app-IN-VCEs2}\textbf{Further  $l_{1.5}$-VCEs for WordNet synset classes} for the Madry\cite{robustness_short}+FT model with varying radii. ILSVRC2012 images are transformed into different classes from the same WordNet synset as the original.}
     \end{figure*}

     \begin{figure*}
       
     \centering
      \begin{tabular}{c|cccccc}
                   \hline
                   \multicolumn{1}{c|}{Orig.} & \multicolumn{1}{C{.13\textwidth}}{RN-50}&\multicolumn{1}{C{.13\textwidth}}{Madry $l_2$}&\multicolumn{1}{C{.13\textwidth}}{Madry $l_2$+FT}&\multicolumn{1}{C{.13\textwidth}}{Madry $l_{\infty}$}&\multicolumn{1}{C{.13\textwidth}}{Madry $l_{\infty}$+FT}&\multicolumn{1}{C{.13\textwidth}}{Madry $l_{2}$+$l_{1}$ FT}\\
\hline
     \end{tabular}
     
      \caption{\label{fig:ablation-imagenet-1-5}\textbf{ImageNet: Ablation of the threat model during training.} $l_{1.5}$-VCEs at $\epsilon_{1.5}=50$ for ImageNet models trained adversarially with varying threat models.}

     \centering
     \begin{tabular}{c|cccccc}
                   \hline
                   \multicolumn{1}{c|}{Orig.} & \multicolumn{1}{C{.13\textwidth}}{RN-50}&\multicolumn{1}{C{.13\textwidth}}{Madry $l_2$}&\multicolumn{1}{C{.13\textwidth}}{Madry $l_2$+FT}&\multicolumn{1}{C{.13\textwidth}}{Madry $l_{\infty}$}&\multicolumn{1}{C{.13\textwidth}}{Madry $l_{\infty}$+FT}&\multicolumn{1}{C{.13\textwidth}}{Madry $l_{2}$+$l_{1}$FT}\\
\hline
     \end{tabular}

     \caption{\label{fig:ablation-imagenet-l2}\textbf{ImageNet: Ablation of the threat model during training.} $l_2$-VCEs at $\epsilon_2=12$ for ImageNet models trained to be adversarially robust with respect to different threat models. Here $l_1$-finetuning \cite{croce2021adversarial_short} (outermost right column) is fine-tuned to be $l_1$-robust.}
     \end{figure*}
            
    \begin{figure*}
     
     \centering
     \begin{tabular}{cc||ccccccc}
                   \hline
                   \multicolumn{1}{C{0.1\textwidth}}{$\;\;$ Original} & \multicolumn{1}{C{0.1\textwidth}||}{$\;\;$ GU+FT}&\multicolumn{1}{C{0.1\textwidth}}{$l_2$,\par$\epsilon$=$0.1$}&\multicolumn{1}{C{0.1\textwidth}}{$l_2$,\par$\epsilon$=$0.25$}&\multicolumn{1}{C{0.1\textwidth}}{$l_2$,\par$\epsilon$=$0.5$}&\multicolumn{1}{C{0.1\textwidth}}{$l_2$,\par $\epsilon$=$0.75$}&\multicolumn{1}{C{0.1\textwidth}}{$l_2$,\par $\epsilon$=$1$}&\multicolumn{1}{C{0.1\textwidth}}{$l_{\infty}$,\par$\epsilon$=$8/255$}&\multicolumn{1}{C{0.1\textwidth}}{$l_1$,\par
                       $\epsilon$=$12$}\\
\hline
     \end{tabular}
     
      \caption{\label{tab:ablation-cifar10-1-5}\textbf{CIFAR-10: Ablation of the threat model during training.} $l_{1.5}$-VCEs at $\epsilon_{1.5}=6$ for CIFAR-10 models trained adversarially with varying threat models.}

     \centering
     \begin{tabular}{cc||ccccccc}
                   \hline
                  \multicolumn{1}{C{0.1\textwidth}}{$\;\;$ Original} & \multicolumn{1}{C{0.1\textwidth}||}{$\;\;$ GU+FT}&\multicolumn{1}{C{0.1\textwidth}}{$l_2$,\par$\epsilon$=$0.1$}&\multicolumn{1}{C{0.1\textwidth}}{$l_2$,\par$\epsilon$=$0.25$}&\multicolumn{1}{C{0.1\textwidth}}{$l_2$,\par$\epsilon$=$0.5$}&\multicolumn{1}{C{0.1\textwidth}}{$l_2$,\par $\epsilon$=$0.75$}&\multicolumn{1}{C{0.1\textwidth}}{$l_2$,\par $\epsilon$=$1$}&\multicolumn{1}{C{0.1\textwidth}}{$l_{\infty}$,\par$\epsilon$=$8/255$}&\multicolumn{1}{C{0.1\textwidth}}{$l_1$,\par
                       $\epsilon$=$12$}\\
\hline
     \end{tabular}
     
      \caption{\label{tab:ablation-cifar10-l2}\textbf{CIFAR-10: Ablation of the threat model during training.} $l_2$-VCEs at $\epsilon_2=2.5$ for CIFAR-10 models trained to be adversarially robust with respect to different threat models.}
     
     \end{figure*}

     \begin{table*}[ht!]
        
         \centering
         \caption{\textbf{CIFAR10:} Evaluation of employed (robust) classifiers trained with different threat models for standard accuracy, $l_1$-, $l_{1.5}$- and $l_2$-robust accuracy (RA) evaluated at $\epsilon_1=12$, $\epsilon_{1.5}=1.5$, and $\epsilon_2=0.5$ respectively (first 1k test points). Further, FID scores for $l_2$- and $l_{1.5}$-VCEs for in-and out-of-distribution inputs and their average is shown. The threat model is indicated in the headers.}
          \begin{tabular}{cc|c||c|c|c|c|c|c|c}
          &
           \multicolumn{1}{c|}{}&\multicolumn{1}{c||}{\textbf{WRN-70-16}}& \multicolumn{7}{c}{\textbf{PreActResNet-18}} \\
                       
                       \cline{2-10}
                       
                       &\multicolumn{1}{c|}{}& \multicolumn{1}{C{.07\textwidth}||}{$\;\;\;\;$GU+FT}&\multicolumn{1}{C{.07\textwidth}|}{$l_2$,\par$\epsilon=0.1$}&\multicolumn{1}{C{.07\textwidth}|}{$l_2$,\par$\epsilon=0.25$}&\multicolumn{1}{C{.07\textwidth}|}{$l_2$,\par$\epsilon=0.5$}&\multicolumn{1}{C{.07\textwidth}|}{$l_2$,\par $\epsilon=0.75$}&\multicolumn{1}{C{.07\textwidth}|}{$l_2$,\par $\epsilon=1$}&\multicolumn{1}{C{.07\textwidth}|}{$l_{\infty}$,\par$\epsilon=8/255$}&\multicolumn{1}{C{.07\textwidth}}{$l_1$,\par
                       $\epsilon=12$}\\
        
        \cline{2-10} \cline{2-10}
       & Acc.  & 90.8  & \cellcolor{green!25}  91.6 & 90.8 & 88.8 & 84.8 & 80.6 & 82.8 & 87.1 \\

        \cline{2-10}
      &$l_1$-RA & 58.0 & 9.8 & 15.3 & 25.7 & 35.2 & 44.0 & 7.1 & \cellcolor{green!25} 60.1\\
       
       \cline{2-10}
      & $l_{1.5}$-RA & \cellcolor{green!25}  76.7 & 54.8 & 62.2 & 66.2 & 67.1  & 64.9 & 44.6 & 66.8  \\
       
       \cline{2-10}
      & $l_2$-RA & \cellcolor{green!25}  79.2 & 61.8 & 66.5 & 68.6 &  67.7 & 65.3 & 59.9 & 64.9 \\
      \\

         \cline{2-10}
          
           \multirow{3}{2.3cm}{\textbf{FID scores for $l_{1.5}$-VCE}}

        & ID 
        & 11.4 & 16.3 & 14.1 & 13.7 & 12.9 & 11.9 & 22.9 & 13.6 \\
        
        \cline{2-10}
     &
        OD 
        & 46.2 & 46.9 & 45.8 & 48.8 & 51.6 & 52.2 & 65.2 & 48.9\\
        
        \cline{2-10}
     &
        Avg. 
        & 28.8 & \textbf{31.6}  & \textbf{29.9} & \textbf{31.2} & \textbf{32.2} & 32.1 & 44.0 & \textbf{31.2} \\
        \\
       \multirow{3}{2.3cm}{\textbf{FID scores for $l_{2}$-VCE}}&
         
        ID 
        & 11.9 & 21.3 & 18.2 & 16.3 & 15.0 & 13.7 & 20.2 & 20.1\\
        
        \cline{2-10}
     &
        OD 
        & 41.2 & 47.2 & 44.7 & 47.1 & 49.6 & 50.1 &60.7 & 48.7\\
        
        \cline{2-10}
     &
        Avg. 
        & \textbf{26.7}  & 34.3 & 31.4 & 31.7 & 32.3  & \textbf{31.9} & \textbf{40.5} & 34.4\\

        \cline{2-10}
        
      \end{tabular}

        \label{tab:cifar10-abl-fid}

         \centering
          \caption{\textbf{ImageNet:} Evaluation of employed (robust) classifiers trained with different threat models for standard accuracy, $l_1$-, $l_{1.5}$- and $l_2$-robust accuracy (RA) evaluated at $\epsilon_1=255$, $\epsilon_{1.5}=12.5$, and $\epsilon_2=2$ respectively (1k test points). Further, FID scores for $l_2$- and $l_{1.5}$-VCEs for in-and out-of-distribution inputs and their average is shown. The threat model is indicated in the headers.}
         \label{tab:in1k-abl-fid}
          \begin{tabular}{cc|c|c|c|c|c|c}

         & \multicolumn{7}{c}{\textbf{ResNet50}} \\
                 
                 \cline{2-8}
                 
        &\multicolumn{1}{c|}{}& \multicolumn{1}{C{.11\textwidth}|}{RN-50}&\multicolumn{1}{C{.11\textwidth}|}{Madry $l_2$}&\multicolumn{1}{C{.11\textwidth}|}{Madry $l_2$+FT}&\multicolumn{1}{C{.11\textwidth}|}{Madry $l_{\infty}$}&\multicolumn{1}{C{.11\textwidth}|}{Madry $l_{\infty}$+FT}&\multicolumn{1}{C{.11\textwidth}}{Madry $l_2$+$l_1$FT} \\
         
         \cline{2-8}
       & Acc.  & \cellcolor{green!25} 76.0 &  57.9 & 57.5 & 62.4 & 57.6 &  56.9  \\

        \cline{2-8}
        & $l_1$-RA & 0.0 & 13.0 & 25.5 & 0.0 & 22.6 & \cellcolor{green!25} 29.2 \\
        
        \cline{2-8}
        & $l_{1.5}$-RA & 0.0 & 37.4 & \cellcolor{green!25}  40.1 & 2.8 & 39.4 & 37.9  \\
        
        \cline{2-8}
       & $l_2$-RA & 0.0 & \cellcolor{green!25} 45.7 & 44.6 & 18.3 & 44.5 & 40.3   \\

        \\

         \cline{2-8}

       \multirow{3}{2.3cm}{\textbf{FID scores for $l_{1.5}$-VCE}} &
       ID & 
         9.4 & 8.4 & 6.9 & 8.3 & 6.8 & 7.1 \\
        
        \cline{2-8}
       & OD & 
        50.3 & 24.3 & 22.6 & 39.3 & 23.3 & 25.5 \\
        
        \cline{2-8}
        & Avg. & 
        \textbf{29.8} & 16.4 & \textbf{14.8} & 23.8 & \textbf{15.1}  & \textbf{16.3} \\
         \\
         
         \multirow{3}{2.3cm}{\textbf{FID scores for $l_{2}$-VCE}}
        &ID & 
         20.0 & 8.4 & 7.9  & 9.1 & 7.8 & 9.5 \\
        
        \cline{2-8}
        & OD & 
         60.3 & 22.8 & 23.1 & 34.9 & 23.3 & 28.5\\
        
        \cline{2-8}
        & Avg. & 
        40.1 & \textbf{15.6} & 15.5 & \textbf{22.0} & 15.6 & 19.0  \\

        \cline{2-8}
       \end{tabular}
       
     \end{table*}

    \clearpage
    
    \section{Further evaluation}
    First, we show the qualitative and quantitative (with FIDs) evaluation of the different models on CIFAR10.
    \begin{figure*}[htp!]
     \centering
     \small
     \begin{tabular}{cccccccc}
                   \hline
                   \multicolumn{1}{c}{Orig.} & \multicolumn{1}{C{.121\textwidth}}{
                  
                  \small BiT-M\cite{KolesnikovEtAl2019_short}}&\multicolumn{1}{C{.12\textwidth}}{RST-S\cite{CarEtAl19_short}}&\multicolumn{1}{C{.12\textwidth}}{RATIO\cite{augustin2020_short}}&\multicolumn{1}{C{.12\textwidth}}{GU\cite{gowal2020uncovering_short}}&\multicolumn{1}{C{.12\textwidth}}{GU+FT\cite{croce2021adversarial_short}}&\multicolumn{1}{C{.12\textwidth}}{PAT\cite{laidlaw2020perceptual_short}}&\multicolumn{1}{C{.12\textwidth}}{HenC\cite{hendrycks2020augmix_short}}\\
                   \hline
                   \begin{subfigure}{0.12\textwidth}\centering
     
     \caption*{car $\rightarrow$ truck}
     \includegraphics[width=1\textwidth]{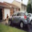}
     \end{subfigure}&\begin{subfigure}{0.12\textwidth}\centering
                
                \caption*{$p_i$:$0.01$, $p_e$:$1.00$}
                \includegraphics[width=1\textwidth]{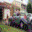}
                
                \end{subfigure}&\begin{subfigure}{0.12\textwidth}\centering
                
                \caption*{$p_i$:$0.51$, $p_e$:$1.00$}              \includegraphics[width=1\textwidth]{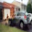}
                
                \end{subfigure}&\begin{subfigure}{0.12\textwidth}\centering
                
                \caption*{$p_i$:$0.78$, $p_e$:$1.00$}
                \includegraphics[width=1\textwidth]{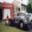}
                
                \end{subfigure}&\begin{subfigure}{0.12\textwidth}\centering
                
                \caption*{$p_i$:$0.44$, $p_e$:$1.00$}
                \includegraphics[width=1\textwidth]{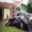}
                
                \end{subfigure}&\begin{subfigure}{0.12\textwidth}\centering
                
                \caption*{$p_i$:$0.83$, $p_e$:$1.00$}
                \includegraphics[width=1\textwidth]{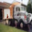}
                
                \end{subfigure}&\begin{subfigure}{0.12\textwidth}\centering
                
                \caption*{$p_i$:$0.57$, $p_e$:$0.99$}
                \includegraphics[width=1\textwidth]{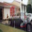}
                
                \end{subfigure}&\begin{subfigure}{0.12\textwidth}\centering
                
                \caption*{$p_i$:$0.04$, $p_e$:$1.00$}
                \includegraphics[width=1\textwidth]{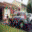}
                
                \end{subfigure}\\
                \begin{subfigure}{0.12\textwidth}\centering

    \vspace*{3.6mm}
     \caption*{$\rightarrow$ cat}
     \includegraphics[width=1\textwidth]{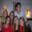}
     \end{subfigure}&\begin{subfigure}{0.12\textwidth}\centering
                
                \caption*{$p_i$:$0.05$, $p_e$:$1.00$}
                \includegraphics[width=1\textwidth]{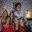}
                
                \end{subfigure}&\begin{subfigure}{0.12\textwidth}\centering
                
                \caption*{$p_i$:$0.26$, $p_e$:$0.95$}
                \includegraphics[width=1\textwidth]{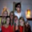}
                
                \end{subfigure}&\begin{subfigure}{0.12\textwidth}\centering
                
                \caption*{$p_i$:$0.10$, $p_e$:$1.00$}
                \includegraphics[width=1\textwidth]{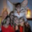}
                
                \end{subfigure}&\begin{subfigure}{0.12\textwidth}\centering
                
                \caption*{$p_i$:$0.54$, $p_e$:$1.00$}
                \includegraphics[width=1\textwidth]{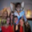}
                
                \end{subfigure}&\begin{subfigure}{0.12\textwidth}\centering
                
                \caption*{$p_i$:$0.35$, $p_e$:$1.00$}
                \includegraphics[width=1\textwidth]{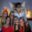}
                
                \end{subfigure}&\begin{subfigure}{0.12\textwidth}\centering
                
                \caption*{$p_i$:$0.45$, $p_e$:$0.95$}
                \includegraphics[width=1\textwidth]{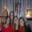}
                
                \end{subfigure}&\begin{subfigure}{0.12\textwidth}\centering
                
                \caption*{$p_i$:$0.07$, $p_e$:$1.00$}
                \includegraphics[width=1\textwidth]{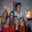}
                
                \end{subfigure} 
    
     \end{tabular}
     
     \caption{\label{tab:benchmark-cifar10} \textbf{CIFAR10:} $l_2$-VCEs  of radius $\epsilon=2.5$ of different classifiers  for the change ``car $\rightarrow$ truck''
     (top) and ``frog $\rightarrow$ cat'' (middle)
     and ``$\rightarrow$ cat'' for an OOD image from 80M TM \cite{torralba200880} not belonging to any of the CIFAR10 classes (bottom).  
     We denote by $p_i$ resp. $p_e$ the confidence in the target class for the original image and the generated VCE. All generated VCEs are valid as high confidence in the target class is achieved but only adversarially robust models, see \cref{tab:cifar10-bench-fid}, show class-specific changes.}
     \vspace{-3mm}
     \end{figure*}

    For this purpose, we qualitatively compare different CIFAR10 classifiers in \cref{tab:benchmark-cifar10} and quantitatively in \cref{tab:cifar10-bench-fid}: \textit{BiT-M:} a non-robust BiT-M-R50x1 model \cite{KolesnikovEtAl2019_short} with high accuracy, \textit{RST-s:} a WRN-28-10 trained only with additive noise using stability training \cite{CarEtAl19_short}, \textit{RATIO:} a WRN-34-10 using $l_2$-adv. training on the in-and out-distribution, \textit{GU:} the SOTA $l_2$-robust WRN-70-16 from \cite{gowal2020uncovering_short} trained with extra data, \textit{GU+FT:} we fine-tune the GU model \cite{croce2021adversarial_short} to get multiple-norm-robust, \textit{PAT:} the PAT-self ResNet50 from  \cite{laidlaw2020perceptual_short} which uses adversarial training with respect to a perceptual distance, \textit{HenC:} a ResNext29 trained to be robust against common corruptions (no adversarial training). Note that all $l_2$-adversarially trained models were trained using a radius of $0.5$. All models are the original models from the authors, most of them are available in RobustBench \cite{croce2020robustbench}, and we report their standard and robust accuracy against adversarial $l_1$- and $l_2$-perturbations in \cref{tab:cifar10-bench-fid}.
    The perturbation model for the generation of the VCEs in \cref{eq:VCE-RATIO} is a $l_2$-ball of radius 2.5 following \cite{augustin2020_short}. From \cref{tab:benchmark-cifar10} one can observe that the two non-robust models BiT-M and HenC do not produce any meaningful counterfactuals.

    Surprisingly, the RST-s model has some adversarial robustness but its $l_2$-VCEs do minimal changes to the image which show little class-specific features of the target class. Thus the FID score for the in-distribution is low, but the FID score of the out-distribution is high. Moreover, the PAT-model, which uses a threat model with respect to a perceptual distance but also has non-trivial $l_1$- and $l_2$-adversarial robustness, produces VCEs that show strong artefacts. The best VCEs are generated by RATIO, GU and GU+FT, which also have the highest $l_2$-adversarial robustness. Among them, RATIO and GU+FT produce the most visually realistic VCEs and also have the best FID scores for in- and out-distribution. In particular, the multiple-norm finetuning of the GU model seems to significantly boost the generative properties, both for $l_2$-VCEs and the $l_{1.5}$-VCEs (see \cref{sec:threatmodel}). High $l_2$-adversarial robustness alone, as for the SOTA GU model, is not the only factor which determines the quality of the generated VCEs. In \cref{sec:app-robustness} we provide a more detailed analysis, \eg 
    which radius for the threat model for adversarial training is required for good VCEs, and also repeat a similar experiment for ImageNet even though there are fewer adversarially robust models available.
    
    \begin{table}[htp!]
         \caption{\textbf{CIFAR10:} Evaluation of (robust) classifiers for standard accuracy, $l_1$-, $l_{1.5}$- and $l_2$-robust accuracy (RA) evaluated at $\epsilon_1=12$, $\epsilon_{1.5}=1.5$, and $\epsilon_2=0.5$ respectively (first 1k test points). Further, FID scores for $l_1$-, $l_{1.5}$-, and $l_2$-VCEs for in-and out-of-distribution inputs and their average are shown. For all classifiers except RATIO $l_{1.5}$-VCEs attain the best average FID-score.} 
        
        \label{tab:cifar10-bench-fid}

         \centering
          \begin{tabular}{cc|c|c|c|c|c|c|c}
                       \cline{2-9}
                       
                       & & \multicolumn{1}{C{0.09\textwidth}|}{BiT-M}&\multicolumn{1}{C{0.09\textwidth}|}{RST-s}&\multicolumn{1}{C{0.09\textwidth}|}{RATIO}&\multicolumn{1}{C{0.09\textwidth}|}{GU}&\multicolumn{1}{C{0.1\textwidth}|}{ GU+FT}&\multicolumn{1}{C{0.09\textwidth}|}{PAT}&\multicolumn{1}{C{0.09\textwidth}}{HenC}\\
        \cline{2-9} \cline{2-9}
       &  Acc.  & \cellcolor{green!25}
       97.4 & 87.9 & 94.0 & 94.7 & 90.8 & 82.4 & 95.8 \\

        \cline{2-9}
      &   $l_1$-RA & 0.0 & 36.5 & 34.3 & 33.4 & \cellcolor{green!25} 58.0 & 32.9 & 0.0 \\
        \cline{2-9}
      &   $l_{1.5}$-RA & 0.0 & 70.4 & 75.4 & \cellcolor{green!25} 76.8 & 76.7 & 59.2 & 0.3 \\
        \cline{2-9}
       &  $l_2$-RA & 0.0 & 71.4 & 79.9 & \cellcolor{green!25} 81.7 & 79.2 & 62.4 & 0.1 \\
       \\

         \cline{2-9}
          
       \multirow{3}{2.3cm}{\textbf{FID scores for $l_{1}$-VCE}} &     ID 
      & 25.1 & 26.0 & 24.4 & 31.1 & 10.2 & 29.1 & 22.7 \\
         \cline{2-9}
     &     OD
        & 79.5 & 72.6 & 57.8 & 71.4 & 52.7  & 72.2 & 79.5\\
         \cline{2-9}
       & Avg. 
        &  52.3 & 49.8  & 41.1  & 51.3 & 31.5  & 50.6 & 51.1 \\ 
        \\

          \multirow{3}{2.3cm}{\textbf{FID scores for $l_{1.5}$-VCE}}  &  ID 
        & 12.2 & 8.5   &  11.7 & 12.3 & 9.2  & 14.4 & 18.8 \\
        \cline{2-9}
       &  OD 
        & 62.7 & 51.6 & 30.4 
        &  52.5 & 43.4 & 51.6 & 62.4  \\
         \cline{2-9}
     &    Avg. 
        & \textbf{42.5} & \textbf{30.1} & 
        19.5 & \textbf{32.4} & \textbf{26.3}  & \textbf{33.0} & \textbf{40.6} \\
        \\
         
   \multirow{3}{2.3cm}{\textbf{FID scores for $l_{2}$-VCE}} &      ID 
        & 55.4 & 10.3   &  12.2 & 15.8 & 11.9  & 18.8 & 37.9 \\
         \cline{2-9}
       & OD 
        & 83.9 & 50.7 &  26.0 & 53.9 & 41.2 & 49.0 & 67.2 \\
        \cline{2-9}
       &  Avg. 
        & 69.7 & 30.5 & 
        \textbf{19.1} & 34.9 & 26.7 & 33.9 & 52.6
        \\

         \cline{2-9}
       \end{tabular}
      
    \end{table}
    
    \label{app:further-evaluation}
    We additionally evaluate the $l_p$-VCEs in a user study and evaluate their object-relatedness using ImageNet-S dataset \cite{gao2021luss}.

    \paragraph{ImageNet-S Evaluation}
     We use the recently introduced dataset ImageNet-S \cite{gao2021luss} with pixel-level segmentations of ImageNet images to evaluate how well the $l_p$-VCEs are located on the object. We limit the evaluation to a subset of 2048 images that only contain a single class 
     and where the segmented pixels are connected (one object). The target class is chosen using the WordNet hierarchy. For each VCE we compute the absolute difference to the original image and sum up over color channels. We normalize this to get a distribution $p$ of pixel changes over the image.
    
     Given $p$, we evaluate three metrics:
    a) the expected distance $\mathbb{E}[d]$ calculates  the distance to the closest mask pixel for each pixel in the image and then calculates a weighted average using the distribution of changes $p$. The larger the expected distance $\mathbb{E}[d]$ the more far away are most changes from the object. However, note that a VCE might need to change also pixels outside the object so that an ``ideal'' VCE need not have zero expected distance; b) the probability mass $\int p$ of changes located in the segmentation; c) the Intersection over Union (IOU). As this is a measure between two binary images, we have to discretize $p$. To do this, we sort $p$ in a descending fashion and activate pixels until their cumulative probability is at least $0.95$, thus they explain most of the changes of the image. While the previous metrics are optimal if all changes are inside the mask, the IOU requires the changes to also cover most of the object. Thus if the changes are too sparse, the IOU will be small even if all of them are located on the object.

    \begin{figure*}[htp!]
\centering
    \begin{tabular}{c c c}
         
         \begin{tabular}{|c|c|c|c|}
             \hline
             & 
             {\footnotesize $l_{1}$, $\epsilon=400$}
             &
             {\footnotesize $l_{1.5}$, $\epsilon=50$}
             &
             {\footnotesize $l_{2}$, $\epsilon=12$}\\
             \hline
             \begin{subfigure}{0.15\textwidth}\centering
             \vspace{4mm}
             \caption*{ GT: Samoyed}
             \includegraphics[width=\textwidth]{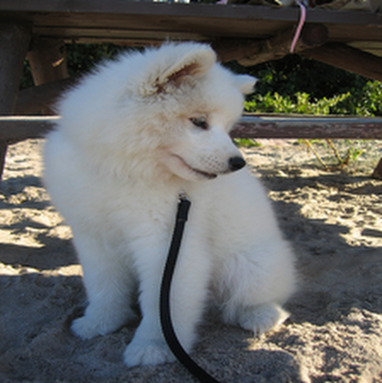}
             \end{subfigure} &
             \begin{subfigure}{0.15\textwidth}\centering
             \caption*{  $\rightarrow$ Pomeranian: 0.33}
             \includegraphics[width=\textwidth]{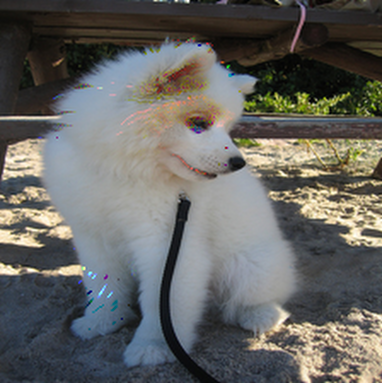}
             \end{subfigure} &
             \begin{subfigure}{0.15\textwidth}\centering
             \caption*{  $\rightarrow$ Pomeranian: 0.94}
             \includegraphics[width=\textwidth]{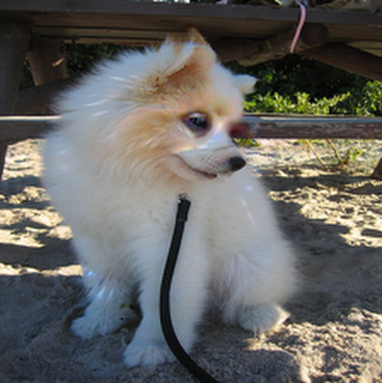}
             \end{subfigure} &
             \begin{subfigure}{0.15\textwidth}\centering
             \caption*{  $\rightarrow$ Pomeranian: 0.99}
             \includegraphics[width=\textwidth]{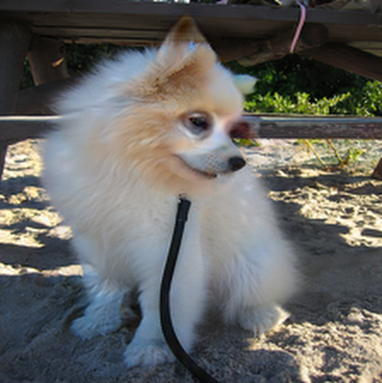}
             \end{subfigure}\\
             \hline
             \begin{subfigure}{0.15\textwidth}\centering
              \caption*{ Mask}
             \includegraphics[width=\textwidth]{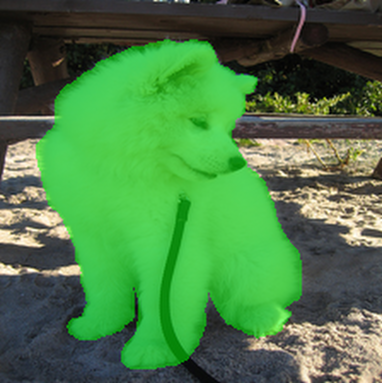}
             \end{subfigure} &
             \begin{subfigure}{0.15\textwidth}\centering
             \caption*{ $\mathbb{E}[d] = 0.04$ }  
             \includegraphics[width=\textwidth]{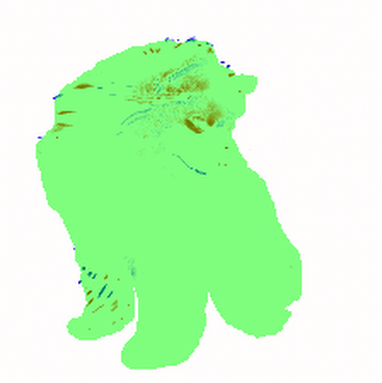}
             \end{subfigure} &
             \begin{subfigure}{0.15\textwidth}\centering
              \caption*{ $\mathbb{E}[d] = 2.85$ }  
             \includegraphics[width=\textwidth]{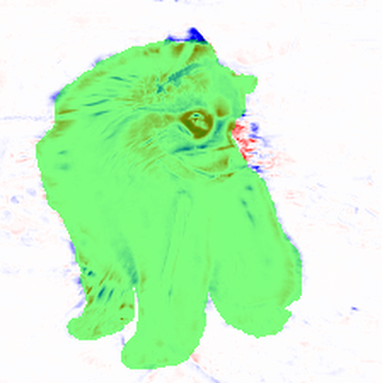}
             \end{subfigure} &
             \begin{subfigure}{0.15\textwidth}\centering
             \caption*{ $\mathbb{E}[d] = 77.82$ }  
             \includegraphics[width=\textwidth]{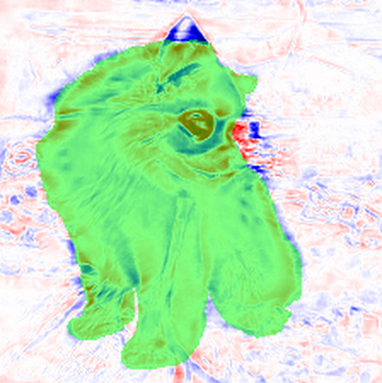}
             \end{subfigure}\\
             \hline
         \end{tabular}
         &
         & 
   \raisebox{5mm}{ \begin{tabular}{|c|c|c|c|}
              \hline
              \multicolumn{4}{|c|}{\makecell{ImageNet-S\\Avg statistics }}\\
              \hline
                &  $l_1$ & $l_{1.5}$ & $l_2$\\
              \hline
              $\mathbb{E}[d]$ & 7.63 & 14.38 & 21.20\\
              $\int p $ & 0.69 &  0.57 & 0.45\\
              $\text{IOU}@0.95$ & 0.05 & 0.38 & 0.35\\
              \hline
            
          \end{tabular}}
         \end{tabular}
    \caption{Evaluation of changes of $l_p$-VCE based on pixel-level segmentations in the ImageNet-S \cite{gao2021luss} dataset. $l_1$-VCEs are too sparse, while $l_2$-VCEs induce a lot of changes in the background. The $l_{1.5}$-VCE induce more subtle changes which are more concentrated on the object.}
    \label{fig:imagenet_s}
    
\end{figure*}

    Results and an illustration in Fig. \ref{fig:imagenet_s}. The $l_1$-VCEs are well located on the object but too sparse which results in a very small IOU. The $l_2$-VCEs have a large expected distance $\mathbb{E}[d]$ which means that a lot of pixels outside of the object are changed: this can also be verified from the given example in Fig. \ref{fig:imagenet_s} where most background pixels are changed. $l_{1.5}$-VCEs offer a good balance between the two extremes. They are largely located on or close to the object without being too sparse, which results in more realistic images than the $l_1$-VCEs.
    
    \section{Auto-Frank-Wolfe} 
    \label{app:afw}
    \subsection{Details of AFW algorithm}\label{app:afw_details}
    We extend the scheme of APGD for selecting the step size to the FW scheme to make the choice of $\gamma^k$ adaptive (see Eq.~\eqref{eq:fw_conv_comb}).
    The algorithm follows Algorithm 1 from \cite{croce2020reliable} where the step size at each iteration is indicated by $\eta^k$: however, the update in lines $7,8$ becomes
    \begin{align}\label{eq:afw_update}
     \hat{x}^{k+1} = (1-\eta^{k})\hat{x}^{k} + \eta^k s^k,
     \end{align}
     where $s^k$ is the solution of \cref{eq:lmo}, computed according to the \cref{thm:fw_intersection}, and $\eta^k = \frac{M}{\sqrt{k}+2},$ where in the beginning $M=2$, and then at steps that satisfy 
     the same \textit{Condition 1} and \textit{Condition 2} as in \cite{croce2020reliable}, we update $M \leftarrow M \cdot 0.75$.
     
    \subsection{Omitted proof}\label{app:afw_proof}
    \begin{refproof}[Proof of \cref{thm:fw_intersection}]

    Introducing $$v = |w|, \; \gamma = \max\{-x \sign{w}, (1-x)\sign{w}\}
    $$
    
    we have an equivalent problem 
    \begin{align*} \argmin_{\eta \in \R^d}\, - \inner{v, \eta} \quad \mathrm{s. th.} 
    &\sum_i \eta_i^p \leq \epsilon^p,\\ 
    & \eta_i \in [0, \gamma_i]\quad i = 1, \ldots, d,
    \end{align*}
    with solution $\eta^*$ for which $\delta^* = \eta^* \sign{w}$. Note that $\gamma_i=0$ implies $\eta_i=0$, 
    and similarly, if $w_i = v_i = 0$ then $\eta_i = 0$ since the $i$-th coordinate does not contribute to the objective function. Thus in the following we can assume $\gamma_i, v_i > 0$.
    The Lagrangian is given by $$L(\eta, \mu, \alpha, \beta) = - \inner{v, \eta} + \mu (\inner{\mathbf{1}, \eta^p} - \epsilon^p) - \inner{\alpha, \eta} + \inner{\beta, \eta - \gamma},$$
    with $\mu \geq 0$, $\alpha, \beta \in \R^d_+$, and the power of a vector is meant componentwise. It has gradient wrt $\eta$ $$\nabla_\eta L(\eta, \mu, \alpha, \beta) = -v + p\mu \eta^{p - 1} - \alpha + \beta,$$
    which yields the optimality conditions \begin{align*} -v_i + p\mu \eta_i^{p-1} - \alpha_i + \beta_i =& 0 \quad i=1, \ldots, d\\
    \alpha_i \eta_i =& 0 \quad i=1, \ldots, d\\
    \beta_i (\gamma_i - \eta_i) =& 0 \quad i=1, \ldots, d \\
    \mu (\inner{\mathbf{1}, \eta^p} - \epsilon^p) =& 0\\
    \alpha_i, \beta_i, \mu \geq& 0 \quad i=1, \ldots, d.
    \end{align*}
    We can distinguish three cases: \begin{itemize} \item $\alpha_i=0, \beta_i > 0 \Rightarrow \eta_i = \gamma_i, \quad p\mu \gamma_i^{p-1} < v_i$,
    \item $\alpha_i > 0, \beta_i = 0 \Rightarrow \eta_i = 0, \quad v_i = -\alpha_i < 0$, 
    \item $\alpha_i=0, \beta_i = 0 \Rightarrow \eta_i \in [0, \gamma_i], \quad p\mu \eta_i^{p-1} = v_i$.
    \end{itemize}
    Note that the second case is not possible since we assume $v_i > 0$, while in the third case we have $p\mu \gamma_i^{p-1}\geq v_i$ because of the interval $\eta_i$ belongs to. Thus, we have $$\mu < \frac{v_i}{p\gamma_i^{p-1}} \Longrightarrow \eta_i = \gamma_i, \quad \mu \geq \frac{v_i}{p\gamma_i^{p-1}} \Longrightarrow \eta_i = \left(\frac{v_i}{p\mu}\right)^\frac{1}{p - 1}
    $$
    If we have $\mu > 0$, then $$\sum_i\eta_i^p = \sum_{i\in I^-(\mu)} \gamma_i^p + \sum_{i \in I^+(\mu)}\left(\frac{v_i}{p\mu}\right)^\frac{p}{p - 1} = \epsilon^p,
    $$
    
    with $I^-(\mu) = \{i: \mu < \frac{v_i}{p\gamma_i^{p-1}}\}$ and $I^+(\mu) = \{i: \mu \geq \frac{v_i}{p\gamma_i^{p-1}}\}$.
    
    This is equivalent to finding the solution $\mu^*$ of the equation
    
    \begin{equation} f(\mu) := \sum_{i\in I^-(\mu)} \gamma_i^p + \sum_{i \in I^+(\mu)}\left(\frac{v_i}{p\mu}\right)^\frac{p}{p - 1} - \epsilon^p = 0.
     \end{equation}

    which exists if $f(0) \geq 0$ since $f$ is continuous on $[0, +\infty)$ and converges to $-\epsilon^p$ for $\mu \rightarrow + \infty$. The solution is also unique as $f$ is decreasing. To find $\mu^*$ it is possible to sort the set $M = \{\frac{v_i}{p\gamma_i^{p-1}}, i=1, \ldots, d \} \cup \{0, +\infty\}$. There exists $m_j \in M$ is such that $f(m_j) \geq 0$ and $f(m_{j+1}) < 0$, and $\mu^*$ solves $$\sum_{i\in I^-(m_j)} \gamma_i^p + \sum_{i \in I^+(m_j)}\left(\frac{v_i}{p\mu}\right)^\frac{p}{p - 1} - \epsilon^p = 0,
    $$
    that is $$(p\mu^*)^\frac{p}{p - 1} = \sum_{i\in I^+(m_j)} v_i^\frac{p}{p - 1} \left(\epsilon^p - \sum_{i\in I^-(m_j)} \gamma_i^p\right)^{-1},
    $$
    and $\eta^*_i = \min\{\gamma_i, \left(\frac{v_i}{p\mu^*}\right)^\frac{1}{p - 1}\}$. Finally note that if $f(0)< 0$, then $\mu = 0$ which implies $\eta_i = \gamma_i$ if $v_i > 0$, $\eta_i = 0$ else. Note the most complex operation involved is the sorting of the set $M$, which gives the complexity of the algorithm.
    \end{refproof}
    \subsection{Additional experiments}\label{app:afw_additional}
    \begin{figure}[hbt!]
    \centering
    \begin{minipage}[t]{.5\textwidth}
        \centering
        \captionsetup{width=.9\linewidth}
        \includegraphics[width=0.9\columnwidth]{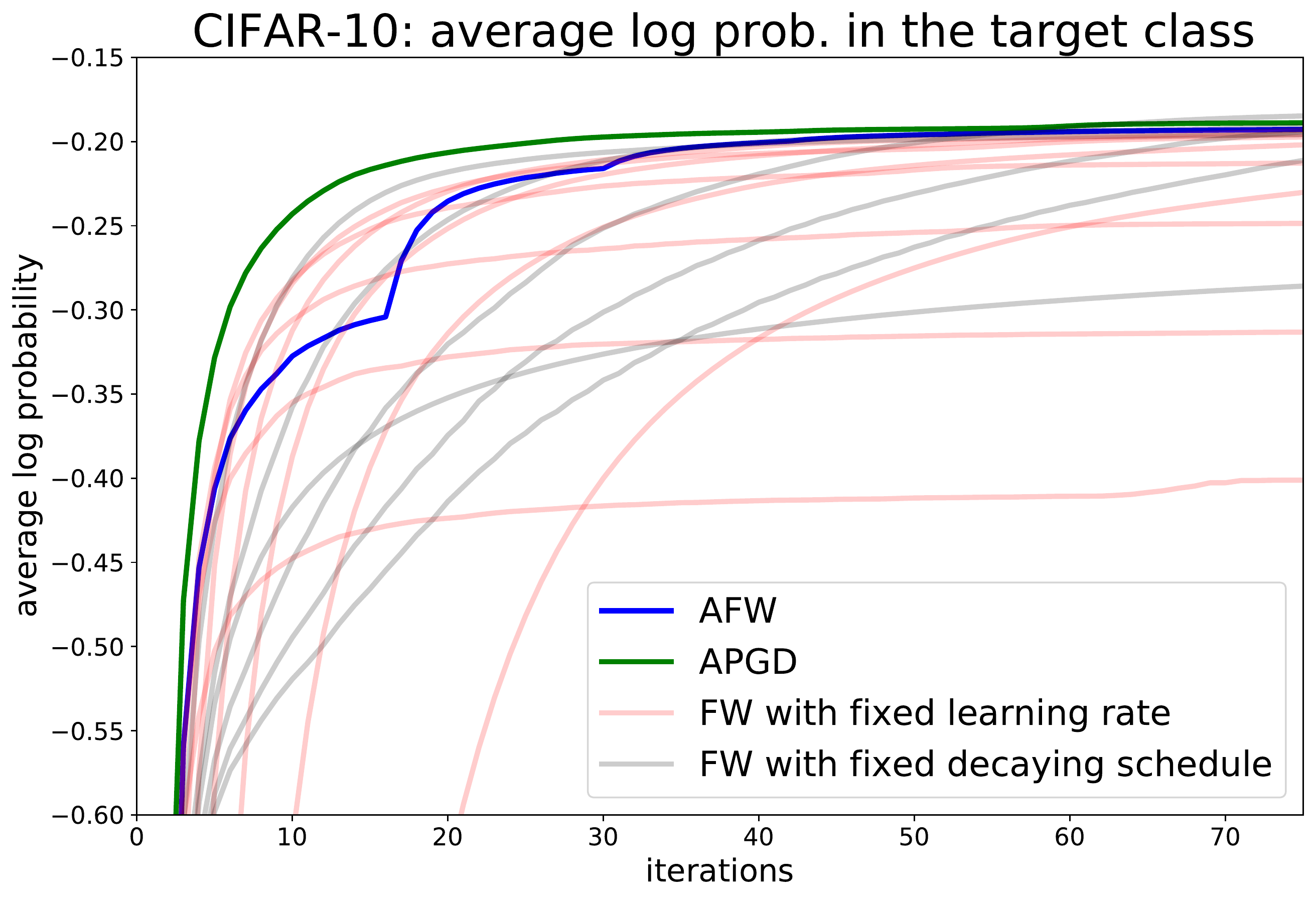}
        \captionof{figure}{\textbf{APGD and AFW vs FW.} APGD and AFW outperform most of the FW versions except for 2 in terms of final mean (over 1000 images) log probability.}
        \label{fig:apgd_vs_afw_cifar10_plot}
    \end{minipage}%
    \begin{minipage}[t]{.5\textwidth}
        \centering
        \captionsetup{width=.9\linewidth}
        \includegraphics[width=0.9\columnwidth]{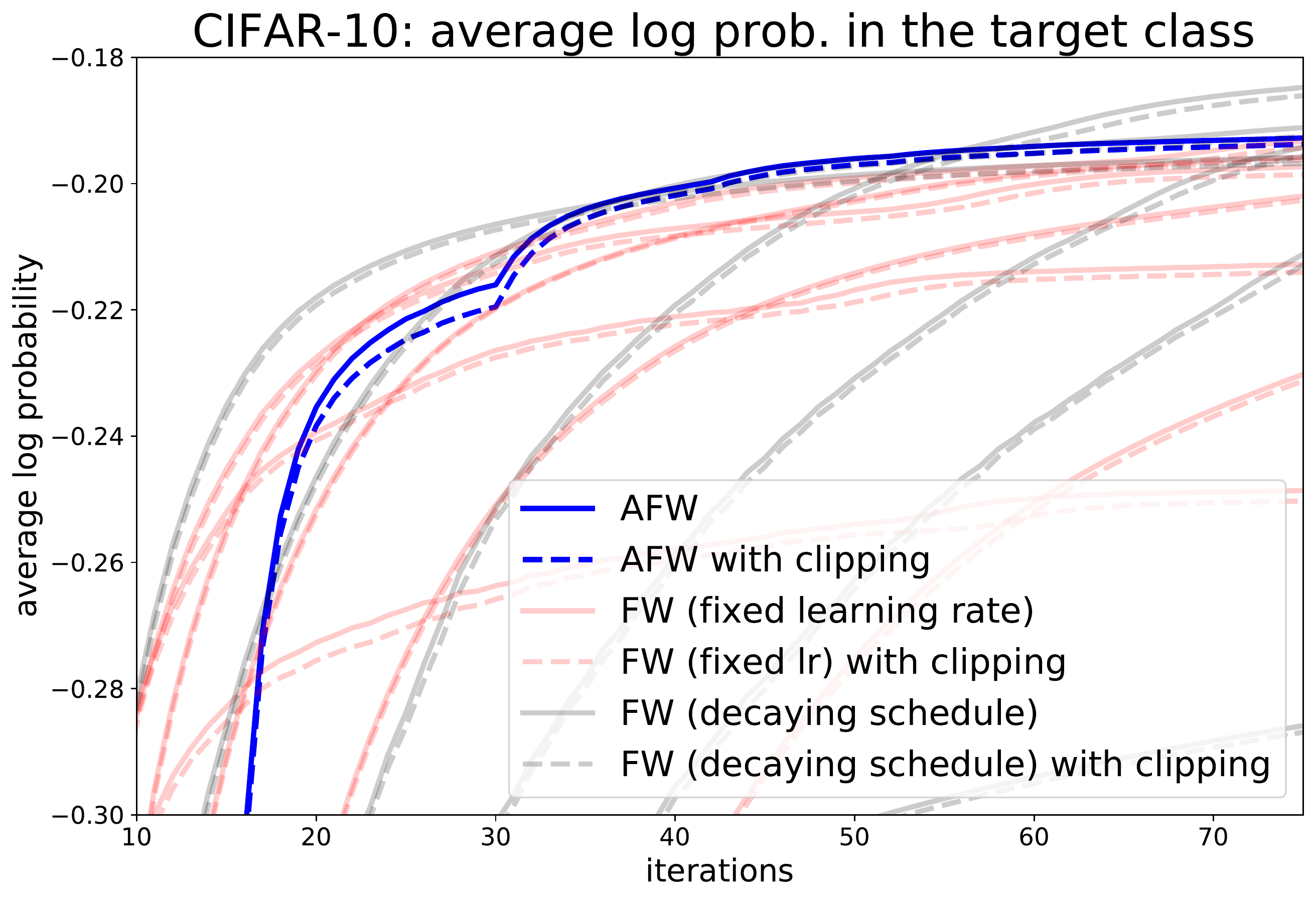}
        \captionof{figure}{\textbf{AFW, FW with vs without clipping.} Applying \cref{thm:fw_intersection} to do AFW in the intersection of $l_p$-ball and $[0, 1]^d$ consistently outperforms constrained optimization in the $l_p$-ball followed by clipping.}
        \label{fig:clipping_fw_cifar10}
    \end{minipage}
    \end{figure}
    First, we repeat the experimental comparison of AFW to APGD and the existing variants of FW reported in \cref{sec:afw} on CIFAR10, using the robust models, in the $l_2$-threat model with $\epsilon=2.5$. \Cref{fig:apgd_vs_afw_cifar10_plot} shows that even in this case AFW is competitive with the best methods and then effective optimizer. Moreover, while the highest loss is attained by a version of FW with decaying schedule, this is achieved only by tuning $\gamma_0$ which needs to be done for each classifier and threat model, unlike for AFW and APGD.
    
    Second, we study the effect of including the box-constraints in the FW scheme rather than clipping after each iteration as done by prior works. Then, we compare in \cref{fig:clipping_fw_cifar10} the performance of AFW and FW (same setup as above) with either the linear optimization suggested by \cref{thm:fw_intersection} (solid lines) or the standard optimization in the $l_p$-ball followed by clipping (dashed lines). We observe that considering the true threat models, i.e. the intersection of $l_p$-ball and $[0, 1]^d$, yields consistently a small improvement over the baseline.

    \subsection{Effect of AFW versus APGD on the resulting $l_2$-VCEs}\label{app:afw-vs-apgd}
    
    \begin{figure}[hbt!]
  \includegraphics[width=0.5\textwidth]{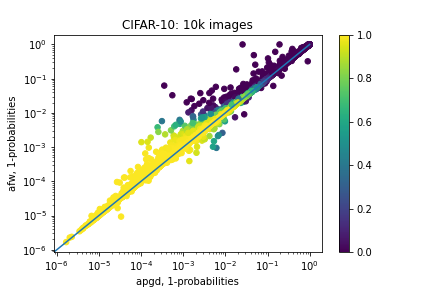}
  \includegraphics[width=0.5\textwidth]{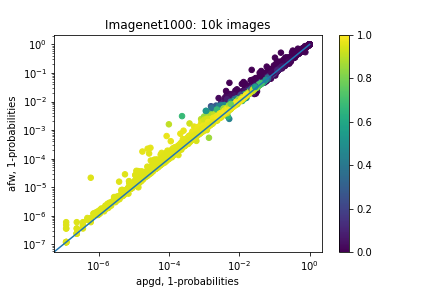}
  \caption{\label{fig:apgd_vs_afw_diagonal}\textbf{End probabilities for APGD vs AFW.} APGD performs similarly to AFW in terms of end probabilities for 10.000 VCEs used for the FID computation. For CIFAR10 GU+FT model is used, and for ImageNet - Madry+FT. Color represents the density of the datapoints with yellow indicating the highest density.}
  
    \end{figure}

    In this section we compare $l_2$-VCEs for AFW and APGD quantitatively. We use for CIFAR10 the GU+FT model and for ImageNet the Madry+FT models to compare the end probabilities for 10.000 VCEs used for the FID computation 
    in \cref{fig:apgd_vs_afw_diagonal}. One sees that the two methods achieve similar results, with APGD producing only slightly higher probabilities.

    \begin{table*}[!htb] 
    \begin{minipage}{1\textwidth}
         
         \centering
         
          \caption{Temperature obtained after the temperature scaling for CIFAR10 models.}
         \begin{tabular}{c|c|c|c|c|c|c|c}
          \multicolumn{8}{c}{\textbf{Temperature for CIFAR10 models introduced in \cref{sec:vces_intro}}}\\
                       \hline
                       
                       & \multicolumn{1}{C{0.12\textwidth}|}{BiT-M}&\multicolumn{1}{C{0.12\textwidth}|}{RST-s}&\multicolumn{1}{C{0.12\textwidth}|}{RATIO}&\multicolumn{1}{C{0.12\textwidth}|}{GU}&\multicolumn{1}{C{0.12\textwidth}|}{GU+FT}&\multicolumn{1}{C{0.12\textwidth}|}{PAT}&\multicolumn{1}{C{0.12\textwidth}}{HenC}\\
        \hline \hline
        Temp.  & 1.02 & 1.12  & 0.42 & 0.69  & 0.59 & 0.52 & 1.31 \\

        \hline
        
        \end{tabular} 
        
        \begin{tabular}{c|c|c|c|c|c|c|c}
          \multicolumn{8}{c}{\textbf{Temperature for CIFAR10 models introduced in \cref{tab:cifar10-abl-fid}}}\\
                       \hline
                       
                       & \multicolumn{1}{C{0.12\textwidth}|}{$l_2$, $\epsilon$=$0.1$}&\multicolumn{1}{C{0.12\textwidth}|}{$l_2$, $\epsilon $=$0.25$}&\multicolumn{1}{C{0.12\textwidth}|}{$l_2$, $\epsilon$=$0.5$}&\multicolumn{1}{C{0.12\textwidth}|}{$l_2$, $\epsilon$=$0.75$}&\multicolumn{1}{C{0.12\textwidth}|}{$l_2$,  $\epsilon$=$1$}&\multicolumn{1}{C{0.12\textwidth}|}{$l_{\infty}$, $\epsilon$=$8/255$}&\multicolumn{1}{C{0.12\textwidth}}{$l_{1}$, $\epsilon$=$12$}\\
        \hline \hline
        Temp. & 1.49  & 1.44 & 1.31 & 0.90 & 0.79 & 0.8 & 1.03 \\

        \hline
        \end{tabular}

        \label{table:cifar10-temperature}

    \end{minipage}
    \begin{minipage}{1\textwidth}
         
         \centering
        
          \begin{tabular}{c|c|c|c|c|c|c}
          \multicolumn{7}{c}{\textbf{Temperature for ImageNet models introduced in \cref{tab:in1k-abl-fid}}}\\
                       \hline
                       
                       & \multicolumn{1}{C{0.14\textwidth}|}{RN50}&\multicolumn{1}{C{0.14\textwidth}|}{Madry}&\multicolumn{1}{C{0.14\textwidth}|}{Madry + FT}&\multicolumn{1}{C{0.14\textwidth}|}{Madry $l_{\infty}$}&\multicolumn{1}{C{0.14\textwidth}|}{Madry $l_{\infty}$ + FT}&\multicolumn{1}{C{0.14\textwidth}}{Madry + $l_1$ FT}\\
        \hline \hline
        Temp.  & 1.16 & 0.84 & 0.72 & 0.82 & 0.73 & 0.74\\

        \hline  
        \end{tabular}

    \end{minipage}
    \end{table*}
    
    \section{Experimental Details}\label{sec:exp-details}
    \subsection{Calibration} \label{app:calibration}

    In order to have comparable confidence values, we use CIFAR10.1 \cite{recht2018cifar10.1,torralba200880} for CIFAR10 models, respectively 20.000 images from the ImageNet test set for ImageNet models to do temperature scaling. In \cref{table:cifar10-temperature} 
    we show the computed values of temperature $T$, by which the output of each classifier is divided before computing applying the softmax function. We observe that models with standard training or with adversarial training with small radii are in general overconfident ($T > 1$), while the most robust ones, especially trained for multiple norm robustness, are underconfident ($T < 1$).

    \subsection{VCEs generation}
    For generating the VCEs with AFW and APGD we run the algorithms for 75 restarts and 5 random restarts. The final output is selected as the one attaining the highest value of the objective function, i.e. the log-probability of the target class.
    
    \subsection{Reproducibility}
    Most of the classifiers we use are publicly available. In particular, RATIO, GU and Hen can be found in RobustBench \cite{croce2020robustbench}, RST-s and PAT are provided by the original papers, BiT-M is a BiT-M-R50x1 model that can be fine-tuned following instructions from the authors of \cite{KolesnikovEtAl2019_short}, Madry $l_2$ and $l_\infty$ are part of the Robustness library \cite{robustness_short}, while the remaining ones are obtained via personal communications with the authors. Moreover, we train the classifiers for the ablation study in \cref{sec:app-robustness}.

    \section{Detection of spurious features via VCEs}\label{sec:spurious}
    We revisit in each subsection the spurious feature which we detected with our VCEs.
    Additionally, we illustrate a new spurious feature which is known, that certain classes of fish are associated with humans. For each case we show the VCEs and samples from the training set illustrating the origin of the spurious feature.
    
    \subsection{Failure Mode A: Watermarks}\label{app:watermarks}
\begin{figure*}[hbt!]
\begin{minipage}{1\textwidth}
      \centering
     \small
     \begin{tabular}{cccc}
     \hline
     \multicolumn{1}{C{0.25\textwidth}}{Original} & \multicolumn{1}{C{0.25\textwidth}}{$\epsilon_{1.5}=50$} & 
     \multicolumn{1}{C{0.25\textwidth}}{$\epsilon_{1.5}=75$} &
     \multicolumn{1}{C{0.25\textwidth}}{$\epsilon_{1.5}=100$}\\
     \hline
       
     \begin{subfigure}{00.25\textwidth}\centering
     \caption*{granny sm.: 0.92}
     \includegraphics[width=1\textwidth]{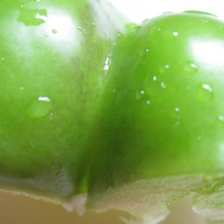}
     \end{subfigure} &
     
     \begin{subfigure}{00.25\textwidth}\centering
     \caption*{$\rightarrow$granny sm.: 1.00}
     \includegraphics[width=1\textwidth]{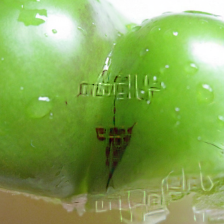}
     \end{subfigure} &

     \begin{subfigure}{00.25\textwidth}\centering
     \caption*{$\rightarrow$granny sm.: 1.00}
     \includegraphics[width=1\textwidth]{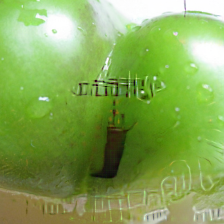}
     \end{subfigure} &

     \begin{subfigure}{00.25\textwidth}\centering
     \caption*{$\rightarrow$granny sm.: 1.00}
     \includegraphics[width=1\textwidth]{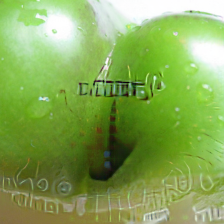}
     \end{subfigure} \\
     
      
      \begin{subfigure}{00.25\textwidth}\centering
     \caption*{lemon: 0.36}
     \includegraphics[width=1\textwidth]{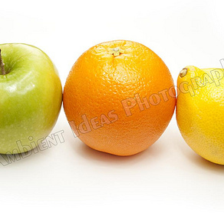}
     \end{subfigure} &
     
     \begin{subfigure}{00.25\textwidth}\centering
     \caption*{$\rightarrow$granny sm.: 0.96}
     \includegraphics[width=1\textwidth]{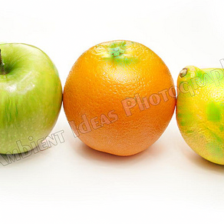}
     \end{subfigure} &

     \begin{subfigure}{00.25\textwidth}\centering
     \caption*{$\rightarrow$granny sm.: 0.99}
     \includegraphics[width=1\textwidth]{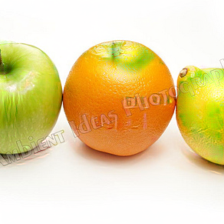}
     \end{subfigure} &

     \begin{subfigure}{00.25\textwidth}\centering
     \caption*{$\rightarrow$granny sm.: 1.00}
     \includegraphics[width=1\textwidth]{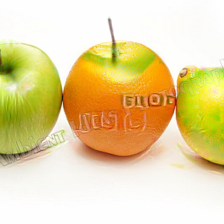}
     \end{subfigure} \\


      \begin{subfigure}{00.25\textwidth}\centering
     \caption*{bell pepper: 1.00}
     \includegraphics[width=1\textwidth]{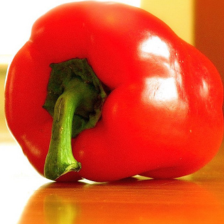}
     \end{subfigure} &
     
     \begin{subfigure}{00.25\textwidth}\centering
     \caption*{$\rightarrow$granny sm.: 0.01}
     \includegraphics[width=1\textwidth]{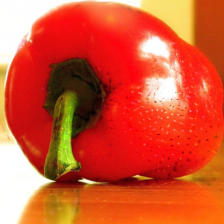}
     \end{subfigure} &

     \begin{subfigure}{00.25\textwidth}\centering
     \caption*{$\rightarrow$granny sm.: 0.09}
     \includegraphics[width=1\textwidth]{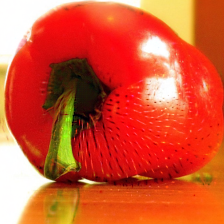}
     \end{subfigure} &

     \begin{subfigure}{00.25\textwidth}\centering
     \caption*{$\rightarrow$granny sm.: 0.75}
     \includegraphics[width=1\textwidth]{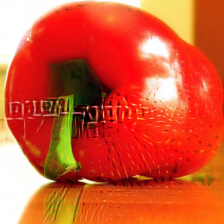}
     \end{subfigure} \\

     \end{tabular}
      \caption{\label{fig:imagent_granny_smith_failure_app}\textbf{Spurious feature of watermarks for the target class ``granny smith''.} $l_{1.5}$-VCEs for Madry\cite{robustness_short}+FT with varying radii for $3$ images for the target class ``granny smith''. They show again that classifier has learned to associate the spurious ``text'' feature with this class.
    }
\end{minipage}
\begin{minipage}{1\textwidth}
     \centering
     \small
     \begin{tabular}{cccc}
     \hline
     \multicolumn{4}{c}{\textbf{Watermarks in the training set of ``granny smith'' in ImageNet}} \\
     \hline \\
       
     \begin{subfigure}{0.25\textwidth}\centering
     \includegraphics[width=1\textwidth]{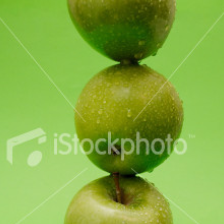}
     \end{subfigure} &
     
     \begin{subfigure}{0.25\textwidth}\centering
     \includegraphics[width=1\textwidth]{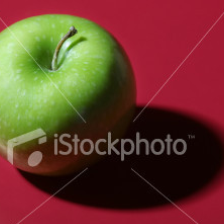}
     \end{subfigure} &

     \begin{subfigure}{0.25\textwidth}\centering
     \includegraphics[width=1\textwidth]{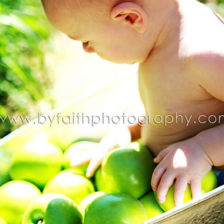}
     \end{subfigure} &

     \begin{subfigure}{0.25\textwidth}\centering
     \includegraphics[width=1\textwidth]{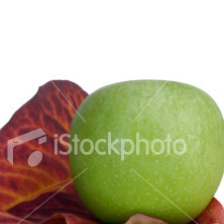}
     \end{subfigure} \\

     \end{tabular}
      \caption{\label{fig:imagent_granny_smith_watermarks_examples}More examples from the training set of the class ``granny smith'' in ImageNet. In total 90 out of 1300 training images contain a watermark, out of which 53 contain the one of ``iStockphoto''.  The fraction of watermarked images seems to be significantly larger than for other classes.
    }
    \end{minipage}
     \end{figure*}

    We show in \cref{fig:imagent_granny_smith_failure_app} more examples of the spurious text like feature appearing in the $l_{1.5}$-VCEs for the target class ``granny smith'' which are obviously unrelated to the class. In Section \ref{sec:debugging} we came up with the hypothesis that the reason for this spurious feature which the model has picked up is the large number of images containing a watermark, in particular one from ``iStockphoto'', in the training set of ``granny smith'', see \cref{fig:imagent_granny_smith_watermarks_examples} for other examples from the training set.  

    \subsection{Failure Mode B: Cages in shark images}\label{app:sharks}
    \begin{figure*}[hbt!]
\begin{minipage}{1\textwidth}
     \centering
     \small
     \begin{tabular}{cccc}
     \hline
     \multicolumn{1}{c}{Original} & \multicolumn{1}{C{0.25\textwidth}}{$\epsilon_{1.5}=50$} & 
     \multicolumn{1}{C{0.25\textwidth}}{$\epsilon_{1.5}=75$} &
     \multicolumn{1}{C{0.25\textwidth}}{$\epsilon_{1.5}=100$}\\
     \hline
       
     \begin{subfigure}{0.25\textwidth}\centering
     \caption*{scuba diver: 0.42}
     \includegraphics[width=1\textwidth]{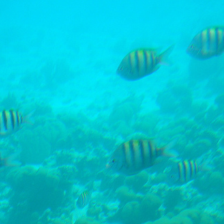}
     \end{subfigure} &
     
     \begin{subfigure}{0.25\textwidth}\centering
     \caption*{$\rightarrow$w. shark: 0.15}
     \includegraphics[width=1\textwidth]{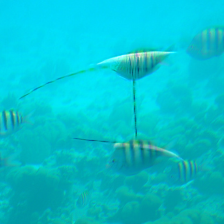}
     \end{subfigure} &

     \begin{subfigure}{0.25\textwidth}\centering
     \caption*{$\rightarrow$w. shark: 0.40}
     \includegraphics[width=1\textwidth]{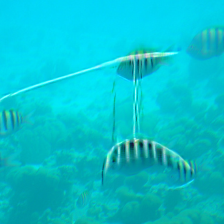}
     \end{subfigure} &

     \begin{subfigure}{0.25\textwidth}\centering
     \caption*{$\rightarrow$w. shark: 0.81}
     \includegraphics[width=1\textwidth]{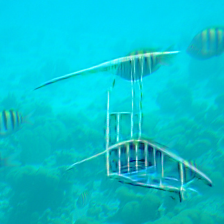}
     \end{subfigure} \\
     
      
      \begin{subfigure}{0.25\textwidth}\centering
     \caption*{coral reef: 0.29}
     \includegraphics[width=1\textwidth]{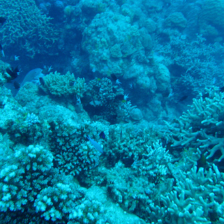}
     \end{subfigure} &
     
     \begin{subfigure}{0.25\textwidth}\centering
     \caption*{$\rightarrow$w. shark: 0.01}
     \includegraphics[width=1\textwidth]{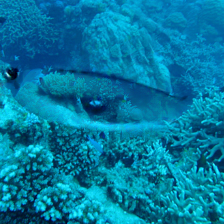}
     \end{subfigure} &

     \begin{subfigure}{0.25\textwidth}\centering
     \caption*{$\rightarrow$w. shark: 0.12}
     \includegraphics[width=1\textwidth]{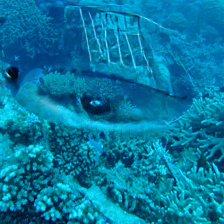}
     \end{subfigure} &

     \begin{subfigure}{0.25\textwidth}\centering
     \caption*{$\rightarrow$w. shark: 0.59}
     \includegraphics[width=1\textwidth]{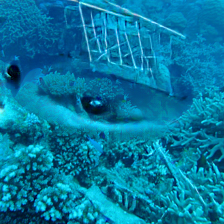}
     \end{subfigure} \\


      \begin{subfigure}{0.25\textwidth}\centering
     \caption*{hammerhead: 0.51}
     \includegraphics[width=1\textwidth]{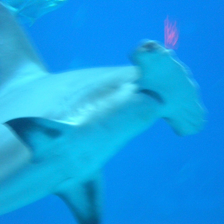}
     \end{subfigure} &
     
     \begin{subfigure}{0.25\textwidth}\centering
     \caption*{$\rightarrow$w. shark: 0.67}
     \includegraphics[width=1\textwidth]{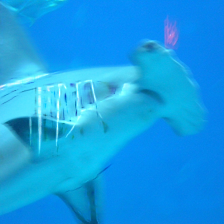}
     \end{subfigure} &

     \begin{subfigure}{0.25\textwidth}\centering
     \caption*{$\rightarrow$w. shark: 0.97}
     \includegraphics[width=1\textwidth]{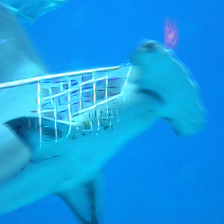}
     \end{subfigure} &

     \begin{subfigure}{0.25\textwidth}\centering
     \caption*{$\rightarrow$w. shark: 1.00}
     \includegraphics[width=1\textwidth]{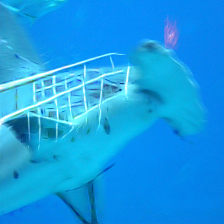}
     \end{subfigure} \\

     \end{tabular}
      \caption{\label{fig:imagent_sharks_cages_failure_app}\textbf{Spurious feature of ``cages'' for the target class ``white shark''.} $l_{1.5}$-VCEs for Madry\cite{robustness_short}+FT with varying radii for $3$ images for the target class ``white shark''. They show again that classifier has learned to associate the spurious ``cage'' feature with this class. 
    }
    \end{minipage}
    \begin{minipage}{1\textwidth}
          \centering
     \small
     \begin{tabular}{cccc}
     \hline
     \multicolumn{4}{c}{\textbf{``Cages'' in the training set of ``white shark'' in ImageNet}} \\
     \hline \\
       
     \begin{subfigure}{0.25\textwidth}\centering
     \includegraphics[width=1\textwidth]{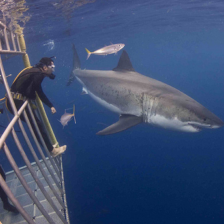}
     \end{subfigure} &
     
     \begin{subfigure}{0.25\textwidth}\centering
     \includegraphics[width=1\textwidth]{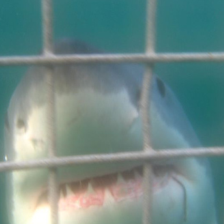}
     \end{subfigure} &

     \begin{subfigure}{0.25\textwidth}\centering
     \includegraphics[width=1\textwidth]{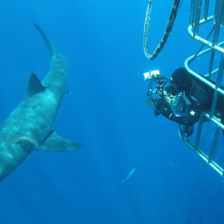}
     \end{subfigure} &

     \begin{subfigure}{0.25\textwidth}\centering
     \includegraphics[width=1\textwidth]{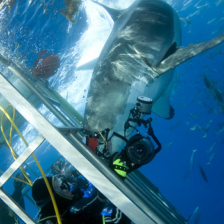}
     \end{subfigure} \\

     \end{tabular}
      \caption{\label{fig:sharks_cages}A large fraction of training images of the class ``white shark'' in ImageNet contains cages as the divers who take the photographs need to be protected.
    }
    \end{minipage}
     \end{figure*}

    In \cref{fig:imagent_sharks_cages_failure_app} we show more examples of the spurious ``cage'' feature the model has picked up for the white shark class. It is interesting that this is even a very dominating feature, in the sense that it easier to paint in some ``cage''-like structures in the image to change the class to ``white shark'' rather than changing some parts of the image into a white shark. In \cref{fig:sharks_cages} we show a sample of training images from the class ``white shark'' which shows this spurious feature of ``cages'' which the model has picked up.

    \subsection{Failure Mode C: Human features in fish images}\label{app:fishes}
    \begin{figure*}[hbt!]
\begin{minipage}{1\textwidth}
     \centering
     \small
     \begin{tabular}{cccc}
     \hline
     \multicolumn{1}{c}{Original} & \multicolumn{1}{C{0.25\textwidth}}{$\epsilon_{1.5}=50$} & 
     \multicolumn{1}{C{0.25\textwidth}}{$\epsilon_{1.5}=75$} &
     \multicolumn{1}{C{0.25\textwidth}}{$\epsilon_{1.5}=100$}\\
     \hline
       
     \begin{subfigure}{0.25\textwidth}\centering
     \caption*{eel: 0.18}
     \includegraphics[width=1\textwidth]{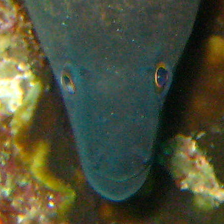}
     \end{subfigure} &
     
     \begin{subfigure}{0.25\textwidth}\centering
     \caption*{$\rightarrow$tench: 0.65}
     \includegraphics[width=1\textwidth]{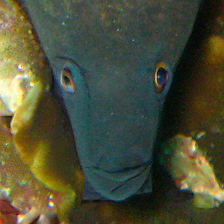}
     \end{subfigure} &

     \begin{subfigure}{0.25\textwidth}\centering
     \caption*{$\rightarrow$tench: 0.99}
     \includegraphics[width=1\textwidth]{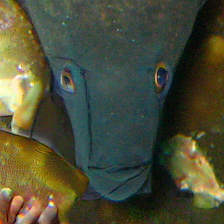}
     \end{subfigure} &

     \begin{subfigure}{0.25\textwidth}\centering
     \caption*{$\rightarrow$tench: 1.00}
     \includegraphics[width=1\textwidth]{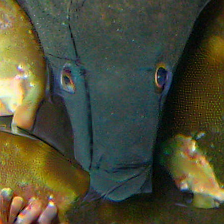}
     \end{subfigure} \\
     
      
      \begin{subfigure}{0.25\textwidth}\centering
     \caption*{goldfish: 0.32}
     \includegraphics[width=1\textwidth]{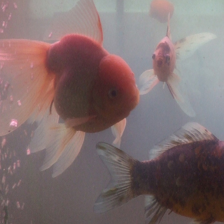}
     \end{subfigure} &
     
     \begin{subfigure}{0.25\textwidth}\centering
     \caption*{$\rightarrow$tench: 0.13}
     \includegraphics[width=1\textwidth]{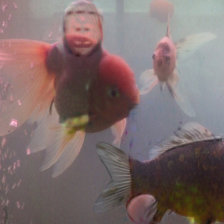}
     \end{subfigure} &

     \begin{subfigure}{0.25\textwidth}\centering
     \caption*{$\rightarrow$tench: 0.86}
     \includegraphics[width=1\textwidth]{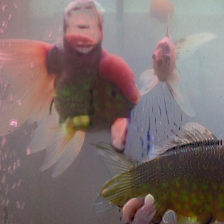}
     \end{subfigure} &

     \begin{subfigure}{0.25\textwidth}\centering
     \caption*{$\rightarrow$tench: 0.99}
     \includegraphics[width=1\textwidth]{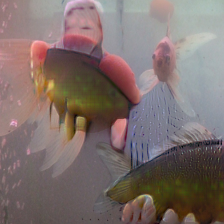}
     \end{subfigure} \\


      \begin{subfigure}{0.25\textwidth}\centering
     \caption*{stingray: 0.17}
     \includegraphics[width=1\textwidth]{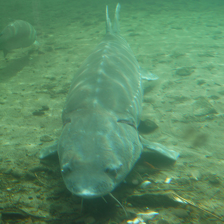}
     \end{subfigure} &
     
     \begin{subfigure}{0.25\textwidth}\centering
     \caption*{$\rightarrow$tench: 0.35}
     \includegraphics[width=1\textwidth]{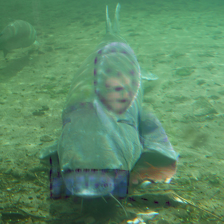}
     \end{subfigure} &

     \begin{subfigure}{0.25\textwidth}\centering
     \caption*{$\rightarrow$tench: 0.88}
     \includegraphics[width=1\textwidth]{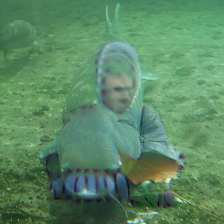}
     \end{subfigure} &

     \begin{subfigure}{0.25\textwidth}\centering
     \caption*{$\rightarrow$tench: 0.99}
     \includegraphics[width=1\textwidth]{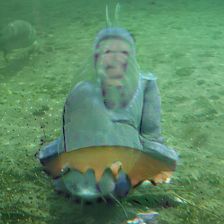}
     \end{subfigure} \\

     \end{tabular}
      \caption{\label{tab:imagent_fishes_hands_failure_app}\textbf{Spurious feature of ``human hands'' for the target class ``tench''.} $l_{1.5}$-VCEs for Madry\cite{robustness_short}+FT with varying radii for $3$ images for the target class ``tench''. They show again that classifier has learned to associate the spurious features of ``human hands'' or ``human faces'' with this class. 
    }
    \end{minipage}
    \begin{minipage}{1\textwidth}
           \centering
     \small
     \begin{tabular}{cccc}
     \hline
     \multicolumn{4}{c}{\textbf{``Human hands'' in the training set of ``tench'' in ImageNet}} \\
     \hline \\
       
     \begin{subfigure}{0.25\textwidth}\centering
     \includegraphics[width=1\textwidth]{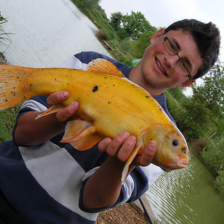}
     \end{subfigure} &
     
     \begin{subfigure}{0.25\textwidth}\centering
     \includegraphics[width=1\textwidth]{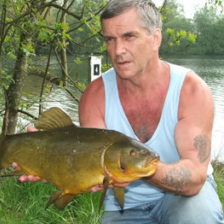}
     \end{subfigure} &

     \begin{subfigure}{0.25\textwidth}\centering
     \includegraphics[width=1\textwidth]{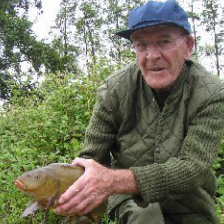}
     \end{subfigure} &

     \begin{subfigure}{0.25\textwidth}\centering
     \includegraphics[width=1\textwidth]{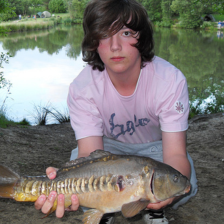}
     \end{subfigure} \\
    
     \end{tabular}
      \caption{\label{fig:tench-trainingset}
      A large fraction of training images of the class ``tench'' contains images of anglers presenting their ``tench'' into the camera. The model learns humans in particular human fingers as a spurious feature associated to ``tench''.
    }
    \end{minipage}
     \end{figure*}

    It is well known that some classes of fish, e.g. ``tench'', appear very often with humans in the training set, as this is a popular fish for anglers which they proudly present to the camera. In some of the $l_{1.5}$-VCEs for the class ``tench'', see examples in \cref{tab:imagent_fishes_hands_failure_app}, we see human fingers or even human faces appearing, which shows that the model has picked up this artefact of the training set. We show in  \cref{fig:tench-trainingset} examples from the training set of the class ``tench'' in ImageNet.
    \clearpage
    
    \subsection{Spurious features using other threat models}
    Here, we show, how the previously found  spurious features look using $l_2$- and $l_1$-VCEs. $l_2$-VCEs allow as well to see indicated spurious features in all three classes, while $l_1$-VCEs have many artefacts and do not seem to be useful for this task.
    \begin{figure*}[hbt!]
   \centering
     \small
     \begin{tabular}{cccc}
     \hline
     \multicolumn{1}{C{0.25\textwidth}}{Original} & \multicolumn{1}{C{0.25\textwidth}}{$\epsilon_{2}=12$} & 
     \multicolumn{1}{C{0.25\textwidth}}{$\epsilon_{2}=18$} &
     \multicolumn{1}{C{0.25\textwidth}}{$\epsilon_{2}=24$}\\
     \hline

     \begin{subfigure}{00.25\textwidth}\centering
     \caption*{granny sm.: 0.92}
     \includegraphics[width=1\textwidth]{images/failure_granny/original.png}
     \end{subfigure} &
     
     \begin{subfigure}{00.25\textwidth}\centering
     \caption*{$\rightarrow$granny sm.: 1.00}
     \includegraphics[width=1\textwidth]{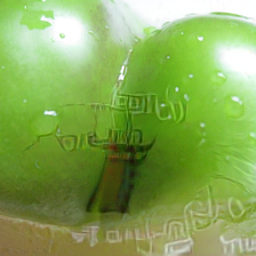}
     \end{subfigure} &

     \begin{subfigure}{00.25\textwidth}\centering
     \caption*{$\rightarrow$granny sm.: 1.00}
     \includegraphics[width=1\textwidth]{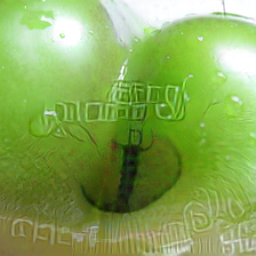}
     \end{subfigure} &

     \begin{subfigure}{00.25\textwidth}\centering
     \caption*{$\rightarrow$granny sm.: 1.00}
     \includegraphics[width=1\textwidth]{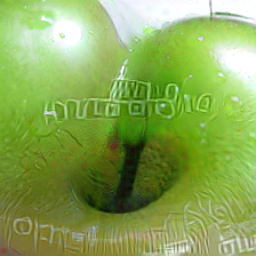}
     
     \end{subfigure} \\

      \begin{subfigure}{00.25\textwidth}\centering
     \caption*{hammerhead: 0.51}
     \includegraphics[width=1\textwidth]{images/failure_sharks_cages/original_re.png}
     \end{subfigure} &
     
     \begin{subfigure}{00.25\textwidth}\centering
     \caption*{$\rightarrow$w. shark: 0.67}
     \includegraphics[width=1\textwidth]{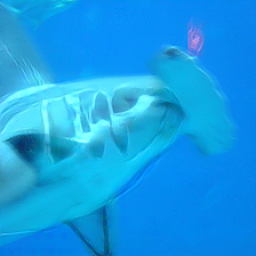}
     \end{subfigure} &

     \begin{subfigure}{00.25\textwidth}\centering
     \caption*{$\rightarrow$w. shark: 0.98}
     \includegraphics[width=1\textwidth]{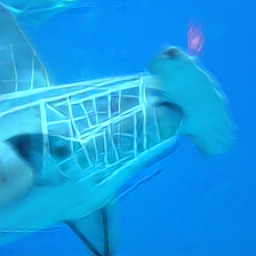}
     \end{subfigure} &

     \begin{subfigure}{00.25\textwidth}\centering
     \caption*{$\rightarrow$w. shark: 1.00}
     \includegraphics[width=1\textwidth]{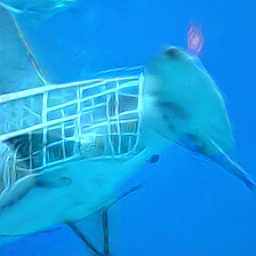}
     \end{subfigure} \\

      \begin{subfigure}{00.25\textwidth}\centering
     \caption*{eel: 0.18}
     \includegraphics[width=1\textwidth]{images/failure_fishes/original.png}
     \end{subfigure} &
     \begin{subfigure}{00.25\textwidth}\centering
     \caption*{$\rightarrow$tench: 0.99}
     \includegraphics[width=1\textwidth]{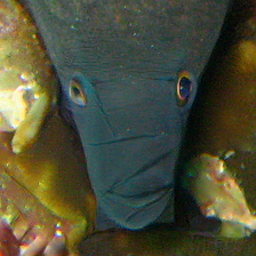}
     \end{subfigure} &

     \begin{subfigure}{00.25\textwidth}\centering
     \caption*{$\rightarrow$tench: 1.00}
    \includegraphics[width=1\textwidth]{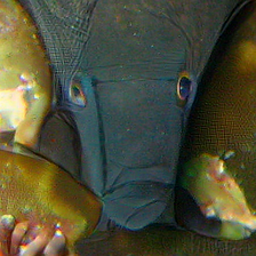}
     \end{subfigure} &

     \begin{subfigure}{00.25\textwidth}\centering
     \caption*{$\rightarrow$tench: 1.00}
     \includegraphics[width=1\textwidth]{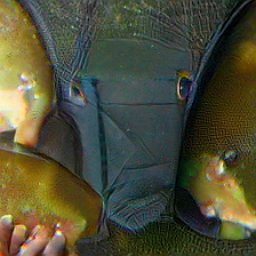}
     \end{subfigure} \\
     \end{tabular}
      \caption{\label{fig:imagent_granny_smith_failure_app}\textbf{Spurious feature of watermarks for the target classes ``granny smith'', ``white shark'', and ``tench''.} $l_{2}$-VCEs for Madry\cite{robustness_short}+FT with varying radii for $3$ images for the target classes ``granny smith'', ``white shark'', and ``tench''. As $l_{1.5}$-VCEs, $l_{2}$-VCEs display spurious features of the classifier.
    }
     \end{figure*}

    \begin{figure*}[hbt!]
   \centering
     \small
     \begin{tabular}{cccc}
     \hline
     \multicolumn{1}{C{0.25\textwidth}}{Original} & \multicolumn{1}{C{0.25\textwidth}}{$\epsilon_{1}=400$} & 
     \multicolumn{1}{C{0.25\textwidth}}{$\epsilon_{1}=600$} &
     \multicolumn{1}{C{0.25\textwidth}}{$\epsilon_{1}=800$}\\
     \hline

     \begin{subfigure}{00.25\textwidth}\centering
     \caption*{granny sm.: 0.92}
     \includegraphics[width=1\textwidth]{images/failure_granny/original.png}
     \end{subfigure} &
     
     \begin{subfigure}{00.25\textwidth}\centering
     \caption*{$\rightarrow$granny sm.: 1.00}
     \includegraphics[width=1\textwidth]{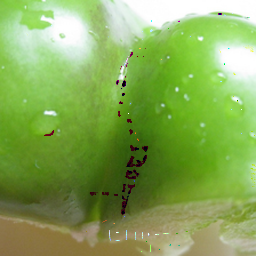}
     \end{subfigure} &

     \begin{subfigure}{00.25\textwidth}\centering
     \caption*{$\rightarrow$granny sm.: 1.00}
     \includegraphics[width=1\textwidth]{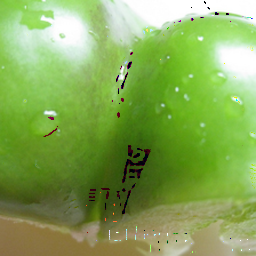}
     \end{subfigure} &

     \begin{subfigure}{00.25\textwidth}\centering
     \caption*{$\rightarrow$granny sm.: 1.00}
     \includegraphics[width=1\textwidth]{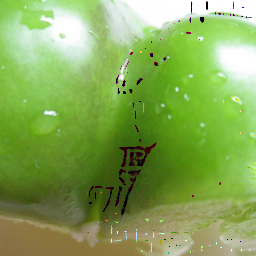}
     
     \end{subfigure} \\

      \begin{subfigure}{00.25\textwidth}\centering
     \caption*{hammerhead: 0.51}
     \includegraphics[width=1\textwidth]{images/failure_sharks_cages/original_re.png}
     \end{subfigure} &
     
     \begin{subfigure}{00.25\textwidth}\centering
     \caption*{$\rightarrow$w. shark: 0.67}
     \includegraphics[width=1\textwidth]{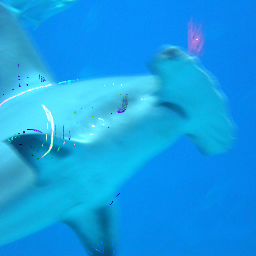}
     \end{subfigure} &

     \begin{subfigure}{00.25\textwidth}\centering
     \caption*{$\rightarrow$w. shark: 0.98}
     \includegraphics[width=1\textwidth]{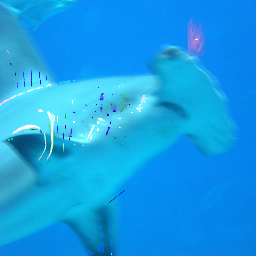}
     \end{subfigure} &

     \begin{subfigure}{00.25\textwidth}\centering
     \caption*{$\rightarrow$w. shark: 1.00}
     \includegraphics[width=1\textwidth]{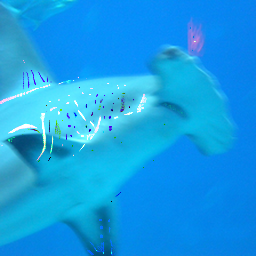}
     \end{subfigure} \\

      \begin{subfigure}{00.25\textwidth}\centering
     \caption*{eel: 0.18}
     \includegraphics[width=1\textwidth]{images/failure_fishes/original.png}
     \end{subfigure} &
     \begin{subfigure}{00.25\textwidth}\centering
     \caption*{$\rightarrow$tench: 0.99}
     \includegraphics[width=1\textwidth]{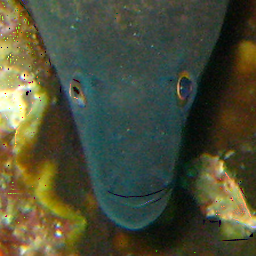}
     \end{subfigure} &

     \begin{subfigure}{00.25\textwidth}\centering
     \caption*{$\rightarrow$tench: 1.00}
    \includegraphics[width=1\textwidth]{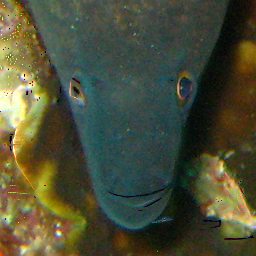}
     \end{subfigure} &

     \begin{subfigure}{00.25\textwidth}\centering
     \caption*{$\rightarrow$tench: 1.00}
     \includegraphics[width=1\textwidth]{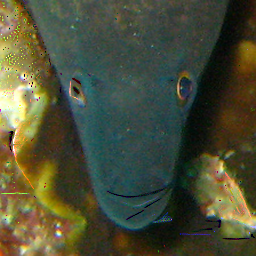}
     \end{subfigure} \\
     \end{tabular}
      \caption{\label{fig:imagent_granny_smith_failure_app}\textbf{Spurious feature of watermarks for the target classes ``granny smith'', ``white shark'', and ``tench''.} $l_{1}$-VCEs for Madry\cite{robustness_short}+FT with varying radii for $3$ images for the target classes ``granny smith'', ``white shark'', and ``tench''. Unlike $l_{1.5}$-VCEs, $l_{1}$-VCEs do not show spurious features of the classifier and have many artefacts.
    }
     \end{figure*}

    \clearpage
    \section{VCEs using guided diffusion process with regularization}\label{app:diffusion-vces}

\begin{figure*}[h]
    \centering
     \small
     \begin{tabular}{c|c|c|c|c|c}
     \hline
     \multicolumn{6}{c}{\textbf{$l_{1.5}$-VCEs for the Madry\cite{robustness_short}+FT model (ours)}} \\
     \hline
     \hline
     \multicolumn{1}{c|}{
     \begin{subfigure}{0.16\textwidth}\centering
     \caption*{Original}
     \includegraphics[width=1\textwidth]{images/teaser_fig/original.png}
     \end{subfigure}}&
     \multicolumn{1}{c}{}
     &
     \multicolumn{1}{c}{
     \begin{subfigure}{0.16\textwidth}\centering
     \caption*{$\epsilon_{1.5}=50$}
     \includegraphics[width=1\textwidth]{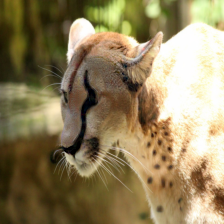}
     \end{subfigure}} &
     \multicolumn{1}{c}{
     \begin{subfigure}{0.16\textwidth}\centering
     \caption*{$\epsilon_{1.5}=75$}
     \includegraphics[width=1\textwidth]{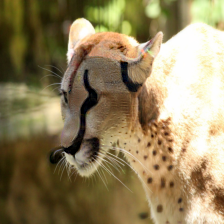}
     \end{subfigure}} &
      \multicolumn{1}{c}{
     \begin{subfigure}{0.16\textwidth}\centering
     \caption*{$\epsilon_{1.5}=100$}
     \includegraphics[width=1\textwidth]{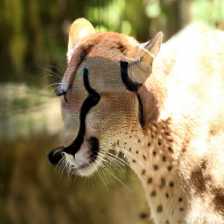}
     \end{subfigure}} & \\
     \hline
     
     \multicolumn{6}{c}{}\\
     \hline
     \multicolumn{6}{c}{\textbf{VCEs using diffusion processes approach of \cite{avrahami2021blended}}} \\
     \hline
     \hline
     &
     \multicolumn{1}{C{.16\textwidth}|}{
     $\lambda_{1.5}$=$0.0125$} & 
     \multicolumn{1}{C{.16\textwidth}|}{$\lambda_{1.5}$=$0.025$} &
     \multicolumn{1}{C{.16\textwidth}|}{$\lambda_{1.5}$=$0.05$}
     & \multicolumn{1}{C{.16\textwidth}|}{
    $\lambda_{1.5}$=$0.1$} & 
     \multicolumn{1}{C{.16\textwidth}}{$\lambda_{1.5}$=$0.2$} \\ 
     \cline{2-6}
     
     \multicolumn{6}{c}{\textbf{Using CLIP model}} \\
     \hline

     $\lambda_g$=$10$ &
     \begin{subfigure}{0.16\textwidth}\centering
     \includegraphics[width=1\textwidth]{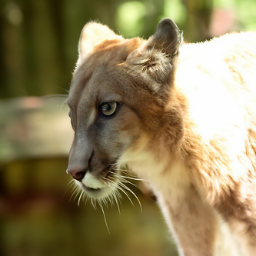}
     \end{subfigure} &
     \begin{subfigure}{0.16\textwidth}\centering
     \includegraphics[width=1\textwidth]{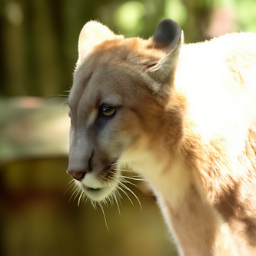}
     \end{subfigure}
     &\begin{subfigure}{0.16\textwidth}\centering
     \includegraphics[width=1\textwidth]{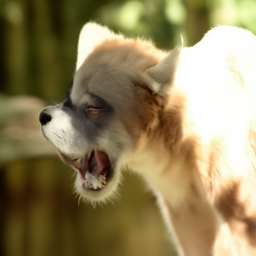}
     \end{subfigure} &
     \begin{subfigure}{0.16\textwidth}\centering
     \includegraphics[width=1\textwidth]{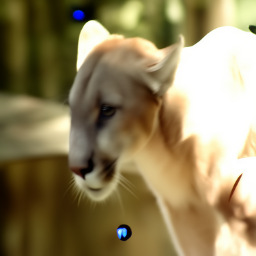}
     \end{subfigure}& 
     \begin{subfigure}{0.16\textwidth}\centering
     \includegraphics[width=1\textwidth]{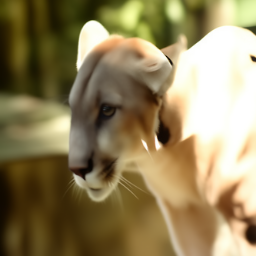}
     \end{subfigure}\\
     \hline
     
     $\lambda_g$=$100$ &
     \begin{subfigure}{0.16\textwidth}\centering
     \includegraphics[width=1\textwidth]{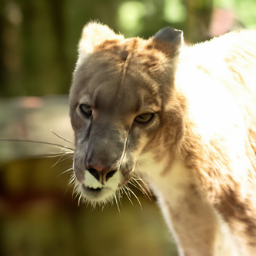}
     \end{subfigure} &
     \begin{subfigure}{0.16\textwidth}\centering
     \includegraphics[width=1\textwidth]{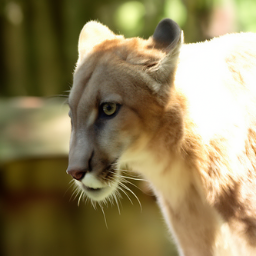}
     \end{subfigure}
     &\begin{subfigure}{0.16\textwidth}\centering
     \includegraphics[width=1\textwidth]{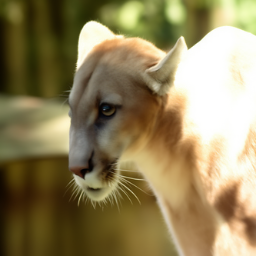}
     \end{subfigure} &
     \begin{subfigure}{0.16\textwidth}\centering
     \includegraphics[width=1\textwidth]{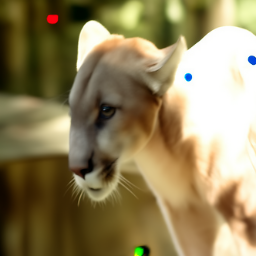}
     \end{subfigure}& 
     \begin{subfigure}{0.16\textwidth}\centering
     \includegraphics[width=1\textwidth]{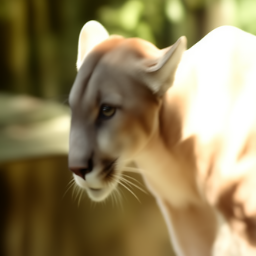}
     \end{subfigure}\\
     \hline
     
     $\lambda_g$=$1000$ &
     \begin{subfigure}{0.16\textwidth}\centering
     \includegraphics[width=1\textwidth]{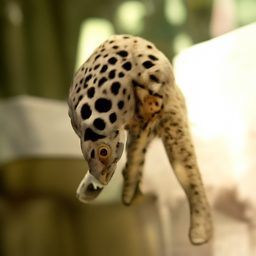}
     \end{subfigure} &
     \begin{subfigure}{0.16\textwidth}\centering
     \includegraphics[width=1\textwidth]{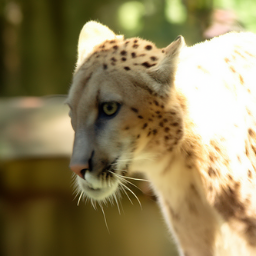}
     \end{subfigure}
     &\begin{subfigure}{0.16\textwidth}\centering
     \includegraphics[width=1\textwidth]{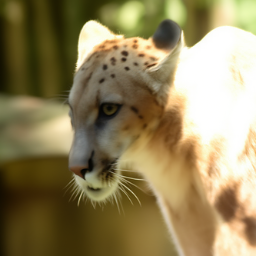}
     \end{subfigure} &
     \begin{subfigure}{0.16\textwidth}\centering
     \includegraphics[width=1\textwidth]{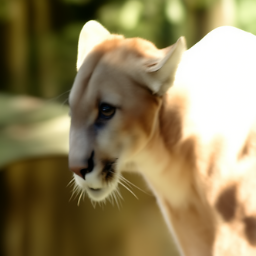}
     \end{subfigure}& 
     \begin{subfigure}{0.16\textwidth}\centering
     \includegraphics[width=1\textwidth]{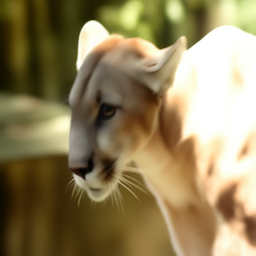}
     \end{subfigure}\\
     
     \hline
     \multicolumn{6}{c}{\textbf{Using Madry\cite{robustness_short}+FT model}} \\
     \hline

     $\lambda_g$=$10$ &
     \begin{subfigure}{0.16\textwidth}\centering
     \includegraphics[width=1\textwidth]{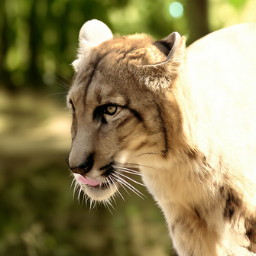}
     \end{subfigure} &
     \begin{subfigure}{0.16\textwidth}\centering
     \includegraphics[width=1\textwidth]{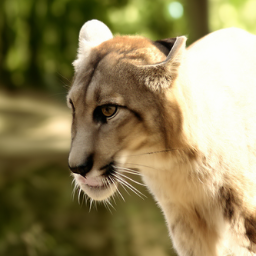}
     \end{subfigure}
     &\begin{subfigure}{0.16\textwidth}\centering
     \includegraphics[width=1\textwidth]{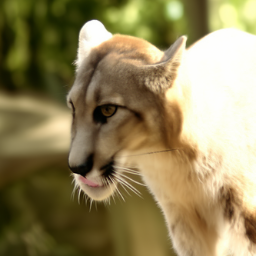}
     \end{subfigure} &
     \begin{subfigure}{0.16\textwidth}\centering
     \includegraphics[width=1\textwidth]{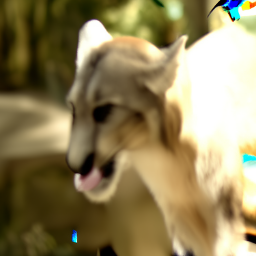}
     \end{subfigure}& 
     \begin{subfigure}{0.16\textwidth}\centering
     \includegraphics[width=1\textwidth]{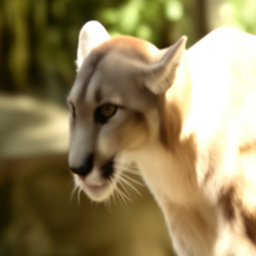}
     \end{subfigure}\\
     \hline
     
     $\lambda_g$=$100$ &
     \begin{subfigure}{0.16\textwidth}\centering
     \includegraphics[width=1\textwidth]{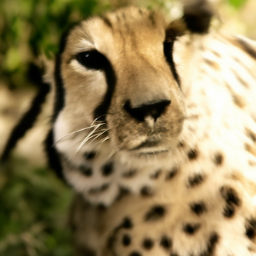}
     \end{subfigure} &
     \begin{subfigure}{0.16\textwidth}\centering
     \includegraphics[width=1\textwidth]{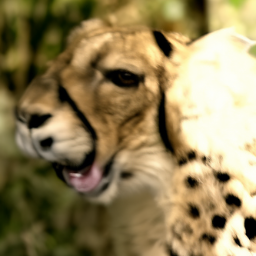}
     \end{subfigure}
     &\begin{subfigure}{0.16\textwidth}\centering
     \includegraphics[width=1\textwidth]{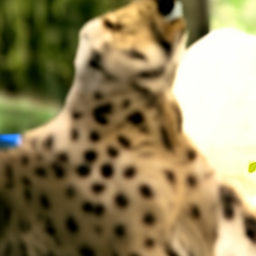}
     \end{subfigure} &
     \begin{subfigure}{0.16\textwidth}\centering
     \includegraphics[width=1\textwidth]{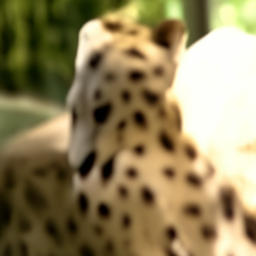}
     \end{subfigure}& 
     \begin{subfigure}{0.16\textwidth}\centering
     \includegraphics[width=1\textwidth]{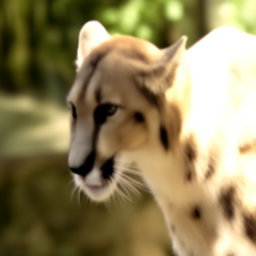}
     \end{subfigure}\\
     \hline
     
     $\lambda_g$=$1000$ &
     \begin{subfigure}{0.16\textwidth}\centering
     \includegraphics[width=1\textwidth]{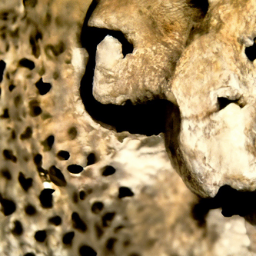}
     \end{subfigure} &
     \begin{subfigure}{0.16\textwidth}\centering
     \includegraphics[width=1\textwidth]{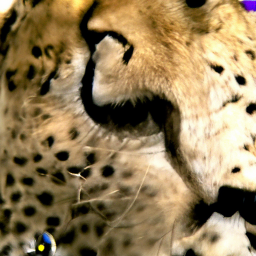}
     \end{subfigure}
     &\begin{subfigure}{0.16\textwidth}\centering
     \includegraphics[width=1\textwidth]{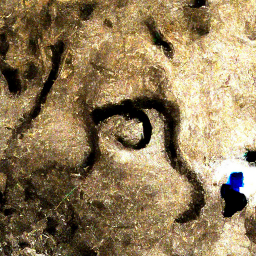}
     \end{subfigure} &
     \begin{subfigure}{0.16\textwidth}\centering
     \includegraphics[width=1\textwidth]{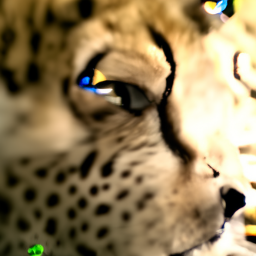}
     \end{subfigure}& 
     \begin{subfigure}{0.16\textwidth}\centering
     \includegraphics[width=1\textwidth]{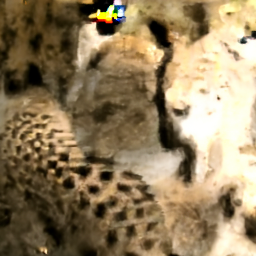}
     \end{subfigure}\\
     \hline
     \end{tabular}
     
      \caption{\label{tab:diffusion-vces} \textbf{$l_{1.5}$-VCEs and VCEs generated using diffusion processes approach of \cite{avrahami2021blended} for CLIP and Madry\cite{robustness_short}+FT model}. Here for VCEs generated using diffusion processes we use $l_{1.5}$ regularization, to have sparser changes for CLIP model used in \cite{avrahami2021blended} and Madry\cite{robustness_short}+FT model. In the columns we vary coefficient for $l_{1.5}$ regularization $\lambda_{1.5}$ and in the rows - coefficient for the CLIP and Madry\cite{robustness_short}+FT guidance $\lambda_g$ for the change \textbf{``cougar $\rightarrow$ cheetah''} for the same seed.
    }
\end{figure*}


\begin{figure*}[h]
    \centering
     \small
     \begin{tabular}{c|c|c|c|c|c}
     \hline
     \multicolumn{6}{c}{\textbf{$l_{1.5}$-VCEs for the Madry\cite{robustness_short}+FT model (ours)}} \\
     \hline
     \hline
     \multicolumn{1}{c|}{
     \begin{subfigure}{0.16\textwidth}\centering
     \caption*{Original}
     \includegraphics[width=1\textwidth]{images/imagenet_valley/original.png}
     \end{subfigure}}&
     \multicolumn{1}{c}{}
     &
     \multicolumn{1}{c}{
     \begin{subfigure}{0.16\textwidth}\centering
     \caption*{$\epsilon_{1.5}=50$}
     \includegraphics[width=1\textwidth]{images/imagenet_valley/target_7_radius_0.png}
     \end{subfigure}} &
     \multicolumn{1}{c}{
     \begin{subfigure}{0.16\textwidth}\centering
     \caption*{$\epsilon_{1.5}=75$}
     \includegraphics[width=1\textwidth]{images/imagenet_valley/target_7_radius_1.png}
     \end{subfigure}} &
      \multicolumn{1}{c}{
     \begin{subfigure}{0.16\textwidth}\centering
     \caption*{$\epsilon_{1.5}=100$}
     \includegraphics[width=1\textwidth]{images/imagenet_valley/target_7_radius_2.png}
     \end{subfigure}} & \\
     \hline
     
     \multicolumn{6}{c}{}\\
     \hline
     \multicolumn{6}{c}{\textbf{VCEs using diffusion processes approach of \cite{avrahami2021blended}}} \\
     \hline
     \hline
     &
     \multicolumn{1}{C{.16\textwidth}|}{
     $\lambda_{1.5}$=$0.0125$} & 
     \multicolumn{1}{C{.16\textwidth}|}{$\lambda_{1.5}$=$0.025$} &
     \multicolumn{1}{C{.16\textwidth}|}{$\lambda_{1.5}$=$0.05$}
     & \multicolumn{1}{C{.16\textwidth}|}{
    $\lambda_{1.5}$=$0.1$} & 
     \multicolumn{1}{C{.16\textwidth}}{$\lambda_{1.5}$=$0.2$} \\ 
     \cline{2-6}
     
     \multicolumn{6}{c}{\textbf{Using CLIP model}} \\
     \hline

     $\lambda_g$=$10$ &
     \begin{subfigure}{0.16\textwidth}\centering
     \includegraphics[width=1\textwidth]{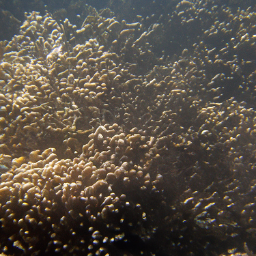}
     \end{subfigure} &
     \begin{subfigure}{0.16\textwidth}\centering
     \includegraphics[width=1\textwidth]{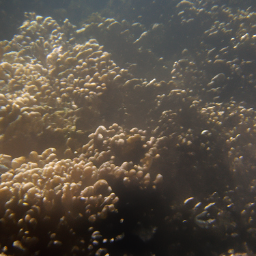}
     \end{subfigure}
     &\begin{subfigure}{0.16\textwidth}\centering
     \includegraphics[width=1\textwidth]{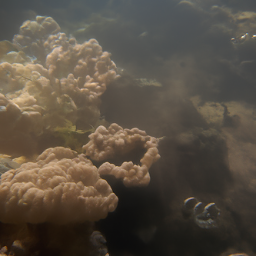}
     \end{subfigure} &
     \begin{subfigure}{0.16\textwidth}\centering
     \includegraphics[width=1\textwidth]{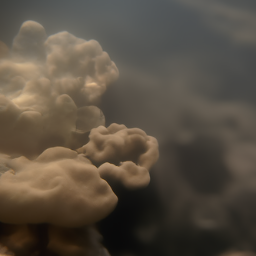}
     \end{subfigure}& 
     \begin{subfigure}{0.16\textwidth}\centering
     \includegraphics[width=1\textwidth]{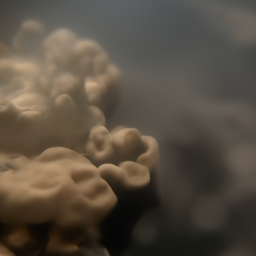}
     \end{subfigure}\\
     \hline
     
     $\lambda_g$=$100$ &
     \begin{subfigure}{0.16\textwidth}\centering
     \includegraphics[width=1\textwidth]{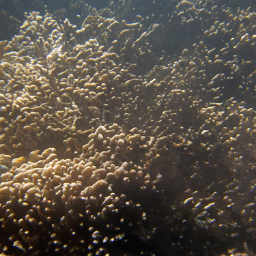}
     \end{subfigure} &
     \begin{subfigure}{0.16\textwidth}\centering
     \includegraphics[width=1\textwidth]{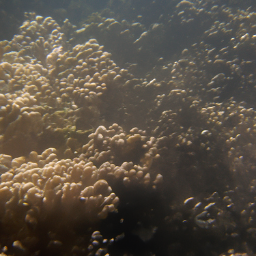}
     \end{subfigure}
     &\begin{subfigure}{0.16\textwidth}\centering
     \includegraphics[width=1\textwidth]{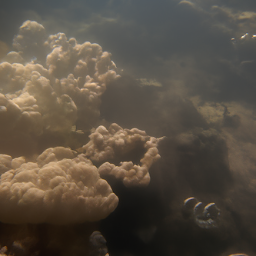}
     \end{subfigure} &
     \begin{subfigure}{0.16\textwidth}\centering
     \includegraphics[width=1\textwidth]{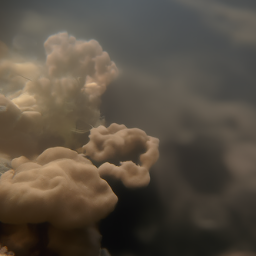}
     \end{subfigure}& 
     \begin{subfigure}{0.16\textwidth}\centering
     \includegraphics[width=1\textwidth]{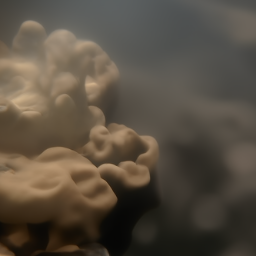}
     \end{subfigure}\\
     \hline
     
     $\lambda_g$=$1000$ &
     \begin{subfigure}{0.16\textwidth}\centering
     \includegraphics[width=1\textwidth]{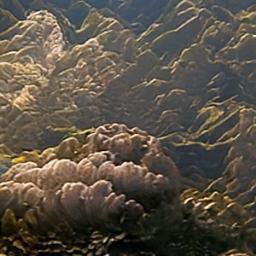}
     \end{subfigure} &
     \begin{subfigure}{0.16\textwidth}\centering
     \includegraphics[width=1\textwidth]{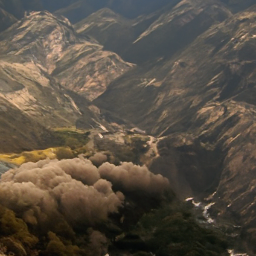}
     \end{subfigure}
     &\begin{subfigure}{0.16\textwidth}\centering
     \includegraphics[width=1\textwidth]{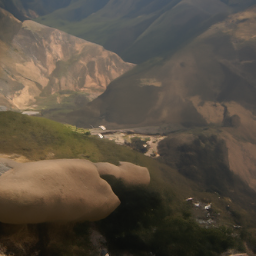}
     \end{subfigure} &
     \begin{subfigure}{0.16\textwidth}\centering
     \includegraphics[width=1\textwidth]{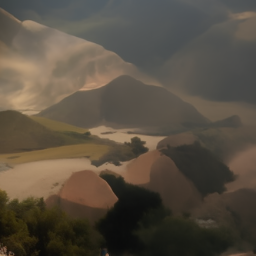}
     \end{subfigure}& 
     \begin{subfigure}{0.16\textwidth}\centering
     \includegraphics[width=1\textwidth]{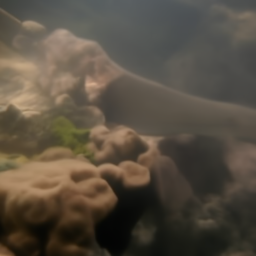}
     \end{subfigure}\\
     \hline
     \multicolumn{6}{c}{\textbf{Using Madry\cite{robustness_short}+FT model}} \\
     \hline
     
      $\lambda_g$=$10$ &
     \begin{subfigure}{0.16\textwidth}\centering
     \includegraphics[width=1\textwidth]{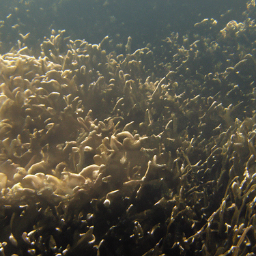}
     \end{subfigure} &
     \begin{subfigure}{0.16\textwidth}\centering
     \includegraphics[width=1\textwidth]{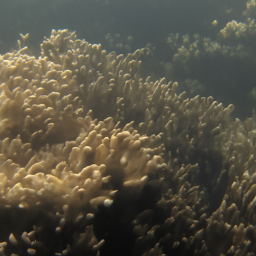}
     \end{subfigure}
     &\begin{subfigure}{0.16\textwidth}\centering
     \includegraphics[width=1\textwidth]{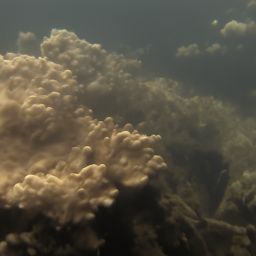}
     \end{subfigure} &
     \begin{subfigure}{0.16\textwidth}\centering
     \includegraphics[width=1\textwidth]{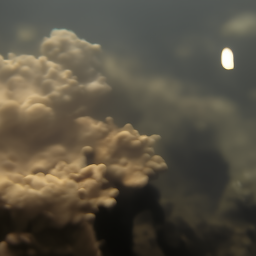}
     \end{subfigure}& 
     \begin{subfigure}{0.16\textwidth}\centering
     \includegraphics[width=1\textwidth]{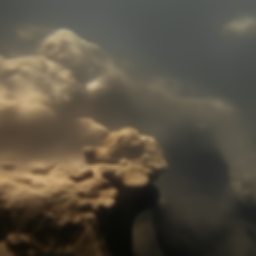}
     \end{subfigure}\\
     \hline
     
     $\lambda_g$=$100$ &
     \begin{subfigure}{0.16\textwidth}\centering
     \includegraphics[width=1\textwidth]{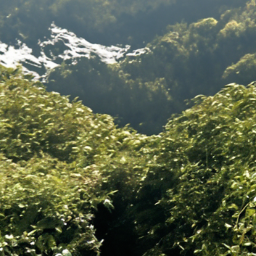}
     \end{subfigure} &
     \begin{subfigure}{0.16\textwidth}\centering
     \includegraphics[width=1\textwidth]{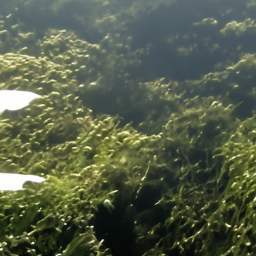}
     \end{subfigure}
     &\begin{subfigure}{0.16\textwidth}\centering
     \includegraphics[width=1\textwidth]{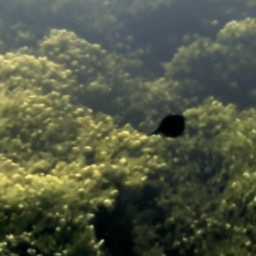}
     \end{subfigure} &
     \begin{subfigure}{0.16\textwidth}\centering
     \includegraphics[width=1\textwidth]{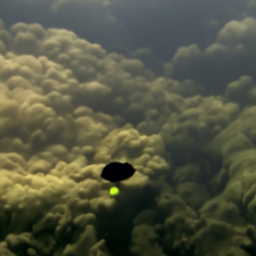}
     \end{subfigure}& 
     \begin{subfigure}{0.16\textwidth}\centering
     \includegraphics[width=1\textwidth]{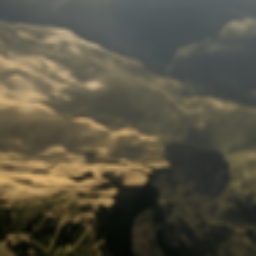}
     \end{subfigure}\\
     \hline
     
     $\lambda_g$=$1000$ &
     \begin{subfigure}{0.16\textwidth}\centering
     \includegraphics[width=1\textwidth]{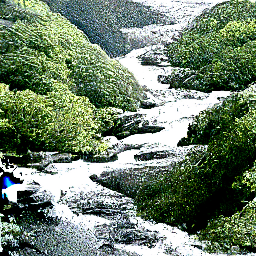}
     \end{subfigure} &
     \begin{subfigure}{0.16\textwidth}\centering
     \includegraphics[width=1\textwidth]{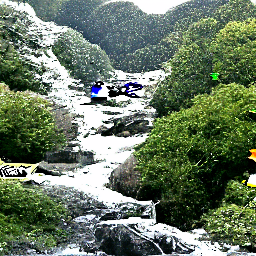}
     \end{subfigure}
     &\begin{subfigure}{0.16\textwidth}\centering
     \includegraphics[width=1\textwidth]{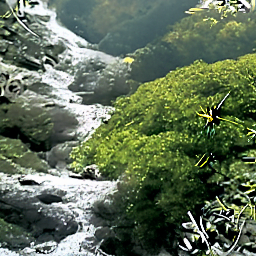}
     \end{subfigure} &
     \begin{subfigure}{0.16\textwidth}\centering
     \includegraphics[width=1\textwidth]{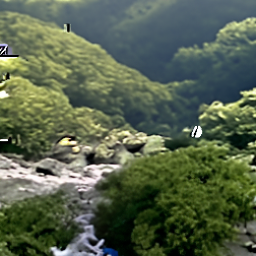}
     \end{subfigure}& 
     \begin{subfigure}{0.16\textwidth}\centering
     \includegraphics[width=1\textwidth]{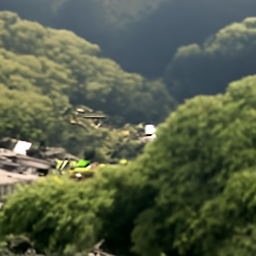}
     \end{subfigure}\\
     \hline
     
     \end{tabular}
     
      \caption{\label{tab:diffusion-vces} \textbf{$l_{1.5}$-VCEs and VCEs generated using diffusion processes approach of \cite{avrahami2021blended} for CLIP and Madry\cite{robustness_short}+FT model}. Here for VCEs generated using diffusion processes we use $l_{1.5}$ regularization, to have sparser changes for CLIP model used in \cite{avrahami2021blended} and Madry\cite{robustness_short}+FT model. In the columns we vary coefficient for $l_{1.5}$ regularization $\lambda_{1.5}$ and in the rows - coefficient for the CLIP and Madry\cite{robustness_short}+FT guidance $\lambda_g$ for the change \textbf{``coral reef $\rightarrow$ valley''} for the same seed.
    }
\end{figure*}


\begin{figure*}[h]
    \centering
     \small
     \begin{tabular}{c|c|c|c|c|c}
     \hline
    \multicolumn{6}{c}{\textbf{$l_{1.5}$-VCEs for the Madry\cite{robustness_short}+FT model (ours)}} \\
     \hline
     \hline
     
     \multicolumn{1}{c|}{
     \begin{subfigure}{0.16\textwidth}\centering
     \caption*{Original}
     \includegraphics[width=1\textwidth]{images/imagenet_appendix_examples/edible_fruit_fig/original.png}
     \end{subfigure}}&
     \multicolumn{1}{c}{}
     &
     \multicolumn{1}{c}{
     \begin{subfigure}{0.16\textwidth}\centering
     \caption*{$\epsilon_{1.5}=50$}
     \includegraphics[width=1\textwidth]{images/imagenet_appendix_examples/edible_fruit_fig/target_4_radius_0.png}
     \end{subfigure}} &
    
     \multicolumn{1}{c}{
     \begin{subfigure}{0.16\textwidth}\centering
     \caption*{$\epsilon_{1.5}=75$}
     \includegraphics[width=1\textwidth]{images/imagenet_appendix_examples/edible_fruit_fig/target_4_radius_1.png}
     \end{subfigure}} &
    
     \multicolumn{1}{c}{
     \begin{subfigure}{0.16\textwidth}\centering
     \caption*{$\epsilon_{1.5}=100$}
     \includegraphics[width=1\textwidth]{images/imagenet_appendix_examples/edible_fruit_fig/target_4_radius_2.png}
     \end{subfigure}}&\multicolumn{1}{c}{} \\
     
     \hline
     
     \multicolumn{6}{c}{}\\
     \hline
     \multicolumn{6}{c}{\textbf{VCEs using diffusion processes approach of \cite{avrahami2021blended}}} \\
     \hline
     \hline
     
     &
     \multicolumn{1}{C{.16\textwidth}|}{
     $\lambda_{1.5}$=$0.0125$} & 
     \multicolumn{1}{C{.16\textwidth}|}{$\lambda_{1.5}$=$0.025$} &
     \multicolumn{1}{C{.16\textwidth}|}{$\lambda_{1.5}$=$0.05$}
     & \multicolumn{1}{C{.16\textwidth}|}{
    $\lambda_{1.5}$=$0.1$} & 
     \multicolumn{1}{C{.16\textwidth}}{$\lambda_{1.5}$=$0.2$} \\ 
     \cline{2-6}
     \multicolumn{6}{c}{\textbf{Using CLIP model}} \\
     \hline

     $\lambda_g$=$10$ &
     \begin{subfigure}{0.16\textwidth}\centering
     \includegraphics[width=1\textwidth]{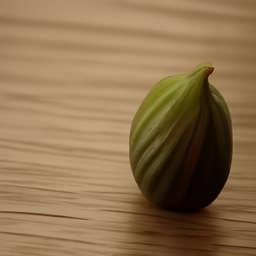}
     \end{subfigure} &
     \begin{subfigure}{0.16\textwidth}\centering
     \includegraphics[width=1\textwidth]{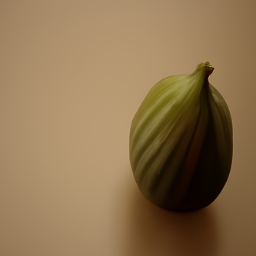}
     \end{subfigure}
     &\begin{subfigure}{0.16\textwidth}\centering
     \includegraphics[width=1\textwidth]{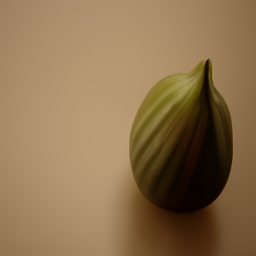}
     \end{subfigure} &
     \begin{subfigure}{0.16\textwidth}\centering
     \includegraphics[width=1\textwidth]{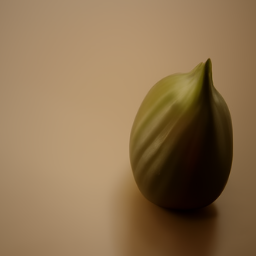}
     \end{subfigure}& 
     \begin{subfigure}{0.16\textwidth}\centering
     \includegraphics[width=1\textwidth]{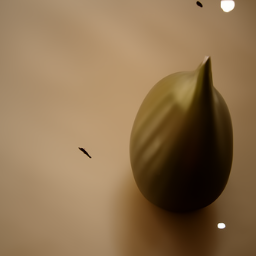}
     \end{subfigure}\\
     \hline
     
     $\lambda_g$=$100$ &
     \begin{subfigure}{0.16\textwidth}\centering
     \includegraphics[width=1\textwidth]{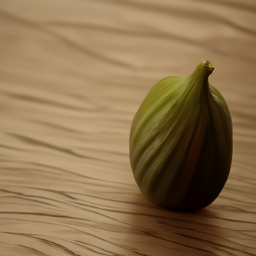}
     \end{subfigure} &
     \begin{subfigure}{0.16\textwidth}\centering
     \includegraphics[width=1\textwidth]{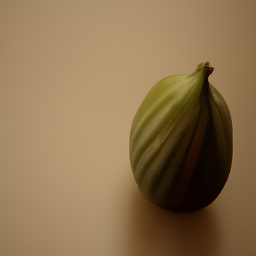}
     \end{subfigure}
     &\begin{subfigure}{0.16\textwidth}\centering
     \includegraphics[width=1\textwidth]{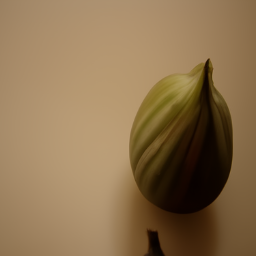}
     \end{subfigure} &
     \begin{subfigure}{0.16\textwidth}\centering
     \includegraphics[width=1\textwidth]{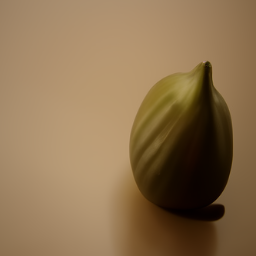}
     \end{subfigure}& 
     \begin{subfigure}{0.16\textwidth}\centering
     \includegraphics[width=1\textwidth]{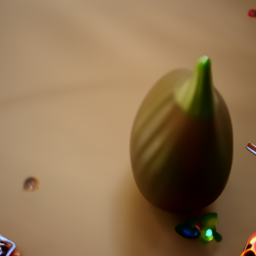}
     \end{subfigure}\\
     \hline
     
     $\lambda_g$=$1000$ &
     \begin{subfigure}{0.16\textwidth}\centering
     \includegraphics[width=1\textwidth]{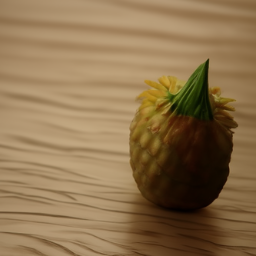}
     \end{subfigure} &
     \begin{subfigure}{0.16\textwidth}\centering
     \includegraphics[width=1\textwidth]{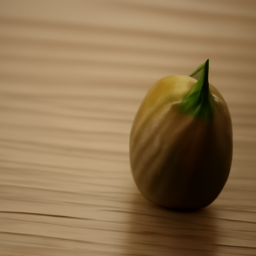}
     \end{subfigure}
     &\begin{subfigure}{0.16\textwidth}\centering
     \includegraphics[width=1\textwidth]{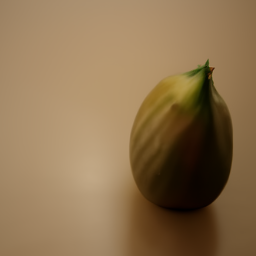}
     \end{subfigure} &
     \begin{subfigure}{0.16\textwidth}\centering
     \includegraphics[width=1\textwidth]{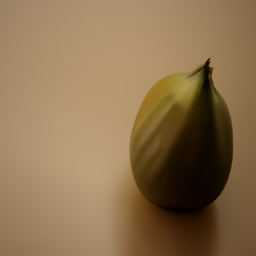}
     \end{subfigure}& 
     \begin{subfigure}{0.16\textwidth}\centering
     \includegraphics[width=1\textwidth]{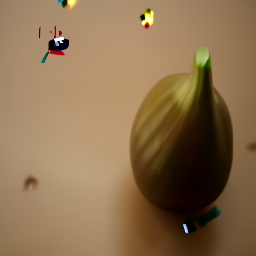}
     \end{subfigure}\\
     \hline
     \multicolumn{6}{c}{\textbf{Using Madry\cite{robustness_short}+FT model}} \\
     \hline
      $\lambda_g$=$10$ &
     \begin{subfigure}{0.16\textwidth}\centering
     \includegraphics[width=1\textwidth]{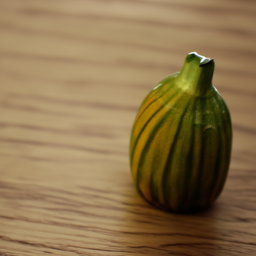}
     \end{subfigure} &
     \begin{subfigure}{0.16\textwidth}\centering
     \includegraphics[width=1\textwidth]{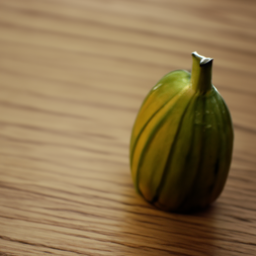}
     \end{subfigure}
     &\begin{subfigure}{0.16\textwidth}\centering
     \includegraphics[width=1\textwidth]{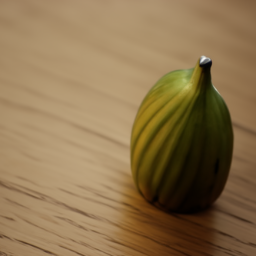}
     \end{subfigure} &
     \begin{subfigure}{0.16\textwidth}\centering
     \includegraphics[width=1\textwidth]{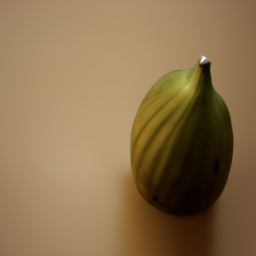}
     \end{subfigure}& 
     \begin{subfigure}{0.16\textwidth}\centering
     \end{subfigure}\\
     \hline
     
     $\lambda_g$=$100$ &
     \begin{subfigure}{0.16\textwidth}\centering
     \includegraphics[width=1\textwidth]{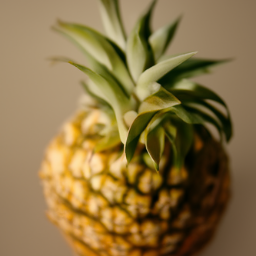}
     \end{subfigure} &
     \begin{subfigure}{0.16\textwidth}\centering
     \includegraphics[width=1\textwidth]{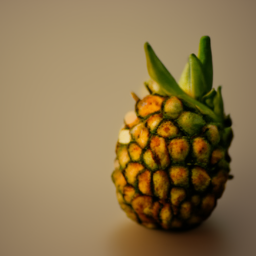}
     \end{subfigure}
     &\begin{subfigure}{0.16\textwidth}\centering
     \includegraphics[width=1\textwidth]{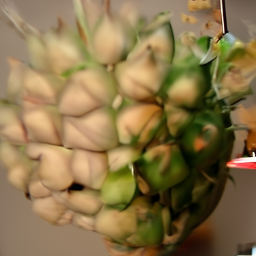}
     \end{subfigure} &
     \begin{subfigure}{0.16\textwidth}\centering
     \includegraphics[width=1\textwidth]{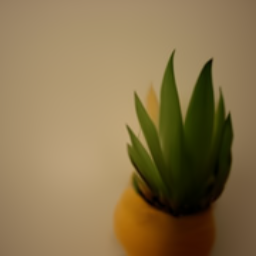}
     \end{subfigure}& 
     \begin{subfigure}{0.16\textwidth}\centering
     \includegraphics[width=1\textwidth]{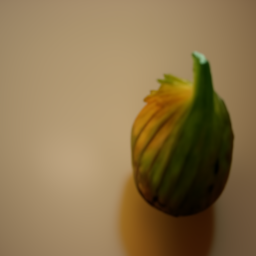}
     \end{subfigure}\\
     \hline
     
     $\lambda_g$=$1000$ &
     \begin{subfigure}{0.16\textwidth}\centering
     \includegraphics[width=1\textwidth]{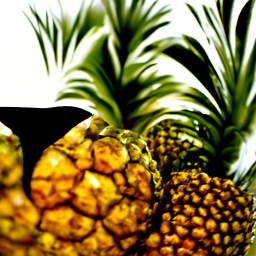}
     \end{subfigure} &
     \begin{subfigure}{0.16\textwidth}\centering
     \includegraphics[width=1\textwidth]{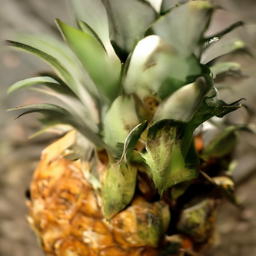}
     \end{subfigure}
     &\begin{subfigure}{0.16\textwidth}\centering
     \includegraphics[width=1\textwidth]{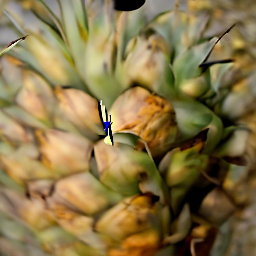}
     \end{subfigure} &
     \begin{subfigure}{0.16\textwidth}\centering
     \includegraphics[width=1\textwidth]{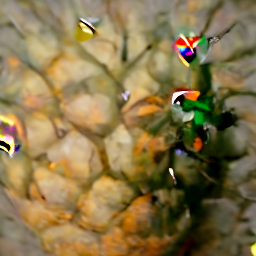}
     \end{subfigure}& 
     \begin{subfigure}{0.16\textwidth}\centering
     \includegraphics[width=1\textwidth]{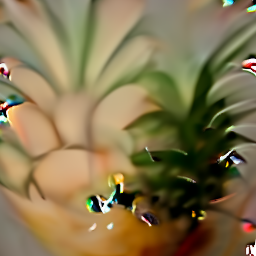}
     \end{subfigure}\\
     \hline
     
     \end{tabular}
     
      \caption{\label{tab:diffusion-vces} \textbf{$l_{1.5}$-VCEs and VCEs generated using diffusion processes approach of \cite{avrahami2021blended} for CLIP and Madry\cite{robustness_short}+FT model}. Here for VCEs generated using diffusion processes we use $l_{1.5}$ regularization, to have sparser changes for CLIP model used in \cite{avrahami2021blended} and Madry\cite{robustness_short}+FT model. In the columns we vary coefficient for $l_{1.5}$ regularization $\lambda_{1.5}$ and in the rows - coefficient for the CLIP and Madry\cite{robustness_short}+FT guidance $\lambda_g$ for the change \textbf{``fig $\rightarrow$ pineapple''} for the same seed. The empty square means that for this seed and this combination the method has encountered a numerical instability.
    }
\end{figure*}

    Diffusion processes \cite{dhariwal2021diffusion} have been used to generate VCEs on natural images when a causal structure is available \cite{sanchez2022diffusion}. We extend such approach to generic classifiers by leveraging the method of \cite{avrahami2021blended}: they show that using a diffusion process, guided by the CLIP model \cite{radford2021learning}, with regularization is effective in a variety of text-driven image generation tasks like object or background replacement, object editing \cite{nichol2021glide}. In particular, in \cite{avrahami2021blended}, sampling exploits a hand-crafted mask, and is guided by image-text consistency according to the CLIP model and an $l_2$-regularization term which ensures consistency with the background (the strength of such terms is controlled by the parameters $\lambda_g$ and $\lambda_2$ respectively).
    
    To apply such approach to our setting, we modify the following components:

    (i) we use $l_{1.5}$ regularization instead of $l_2$ to have sparser changes and to be consistent with our proposed approach, (ii) apply regularization on the whole image, while \cite{avrahami2021blended} did so outside the mask only, and (iii) do not use the mask and operate on the whole image.
    Moreover, to generate VCEs for arbitrary classifiers which do not take text prompt as CLIP, we guide the sampling process by setting
    \begin{equation}
        \mathcal{L} \leftarrow \lambda_g \log \hat{p}(k|\widehat{x}_0) - \lambda_{1.5} \norm{\widehat{x}_0-x}_{1.5}^{1.5} - \lambda_{\text{LPIPS}} \text{LPIPS}(\widehat{x}_0, x) ,
    \end{equation}
    
    as loss function in \textbf{Algorithm 1} of \cite{avrahami2021blended}, where $k$ represents the target class, $\widehat{x}_0$ the output of the denoising step and $x$ the original image (following the notation of 
    \cite{avrahami2021blended}).
    
    In \cref{tab:diffusion-vces} we show the VCEs obtained with this scheme for two images (which we already used in \cref{fig:motivating_example} and \cref{tab:imagent_valley}). In particular we test several combination parameters and repeat for two random seeds. For convenience of the reader for direct comparison to our $l_{1.5}$-VCEs we show again the corresponding images of the main paper. 
    
    We can see that (i) even though for some seeds and images the resulting VCEs can look realistic, they are perceivable less sparse than $l_{1.5}$-VCEs, oftentimes changing the image completely, or are not valid, in the sense that no features of the target class appear (ii) parameters are more difficult to tune as 
    $\lambda_{1.5}$ and $\lambda_g$ are both model but even worse also image dependent, e.g. for the Madry+FT model $\lambda_g=100$ and $\lambda_{1.5}=0.2$ works best for the ``cougar$\rightarrow$ cheetah''-VCE but does not yield any meaningful VCE for ``fig$\rightarrow$ pineapple'' where now $\lambda_{1.5}=0.025$ works best and for ``coral reef $\rightarrow$ valley'' none of these parameters works well, and are thus much more difficult to control than the radius of our $l_p$-VCEs, and (iii) as their  algorithm works not in the image space, and only projects to $[0, 1]^d$ in the end, some images have visible artefacts (black or red dots). Moreover, when zooming in one notices that their generated images are often slightly blurred. In particular, we would argue that our $l_{1.5}$-VCE (shown at the top) induce much more subtle changes of the original image clearly visible for ``fig $\rightarrow$ pineapple'', where their VCEs do not preserve the background. 
    
    \clearpage
    \section{User study examples}\label{app:user-study}
    \begin{figure*}[hbt!]
     
     \centering
     \begin{tabular}{c|c|c|c}
                   \hline 
              \multicolumn{1}{c}{Original}&\multicolumn{1}{C{.27\textwidth}}{APGD, $l_{1}$}&\multicolumn{1}{C{.27\textwidth}}{AFW, $l_{1.5}$}&\multicolumn{1}{C{.27\textwidth}}{APGD, $l_{2}$}\\
                   \hline
     
     \begin{subfigure}{0.14\textwidth}\centering
     
     \caption*{{\tiny promontory:$0.60$}}
     \includegraphics[width=.96\textwidth]{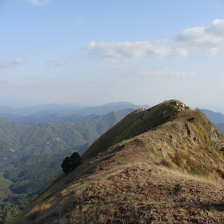}
     \end{subfigure}&
     \begin{subfigure}{0.28\textwidth}\centering

     \caption*{{\tiny r.:$0.00$, m.:$0.56$, s.:$0.39$}}
     \includegraphics[width=.47\textwidth]{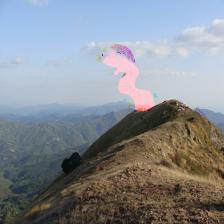}
     \includegraphics[width=.47\textwidth]{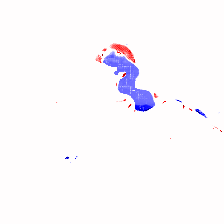}
     
     \end{subfigure}&\begin{subfigure}{0.28\textwidth}\centering
     
     \caption*{{\tiny r.:$0.72$, m.:$1.00$, s.:$0.78$}}
     \includegraphics[width=.47\textwidth]{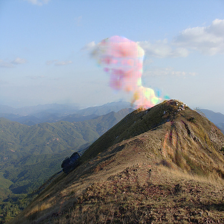}
     \includegraphics[width=.47\textwidth]{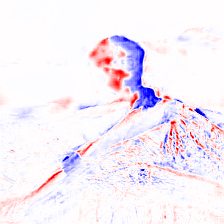}
     
     \end{subfigure} 
     &\begin{subfigure}{0.28\textwidth}\centering

     \caption*{{\tiny r.:$0.94$, m.:$0.94$, s.:$0.83$}}
     \includegraphics[width=.47\textwidth,cfbox=blue 1pt 0pt]{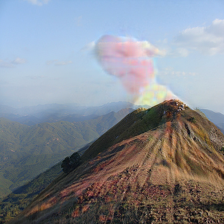}
     \includegraphics[width=.47\textwidth]{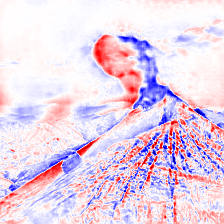}

     \end{subfigure}\\ 
     \hline
     \begin{subfigure}{0.14\textwidth}\centering
     
     \caption*{{\tiny hummingb.:$0.35$}}
     \includegraphics[width=.96\textwidth]{images/user_study_examples/original4.png}
     \end{subfigure}&
     \begin{subfigure}{0.28\textwidth}\centering

     \caption*{{\tiny r.:$0.61$, m.:$0.72$, s.:$0.61$}}
     \includegraphics[width=.47\textwidth]{images/user_study_examples/l1_target_0_radius_0.png}
     \includegraphics[width=.47\textwidth]{images/user_study_examples/l1_target_0_radius_0_diff.png}
     
     \end{subfigure}&\begin{subfigure}{0.28\textwidth}\centering
     
     \caption*{{\tiny r.:$0.89$, m.:$0.89$, s.:$0.72$}}
     \includegraphics[width=.47\textwidth,cfbox=blue 1pt 0pt]{images/user_study_examples/l1_5_target_0_radius_01.png}
     \includegraphics[width=.47\textwidth]{images/user_study_examples/l1_5_target_0_radius_0_diff.png}
     
     \end{subfigure} 
     &\begin{subfigure}{0.28\textwidth}\centering

     \caption*{{\tiny r.:$0.78$, m.:$0.78$, s.:$0.61$}}
     \includegraphics[width=.47\textwidth]{images/user_study_examples/l2_target_0_radius_0.png}
     \includegraphics[width=.47\textwidth]{images/user_study_examples/l2_target_0_radius_0_diff.png}
     
     \end{subfigure}\\

      \end{tabular}
    
     \caption{\label{fig:user-study-best} \textbf{Best rated images from the user study in blue.} $l_p$-VCEs for $p \in \{1,1.5,2\}$ for the change ``promontory $\longrightarrow$ volcano'' (top row) and ``hummingbird $\longrightarrow$ brambling'' (bottom) for Madry\cite{robustness_short}+FT, where the images with blue frame are those for which users answered yes at the same time on \textbf{realism (r)}, \textbf{meaningful (m)}, \textbf{subtle (s)} questions introduced in \cref{app:user-study} most frequently. 
     For each VCE we show the difference to the original image 
     and proportions how often each of the three questions is answered with yes
     .}

     \end{figure*}

     \begin{figure*}[hbt!]
     
     \centering
     \begin{tabular}{c|c|c|c}
                   \hline 
              \multicolumn{1}{c}{Original}&\multicolumn{1}{C{.27\textwidth}}{APGD, $l_{1}$}&\multicolumn{1}{C{.27\textwidth}}{AFW, $l_{1.5}$}&\multicolumn{1}{C{.27\textwidth}}{APGD, $l_{2}$}\\
                   \hline
     
     \begin{subfigure}{0.14\textwidth}\centering
     
     \caption*{{\tiny paddle:$0.11$}}
     \includegraphics[width=.96\textwidth]{images/user_study_examples/original5.png}
     \end{subfigure}&
     \begin{subfigure}{0.28\textwidth}\centering

     \caption*{{\tiny r.:$0.06$, m.:$0.17$, s.:$0.06$}}
     \includegraphics[width=.47\textwidth,cfbox=red 1pt 0pt]{images/user_study_examples/l1_target_6_radius_0.png}
     \includegraphics[width=.47\textwidth]{images/user_study_examples/l1_target_6_radius_0_diff.png}
     
     \end{subfigure}&\begin{subfigure}{0.28\textwidth}\centering
     
     \caption*{{\tiny r.:$0.28$, m.:$0.11$, s.:$0.44$}}
     \includegraphics[width=.47\textwidth]{images/user_study_examples/l1_5_target_6_radius_0.png}
     \includegraphics[width=.47\textwidth]{images/user_study_examples/l1_5_target_6_radius_0_diff.png}
     
     \end{subfigure} 
     &\begin{subfigure}{0.28\textwidth}\centering

     \caption*{{\tiny r.:$0.39$, m.:$0.11$, s.:$0.5$}}
     \includegraphics[width=.47\textwidth]{images/user_study_examples/l2_target_6_radius_0.png}
     \includegraphics[width=.47\textwidth]{images/user_study_examples/l2_target_6_radius_0_diff.png}

     \end{subfigure}\\ 
     \hline
     \begin{subfigure}{0.14\textwidth}\centering
     
     \caption*{{\tiny lakeside:$0.17$}}
     \includegraphics[width=.96\textwidth]{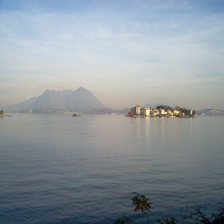}
     \end{subfigure}&
     \begin{subfigure}{0.28\textwidth}\centering

     \caption*{{\tiny r.:$0.06$, m.:$0.50$, s.:$0.44$}}
     \includegraphics[width=.47\textwidth]{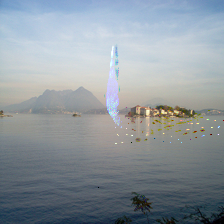}
     \includegraphics[width=.47\textwidth]{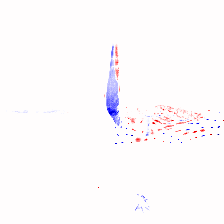}
     
     \end{subfigure}&\begin{subfigure}{0.28\textwidth}\centering
     
     \caption*{{\tiny r.:$0.28$, m.:$0.94$, s.:$0.17$}}
     \includegraphics[width=.47\textwidth]{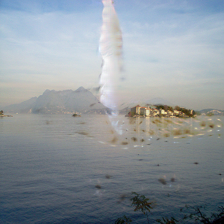}
     \includegraphics[width=.47\textwidth]{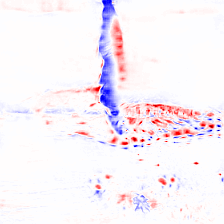}
     
     \end{subfigure} 
     &\begin{subfigure}{0.28\textwidth}\centering

     \caption*{{\tiny r.:$0.06$, m.:$0.00$, s.:$0.11$}}
     \includegraphics[width=.47\textwidth,cfbox=red 1pt 0pt]{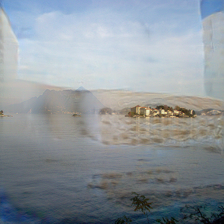}
     \includegraphics[width=.47\textwidth]{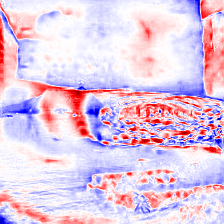}
     
     \end{subfigure}\\

      \end{tabular}
    
     \caption{\label{fig:user-study-worst} \textbf{Worst rated images from the user study in red.} $l_p$-VCEs for $p \in \{1,1.5,2\}$ for the change ``paddle $\longrightarrow$ bearskin'' (top row) and ``lakeside $\longrightarrow$ geyser'' (bottom) for Madry\cite{robustness_short}+FT, where the images with red frame are those for which users have answered least frequently yes to the \textbf{realism (r)}, \textbf{meaningful (m)}, nor \textbf{subtle (s)} questions introduced in \cref{app:user-study}. 
     For each VCE we show the difference to the original image 
     and proportions how often each of the three questions is answered with yes
     .}

     \end{figure*}
    Here we provide two examples for the $l_p$-VCEs ($p \in \{1,1.5,2\}$) that according to the user study \cref{sec:threatmodel} were the best (users have answered yes to all three questions most frequently) in \cref{fig:user-study-best} and two that were the worst  (users have answered yes to all three questions least frequently) in \cref{fig:user-study-worst}. The two worst examples show qualitatively why $l_1$- and $l_2$-VCEs might lead to undesired behaviors. In fact, they introduce changes which are either too localized and with intense colors ($l_1$-VCEs) or cover the entire image ($l_2$-VCEs).

    \section{Randomly selected VCEs}\label{app:random-selection}
\begin{figure*}[htb!]
     \centering

    \begin{tabular}{c|ccc|c|ccc}
     \hline
     \multicolumn{1}{c|}{Original} & \multicolumn{1}{C{.12\textwidth}}{$\epsilon_{1.5}$=$50$} & 
     \multicolumn{1}{C{.12\textwidth}}{$\epsilon_{1.5}$=$75$} &
     \multicolumn{1}{C{.12\textwidth}|}{$\epsilon_{1.5}$=$100$} &
     \multicolumn{1}{c|}{Original} & \multicolumn{1}{C{.12\textwidth}}{$\epsilon_{1.5}$=$50$} & 
     \multicolumn{1}{C{.12\textwidth}}{$\epsilon_{1.5}$=$75$} &
     \multicolumn{1}{C{.12\textwidth}|}{$\epsilon_{1.5}$=$100$}\\
     
     \hline
    
    \begin{subfigure}{0.12\textwidth}\centering
    \caption*{\tiny magnetic compass:\\ 0.58}
    \includegraphics[width=1\textwidth]{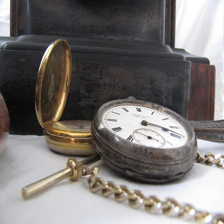}
    \end{subfigure} &
    \begin{subfigure}{0.12\textwidth}\centering
    \caption*{\tiny $\rightarrow$ digital watch: \\0.15}
    \includegraphics[width=1\textwidth]{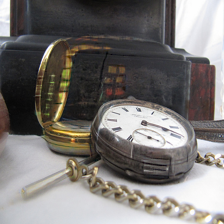}
    \end{subfigure} &
    \begin{subfigure}{0.12\textwidth}\centering
    \caption*{\tiny $\rightarrow$ digital watch: \\0.84}
    \includegraphics[width=1\textwidth]{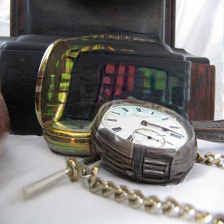}
    \end{subfigure} &
    \begin{subfigure}{0.12\textwidth}\centering
    \caption*{\tiny $\rightarrow$ digital watch: \\0.99}
    \includegraphics[width=1\textwidth]{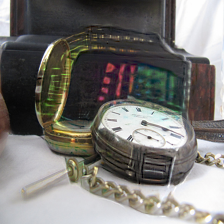}
    \end{subfigure} &
    
    \begin{subfigure}{0.12\textwidth}\centering
    \caption*{\tiny \\sock:\\ 0.54}
    \includegraphics[width=1\textwidth]{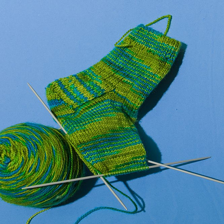}
    \end{subfigure} &
    \begin{subfigure}{0.12\textwidth}\centering
    \caption*{\tiny \\$\rightarrow$ sock: \\1.00}
    \includegraphics[width=1\textwidth]{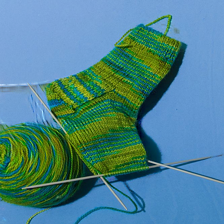}
    \end{subfigure} &
    \begin{subfigure}{0.12\textwidth}\centering
    \caption*{\tiny \\$\rightarrow$ sock: \\1.00}
    \includegraphics[width=1\textwidth]{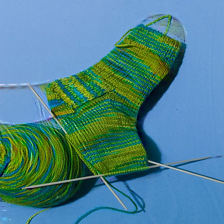}
    \end{subfigure} &
    \begin{subfigure}{0.12\textwidth}\centering
    \caption*{\tiny \\$\rightarrow$ sock: \\1.00}
    \includegraphics[width=1\textwidth]{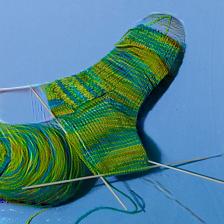}
    \end{subfigure}\\
    \hline

    \begin{subfigure}{0.12\textwidth}\centering
    \caption*{\tiny bearskin:\\ 1.00}
    \includegraphics[width=1\textwidth]{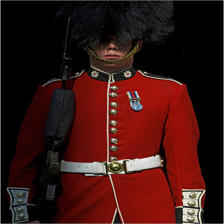}
    \end{subfigure} &
    \begin{subfigure}{0.12\textwidth}\centering
    \caption*{\tiny $\rightarrow$ sombrero: \\0.00}
    \includegraphics[width=1\textwidth]{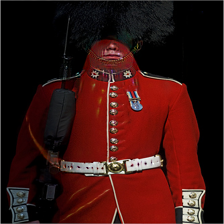}
    \end{subfigure} &
    \begin{subfigure}{0.12\textwidth}\centering
    \caption*{\tiny $\rightarrow$ sombrero: \\0.58}
    \includegraphics[width=1\textwidth]{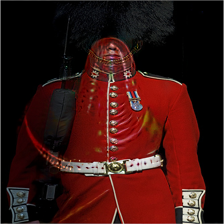}
    \end{subfigure} &
    \begin{subfigure}{0.12\textwidth}\centering
    \caption*{\tiny $\rightarrow$ sombrero: \\0.98}
    \includegraphics[width=1\textwidth]{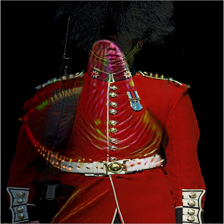}
    \end{subfigure} &
    
    \begin{subfigure}{0.12\textwidth}\centering
    \caption*{\tiny flagpole:\\ 0.72}
    \includegraphics[width=1\textwidth]{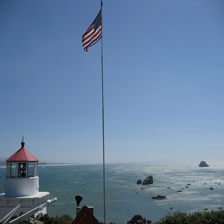}
    \end{subfigure} &
    \begin{subfigure}{0.12\textwidth}\centering
    \caption*{\tiny $\rightarrow$ crutch: \\0.02}
    \includegraphics[width=1\textwidth]{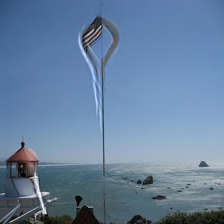}
    \end{subfigure} &
    \begin{subfigure}{0.12\textwidth}\centering
    \caption*{\tiny $\rightarrow$ crutch: \\0.80}
    \includegraphics[width=1\textwidth]{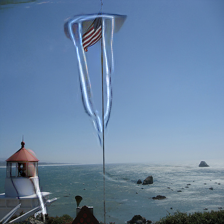}
    \end{subfigure} &
    \begin{subfigure}{0.12\textwidth}\centering
    \caption*{\tiny $\rightarrow$ crutch: \\0.99}
    \includegraphics[width=1\textwidth]{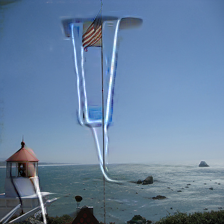}
    \end{subfigure}\\
    \hline

    \begin{subfigure}{0.12\textwidth}\centering
    \caption*{\tiny eel:\\ 0.05}
    \includegraphics[width=1\textwidth]{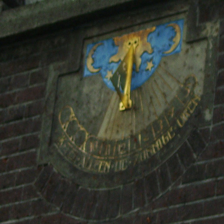}
    \end{subfigure} &
    \begin{subfigure}{0.12\textwidth}\centering
    \caption*{\tiny $\rightarrow$ hourglass: \\0.97}
    \includegraphics[width=1\textwidth]{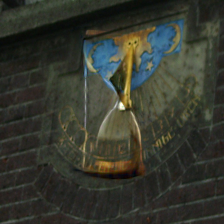}
    \end{subfigure} &
    \begin{subfigure}{0.12\textwidth}\centering
    \caption*{\tiny $\rightarrow$ hourglass: \\1.00}
    \includegraphics[width=1\textwidth]{images/imagenet_appendix_examples/random/4_vce_0.png}
    \end{subfigure} &
    \begin{subfigure}{0.12\textwidth}\centering
    \caption*{\tiny $\rightarrow$ hourglass: \\1.00}
    \includegraphics[width=1\textwidth]{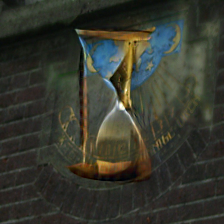}
    \end{subfigure} &
    
    \begin{subfigure}{0.12\textwidth}\centering
    \caption*{\tiny soccer ball:\\ 1.00}
    \includegraphics[width=1\textwidth]{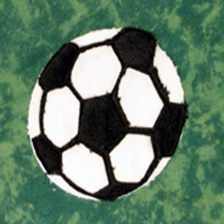}
    \end{subfigure} &
    \begin{subfigure}{0.12\textwidth}\centering
    \caption*{\tiny $\rightarrow$ tennis ball: 0.00}
    \includegraphics[width=1\textwidth]{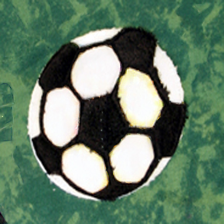}
    \end{subfigure} &
    \begin{subfigure}{0.12\textwidth}\centering
    \caption*{\tiny $\rightarrow$ tennis ball: 0.07}
    \includegraphics[width=1\textwidth]{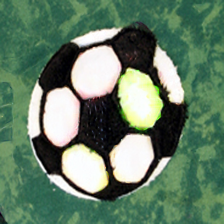}
    \end{subfigure} &
    \begin{subfigure}{0.12\textwidth}\centering
    \caption*{\tiny $\rightarrow$ tennis ball: 0.57}
    \includegraphics[width=1\textwidth]{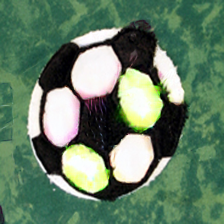}
    \end{subfigure}\\
    \hline
    \begin{subfigure}{0.12\textwidth}\centering
    \caption*{\tiny goldfish:\\ 0.67}
    \includegraphics[width=1\textwidth]{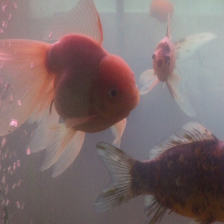}
    \end{subfigure} &
    \begin{subfigure}{0.12\textwidth}\centering
    \caption*{\tiny $\rightarrow$ lionfish: \\0.92}
    \includegraphics[width=1\textwidth]{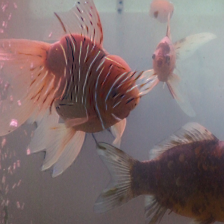}
    \end{subfigure} &
    \begin{subfigure}{0.12\textwidth}\centering
    \caption*{\tiny $\rightarrow$ lionfish: \\1.00}
    \includegraphics[width=1\textwidth]{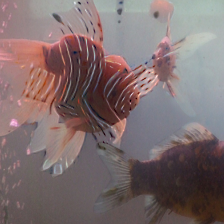}
    \end{subfigure} &
    \begin{subfigure}{0.12\textwidth}\centering
    \caption*{\tiny $\rightarrow$ lionfish: \\1.00}
    \includegraphics[width=1\textwidth]{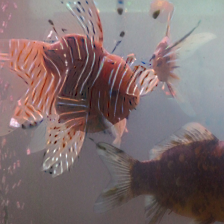}
    \end{subfigure} &
    \begin{subfigure}{0.12\textwidth}\centering
    \caption*{\tiny spindle:\\ 0.48}
    \includegraphics[width=1\textwidth]{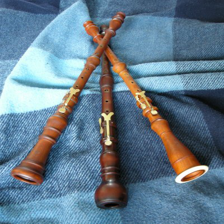}
    \end{subfigure} &
    \begin{subfigure}{0.12\textwidth}\centering
    \caption*{\tiny $\rightarrow$ flute: \\0.44}
    \includegraphics[width=1\textwidth]{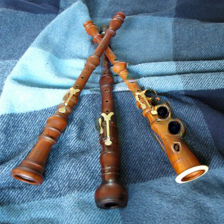}
    \end{subfigure} &
    \begin{subfigure}{0.12\textwidth}\centering
    \caption*{\tiny $\rightarrow$ flute: \\0.80}
    \includegraphics[width=1\textwidth]{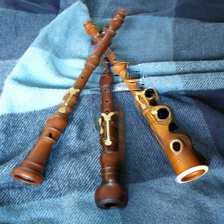}
    \end{subfigure} &
    \begin{subfigure}{0.12\textwidth}\centering
    \caption*{\tiny $\rightarrow$ flute: \\0.95}
    \includegraphics[width=1\textwidth]{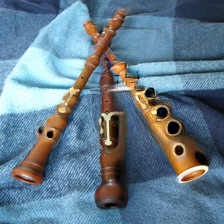}
    \end{subfigure}\\
    \hline
    \end{tabular}
     
      \caption{\label{fig:app-rand-IN} \textbf{Random selection of $l_{1.5}$-VCEs on ILSVRC2012 validation images for Madry\cite{robustness_short}+FT}. The target class is randomly chosen from other classes that are contained in the same WordNet cluster. We note that sometimes it is not possible to transform an image into an image from the target in class with the given budget. This can for example be observed from the soccer ball which mostly contains either black or white pixels. To turn this image into a tennis ball, the attack has to turn those pixel values that lie in the corners of the RGB-cube into yellow, which is not possible with an $l_{1.5}$ budget of 100. We note that even though the soccer ball to tennis ball example can be regarded as failure cases, as it does not visually transform the image into the target class, it is important to note that the end confidence is relatively small. After calibration, it is typically possible to achieve a confidence of $97.0\%$ or higher on valid VCEs whereas the confidence on those two images stays below $80.0\%$ even for the largest radius.
      For other images, like the compass to digital watch, we can see that class-specific features like the numbers from a digital watch appear and integrate well with the remaining image, even though the budget is not sufficiently large to completely change the image subject.
      The flagpole to crutch VCE shows a ghosting artifact that can appear if the attack is not able to integrate the target object into the image. In this case, it paints the crutch on top of the flagpole as there is no reasonable way to integrate it with either the lighthouse or the sea in the background.
      }
     \end{figure*}

    In \cref{fig:app-rand-IN} we show randomly selected pairs of original images and $l_{1.5}$-VCEs where the target class is chosen randomly from the same WordNet clusters described previously. Nevertheless several of the chosen target classes are close to impossible to realize with the given budget.
     
\begin{figure*}[hbtp!]
        \centering
        \includegraphics[width=1\columnwidth]{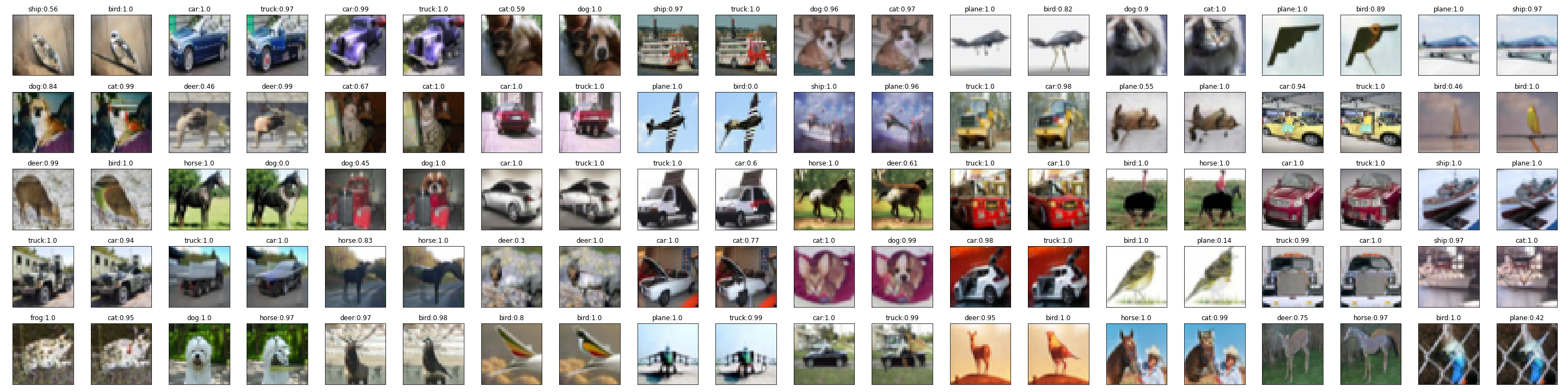}
        \caption{\label{fig:app-rand-CIFAR10} \textbf{Random CIFAR10 images together with their $l_{1.5}$-VCE for $\epsilon=6$} Target class is the second most probable class predicted by the classifier for this image. We see that for CIFAR10 due to the smaller image resolution the budget is sufficient to produce realistic counterfactuals even for pairs of true and target class which are distinct e.g. ``deer'' and ``bird'' or ``ship and ``cat''. However, there are cases where with the given budget the target class cannot be reached respectively there are artefacts remaining from the original image. On the other hand one can note that the changes are often quite subtle and most of the time inserted at the correct position in the image. }
        \label{fig:random-cifar10-imgs}
    \end{figure*}

     For CIFAR10 this is different as due to the lower image resolution of $32 \times 32$ $l_{1.5}$-VCEs are even possible for quite distinct pairs of classes, see \cref{fig:app-rand-CIFAR10}. Note that here the target class is the second most likely class predicted by the classifier for the original image. The model is here again GU+FT, that is the SOTA $l_2$-robust model of \cite{gowal2020uncovering_short} fine-tuned \cite{croce2021adversarial_short} for multiple-norm robustness.
     
    \clearpage
    \section{Comparing threat models}\label{app:comparing-threat-models}
    For the paper, the radii were chosen such that on average the confidence of the classifier in the target class is high and meaningful changes are visible and the perturbation budget is comparable (the avg $\ell_2$-radius of the $\ell_{1.5}$-VCE (we use $\epsilon_{1.5}=50)$ is $11.0$ which is very close to $\epsilon_2=12$ used for $\ell_2$-VCEs). As confirmed by the user study $\ell_2$-VCEs are inferior to $\ell_{1.5}$-VCE. An example VCE is shown below for different radii for $\ell_1/\ell_{1.5}/\ell_2$ where the $\ell_2$-VCEs either do not show meaningful changes for the target class ``geyser'' on the background or are far away from the original image. The $\ell_1$-VCEs show color artefacts and changes are too sparse.   
    \setlength{\tabcolsep}{1pt}
\begin{figure}[hbt]
\vspace{-2mm}
     \centering
     \small
     \begin{tabular}{c|cccc}
     
     \hline
     {{\footnotesize Original}
     } & 
     {{\footnotesize $\epsilon_1$=$200$}
     } & 
     {{\footnotesize $\epsilon_1$=$400$}
     } & 
     {{\footnotesize $\epsilon_1$=$600$}
     } &
     {{\footnotesize $\epsilon_1$=$800$}
     }\\
     \hline
     \begin{subfigure}{0.2\textwidth}\centering
     
     \includegraphics[width=1\textwidth]{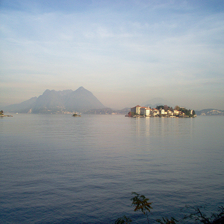}
     \end{subfigure} &
     \begin{subfigure}{0.2\textwidth}\centering
     
     \includegraphics[width=1\textwidth]{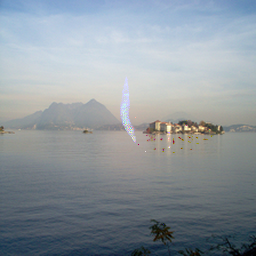}
     \end{subfigure} &
     
     \begin{subfigure}{0.2\textwidth}\centering
     
     \includegraphics[width=1\textwidth]{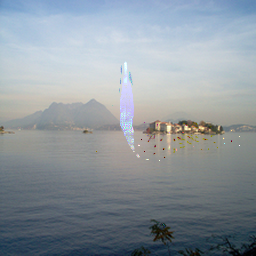}
     \end{subfigure} &
     \begin{subfigure}{0.2\textwidth}\centering
     
     \includegraphics[width=1\textwidth]{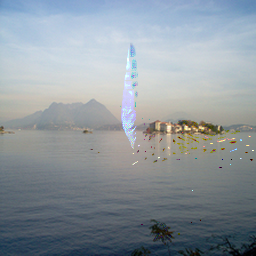}
     \end{subfigure} &
     \begin{subfigure}{0.2\textwidth}\centering
     
     \includegraphics[width=1\textwidth]{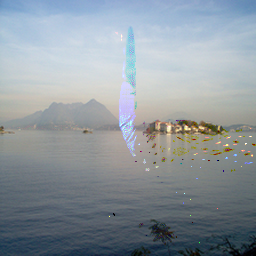}
     \end{subfigure}\\

     \cline{2-5}
      & 
     {{\footnotesize $\epsilon_{1.5}$=$25$}
     } & 
     {{\footnotesize $\epsilon_{1.5}$=$50$}
     } & 
     {{\footnotesize $\epsilon_{1.5}$=$75$}
     } &
     {{\footnotesize $\epsilon_{1.5}$=$100$}
     }\\
     \cline{2-5}
      &
     \begin{subfigure}{0.2\textwidth}\centering
     
     \includegraphics[width=1\textwidth]{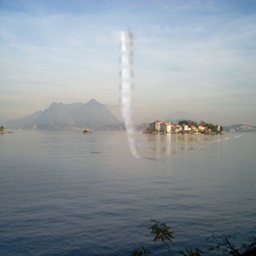}
     \end{subfigure} &
     
     \begin{subfigure}{0.2\textwidth}\centering
     
     \includegraphics[width=1\textwidth]{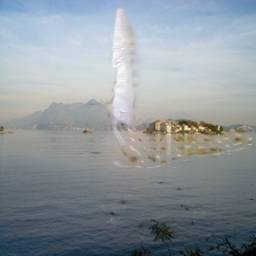}
     \end{subfigure} &
     \begin{subfigure}{0.2\textwidth}\centering
     
     \includegraphics[width=1\textwidth]{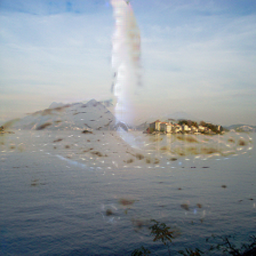}
     \end{subfigure} &
     \begin{subfigure}{0.2\textwidth}\centering
     
     \includegraphics[width=1\textwidth]{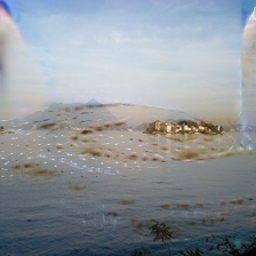}
     \end{subfigure}\\

     \cline{2-5}
     & 
     {{\footnotesize $\epsilon_2$=$6$}
     } & 
     {{\footnotesize $\epsilon_2$=$12$}
     } & 
     {{\footnotesize $\epsilon_2$=$18$}
     } &
     {{\footnotesize $\epsilon_2$=$24$}
     }\\
     \cline{2-5}
      &
     \begin{subfigure}{0.2\textwidth}\centering
     
     \includegraphics[width=1\textwidth]{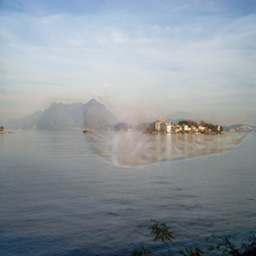}
     \end{subfigure} &
     
     \begin{subfigure}{0.2\textwidth}\centering
     
     \includegraphics[width=1\textwidth]{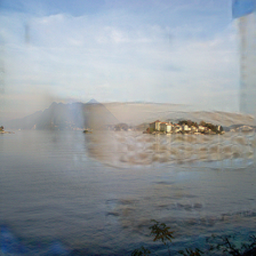}
     \end{subfigure} &
     \begin{subfigure}{0.2\textwidth}\centering
     
     \includegraphics[width=1\textwidth]{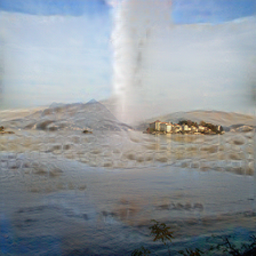}
     \end{subfigure} &
     \begin{subfigure}{0.2\textwidth}\centering
     
     \includegraphics[width=1\textwidth]{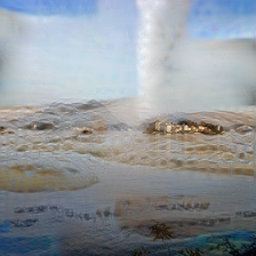}
     \end{subfigure}\\
     \cline{2-5}
     \end{tabular}

     \vspace{-2mm}
     \end{figure}

    \end{document}